\documentclass[twoside]{article}

\usepackage[accepted]{aistats2021}
\usepackage[colorlinks,linkcolor=blue,citecolor=magenta]{hyperref}
\usepackage{url}
\usepackage[pdftex,dvipsnames]{xcolor}
\usepackage{enumitem, comment, xifthen}
\usepackage{graphicx}
\usepackage{booktabs}       
\usepackage{nicefrac}       
\usepackage{microtype}      

\usepackage{subfigure}
\usepackage{amsmath,amssymb,amsthm, amsfonts}
\usepackage[T1]{fontenc}
\usepackage[utf8]{inputenc}
\usepackage{mathtools}
\usepackage[english]{babel}
\usepackage{mwe}
\usepackage{tcolorbox}

\usepackage[toc,page,header]{appendix}
\usepackage{algpseudocode}
\usepackage{algorithm}
\usepackage[ruled,algo2e]{algorithm2e}
\usepackage{setspace}
\usepackage{framed}
\usepackage{mdframed}
\usepackage{cite}
\usepackage{framed}
\usepackage{mdframed}
\usepackage{float}
\usepackage{wrapfig}

\usepackage{blindtext}
\usepackage{enumitem}
\usepackage{amssymb} 
\usepackage{pifont}
\usepackage{xargs}    
\usepackage{soul}
\usepackage[stable]{footmisc}
\usepackage{multirow}
\usepackage{makecell}

\newtheorem{theorem}{Theorem}[section]
\newtheorem{lemma}[theorem]{Lemma}

\newtheorem{assumption}{Assumption}
\newtheorem{proposition}{Proposition}

\newtheorem{definition}{Definition}

\allowdisplaybreaks

\newcommand{\ssymbol}[1]{^{\@fnsymbol{#1}}}

\definecolor{my-green}{cmyk}{0.2, 0.04, 0.1, 0.04, 0.8}
\definecolor{classicrose}{rgb}{0.98, 0.8, 0.91}

\definecolor{blizzardblue}{rgb}{0.67, 0.9, 0.93}

\usepackage{epsf}
\usepackage{graphics}
\usepackage{psfrag}

\usepackage{color}

\usepackage{mathtools}
\usepackage{amsfonts}
\usepackage{amsthm}
\usepackage{amsmath}
\usepackage{amssymb}
\usepackage{xcolor}

\long\def\comment#1{}

\newcommand{\bm}[1]{\boldsymbol{#1}}

\begin{document}
 
\runningtitle{Local Stochastic Gradient Descent Ascent}

\runningauthor{Deng, Mahdavi}

\twocolumn[

\aistatstitle{Local Stochastic Gradient Descent Ascent: \\Convergence Analysis and Communication Efficiency}

\aistatsauthor{Yuyang Deng \And Mehrdad Mahdavi }

\aistatsaddress{The Pennsylvania State University
 \And The Pennsylvania State University
 } ]

\begin{abstract}
Local SGD is a promising approach to overcome the communication overhead in
distributed learning by reducing the synchronization frequency among
worker nodes. Despite the recent theoretical advances of local SGD in empirical risk minimization, the efficiency of its counterpart in minimax optimization remains unexplored. Motivated by large scale minimax learning problems, such as adversarial robust learning and training generative adversarial networks (GANs), we propose  local Stochastic Gradient Descent Ascent (local SGDA), where the primal and dual variables can be trained locally and averaged periodically to significantly reduce the number of communications. We show that local SGDA can provably optimize distributed minimax problems in both homogeneous and heterogeneous data with reduced number of communications and establish convergence rates under strongly-convex-strongly-concave and nonconvex-strongly-concave settings. In addition, we propose a novel variant local SGDA+, to solve nonconvex-nonconcave  problems.  We   give corroborating empirical evidence on different distributed minimax  problems. 
\end{abstract}

 \section{Introduction}

We study  minimax optimization problems of the form
\begin{equation}
    \min_{\bm{x}\in \mathbb{R}^{d_x}} \max_{\bm{y}\in\mathbb{R}^{d_y}} \left\{F(\bm{x},\bm{y}) := \frac{1}{n}\sum_{i=1}^n f_i(\bm{x},\bm{y})\right\}, \label{eq: objective}
\end{equation}
where data are distributed across $n$ nodes so that each node $i$ will have its own objective function $f_i(\cdot,\cdot)$. The local objective function is defined as $f_i(\cdot,\cdot)=\mathbb{E}_{\xi\sim \mathcal{D}_i}[\ell(\cdot,\cdot;\xi)]$, where $\mathcal{D}_i$ is the local data distribution of $i$th client and $\ell$ is the loss function.
Numerous machine learning problems fall in this category. A canonical instance is adversarially robust learning. Consider the following robust linear regression $ \min_{\bm{x} \in\mathbb{R}^{d_x}} \max_{\bm{y}\in \mathbb{R}^{d_y}}  \frac{1}{n}\sum_{i=1}^n \ell (\bm{x}^{\top}(\bm{a}_i+\bm{y});\bm{b}_i) +\frac{\lambda_x}{2}\|\bm{x}\|^2 -\frac{\lambda_y}{2}\|\bm{y}\|^2 
$, where $\{(\bm{a}_i, \bm{b}_i)\}_{i=1}^n$ are input pairs of training data. We wish to learn a predictor $\bm{x}$ that is robust to small perturbation $\bm{y}$. Another popular minimax application is Generative Adversarial Network (GAN)~\cite{goodfellow2014generative}, which can be formulated as:
$   \min_{\bm{x} \in\mathbb{R}^{d_x}} \max_{\bm{y}\in \mathbb{R}^{d_y}} \mathbb{E}_{\bm{a} \sim \mathcal{D}_{real}}[\ell (D_{\bm{y}}(\bm{a}))] + \mathbb{E}_{\bm{a} \sim \mathcal{D}_{\bm{x}}}[\ell (1-D_{\bm{y}}(\bm{a}))],
$ where $\bm{x}$ is the parameter of the generator network  $\mathcal{D}_{\bm{x}}$ and $\bm{y}$ is the parameter of the discriminator network $D_{\bm{y}}$.

The centrality of these applications in machine learning motivates considerable interest in efficiently solving minimax optimization problems. Among all popular algorithms, primal-dual stochastic gradient algorithms are definitely the most popular ones~\cite{nesterov2007dual,nemirovski2004prox,tseng2008accelerated}. The most classic algorithm in this category is stochastic gradient descent ascent (SGDA), which has been  proven to be an effective algorithm for minimax optimization both empirically and theoretically~\cite{lin2019gradient}. However, in practice, due to the huge volume of data, or to protect the privacy of user data (e.g., federated learning scenario~\cite{konevcny2016bfederated,konevcny2016federated}), a distributed algorithm while lowering the communication cost is preferred and is the focus of this paper. A conventional distributed approach to solve~(\ref{eq: objective}) is \emph{parameter server} model, where every client (user) sends its local stochastic gradient to a central node, and the central node performs stochastic gradient descent procedure on primal and dual variables by aggregating local stochastic gradients. Unfortunately, this approach causes heavy communication outage, which has been reported to be the main bottleneck slowing down the distributed optimization~\cite{alistarh2017qsgd,lin2017deep,seide20141,zhang2017zipml}.

\begin{table*}[t!]
\centering
\footnotesize
\resizebox{2\columnwidth}{!}{%
\begin{tabular}{|l|l|c|c|c|}
\hline
Assumption & Setting & Results & Comm. Rounds  &  Convergence Rate \\  
 \hline
 \\[-1em] 
\multirow{2}{*} {Strongly-Convex-Strongly-Concave} & Homogeneous& Theorem~\ref{Thm: IID_SCSC}& $\Tilde{\Omega}\left(n\right)$  &   $ \Tilde{O}\left(\frac{\kappa^2\sigma^2 }{\mu^2nT}\right)$   \\    \cline{2-5}
  & Heterogeneous& Theorem~\ref{Thm:  SCSC}&$\Omega\left(\sqrt{nT}\right)$& $  O\left(\frac{\kappa^2  \left(\Delta_{x}+\Delta_{y}+\sigma^2\right)}{\mu nT} \right)$   \\ \hline
\multirow{2}{*} {Nonconvex-Strongly-Concave} & Homogeneous &Theorem~\ref{Thm: NCSC}&$ \Omega\left(n^{1/3}T^{2/3}\right)$  &  $O\left( \frac{L^2\sigma^2}{(nT)^{1/3}} \right)$   \\    \cline{2-5}
  & Heterogeneous & Theorem~\ref{Thm: NCSC} &$ \Omega\left(n^{1/3}T^{2/3}\right)$ & $O\left(  \frac{L^2\sigma^2}{(nT)^{1/3}} + \frac{L^2\zeta_x}{T^{2/3}}+ \frac{L^2\zeta_y}{n^{2/3}T^{1/3}}\right)$  \\ \hline  
 \multirow{2}{*} {Nonconvex-PL condition} & Homogeneous & Theorem~\ref{Thm: NCNC}  & $ \Omega\left( T^{2/3}\right)$ &  $O\left(\frac{\beta \sigma^2}{(nT)^{1/3}}  \right)$   \\    \cline{2-5}
  & Heterogeneous&Theorem~\ref{Thm: NCNC}   &$ \Omega\left( T^{2/3}\right)$ & $O\left(\frac{\beta \sigma^2}{(nT)^{1/3}}+\frac{\kappa^2 L^2 \zeta_y}{n^{2/3}T^{1/3}} + \frac{\kappa^2 L^2\zeta_x}{ n^{2/3}T} \right)$  \\ \hline  
  \multirow{2}{*} {Nonconvex-One-Point-Concave} & Homogeneous & Theorem~\ref{Thm: NCOC}  & $ \Omega\left( T^{2/3}\right)$ &  $O\left(\frac{L\sigma^2 }{T^{1/6}} \right)$   \\    \cline{2-5}
  & Heterogeneous&Theorem~\ref{Thm: NCOC}   &$ \Omega\left( T^{2/3}\right)$ & $O\left(\frac{L\sigma^2 }{T^{1/6}} +    \frac{L^2\zeta_x}{n^{1/3}T}   + \frac{L^2\zeta_y}{(nT)^{1/3}}\right)$  \\ \hline  
\end{tabular}} \caption{ A summary of our results under different settings. We use $\Tilde{O}(\cdot)$ and $\Tilde{\Omega}(\cdot)$ to hide logarithmic term. $\Delta_x$ and $\Delta_y$ are heterogeneity at the optimum (see Definition~\ref{def: hetero}). $\zeta_x$ and $\zeta_y$ denote gradient dissimilarity (see Definition~\ref{def: dissimilarity}).}
\label{tab:results}
\end{table*}

A notable research effort to reduce the commutation complexity under a computation budget is to employ local SGD with periodic averaging~\cite{mcmahan2017communication,stich2018local}. In local SGD, the idea is to perform multiple local updates, wherein clients update their own local models via SGD for multiple iterations, and the models of the different clients are averaged periodically. While this algorithm introduces additional noise due to local updates over fully synchronous SGD, it is shown that by careful choice of learning rate, local SGD can achieve same asymptotic performance as synchronous SGD, while benefiting from reduced communication rounds~\cite{stich2018local,yu2018parallel,wang2018cooperative,haddadpour2019local,khaled2019better,lin2018don}. Motivated by the success of local SGD and a key observation that in some minimax applications (e.g., aforementioned robust linear regression and GANs), the primal and dual variables can be trained in a distributed manner and locally, we extend local SGD to tackle minimax learning problems and propose \emph{local stochastic gradient descent ascent} (local SGDA) algorithm. In local SGDA, local nodes will optimize their own version of primal and dual variables for multiple steps, and then they synchronize and do model averaging via central server. However, despite it being an extremely simple algorithm, and the thorough understanding of local SGD on minimization problem, local SGDA, as its counter-part in minimax problem, still lacks theoretical foundations. Thus, a natural question that arises is: \emph{\textbf{Does local SGDA provably optimize distributed minimax problems too?}} 

We answer above question in the affirmative, by establishing the  convergence rate of local SGDA in both \textbf{homogeneous data setting}, where  local functions in (\ref{eq: objective}) have the same distribution (IID), i.e., $\mathcal{D}_1 = \ldots =  \mathcal{D}_n = \mathcal{D}$, and \textbf{heterogeneous data setting}, where local functions are not necessarily realized by the same distribution (non-IID). Our main contributions can be summarized as follows. We are the first to show that local SGDA provably optimizes the distributed minimax problem with communication efficiency, on both homogeneous and heterogeneous data.  For strongly-convex-strongly-concave setting,  we obtain the convergence rate of $\Tilde{O}\left(\frac{1}{nT } \right)$ with $ \Tilde{\Omega}(n)$ communication rounds in homogeneous local functions setting, and $O\left(\frac{\Delta_x + \Delta_y}{nT } \right)$ with $ \Omega(\sqrt{nT})$ communication rounds in heterogenous setting, where $\Delta_x + \Delta_y$ is the quantity reflecting heterogeneity. It recovers the same asymptotic rate and communication rounds as local SGD in the smooth strongly-convex minimization problem~\cite{khaled2019tighter,woodworth2020local,woodworth2020minibatch}, up to a constant factor. For nonconvex-strongly-concave problem, we get the rate
$ O\left(\frac{1}{(nT)^{1/3}} \right)$
 with $\Omega\left(T^{2/3}\right)$ communication rounds, under both data allocation settings.  In addition, in order to efficiently solve the nonconvex-nonconcave minimax optimization problems, we propose a variant of local SGDA, dubbed as local SGDA+, a single loop algorithm to solve nonconvex-nonconcave problems. We establish its convergence rate on two classes of functions, which are nonconvex in $\bm{x}$ but satisfies Polyak-{\L}ojasiewicz (PL) condition in $\bm{y}$~\cite{karimi2016linear}, and nonconvex in $\bm{x}$ but one-point-concave in $\bm{y}$. We summarize the  obtained rates for different settings in Table~\ref{tab:results}.

\section{Prior Art}
\noindent\textbf{Single Machine Minimax Optimization.}~The history of minimax optimization dates back to Brown~\cite{brown1951iterative}, where he proposed a bilinear form minimax problem. Korpelevich~\cite{korpelevich1976extragradient} then proposed  the extra gradient (EG) method to solve this bilinear problem. Following their path, Nemirovski~\cite{nemirovski2004prox},  Nesterov~\cite{nesterov2007dual} and Tseng~\cite{tseng2008accelerated} studied the general smooth convex-concave minimax problem, and proposed algorithms which achieve the same asymptotic rate $O\left(1/T\right)$. Du and Hu~\cite{du2019linear} prove the linear convergence of primal-dual gradient method on a class of convex-concave functions.  The other popular algorithm for convex-concave optimization is Optimistic Gradient Descent Ascent (OGDA), which is widely studied and has many applications in machine learning ~\cite{daskalakis2018training,liang2019interaction,mokhtari2019convergence}. For strongly-convex-concave setting, Thekumprampil et al~\cite{thekumparampil2019efficient} proposed an algorithm combing Nesterov accelerated gradient descent and Mirror-Prox, which achieves near optimal rate $\Tilde{O}(1/T^2)$. For strongly-convex-strongly-concave setting, Lin et al~\cite{lin2020near} leveraged the idea of accelerated gradient descent, and gave a nearly optimal minimax algorithm, which matches the lower bound given in~\cite{ouyang2019lower}. Some literature~\cite{lin2019gradient,nouiehed2019solving,rafique2018non,thekumparampil2019efficient} also conduct trials on nonconvex-concave minimax optimization, and among them the most related work to us is~\cite{lin2019gradient}, where they study the single machine SGDA, under nonconvex-(strongly)-concave case. Recently, due to the raise of GANs~\cite{goodfellow2014generative}, a vast amount of work is
devoted to nonconvex-nonconcave optimization~\cite{gidel2018variational,liu2019towards,liu2019decentralized}.

\noindent\textbf{Distributed Minimax Optimization.}~A few recent studies are devoted to decentralized  minimax optimization. Srivastava et al~\cite{srivastava2011distributed} proposed a decentralized algorithm to solve the convex-concave saddle point problem over a network. Mateos and Cortes~\cite{mateos2015distributed} proposed a subgradient method and prove the convergence under convex-concave case. Liu et al~\cite{liu2019decentralized} analyzed the convergence of networked optimistic stochastic gradient descent ascent (OSGDA) on nonconvex-nonconcave setting.~\cite{reisizadeh2020robust} studied a variant of local SGDA, and provided the convergence analysis on PL-PL and nonconvex-PL objective.  We note that~\cite{beznosikov2020local} also studies the convergence of local SGDA on strongly-convex-strongly-concave setting, but their analysis is not as tight as ours.  Recently, federated adversarial training~\cite{reisizadeh2020robust} and FedGAN~\cite{rasouli2020fedgan} are proposed to solve large-scale and privacy-preserving minimax problem, which can be seen as application instances of our work.

\noindent\textbf{Local SGD.}~Communication efficiency has been studied extensively in distributed SGD. The most related idea to this paper is local SGD or FedAvg~\cite{mcmahan2017communication}. FedAvg is firstly proposed by Mcmahan et al~\cite{mcmahan2017communication} to alleviate communication bottleneck in the distributed machine learning. Stich~\cite{stich2018local} was the first to prove that local SGD achieves $O\left(1/T\right)$ convergence rate with only $O(\sqrt{T})$ communication rounds on  IID data for smooth strongly-convex loss functions. Haddadpour et al~\cite{haddadpour2019local} analyzed the convergence of local SGD on nonconvex (PL condition) function, and proposed an adaptive synchronization scheme. \cite{khaled2019tighter} gave the tighter bound of local SGD, which directly reduces the $O(\sqrt{T})$ communication rounds in \cite{stich2018local} to $O(n)$, under smooth strongly-convex setting. Recently, Yuan and Ma~\cite{yuan2020federated} proposed the first accelerated local SGD, which further reduced the communication rounds to  $O(n^{1/3})$.~\cite{haddadpour2019convergence} gave the  analysis of local GD and SGD on smooth nonconvex functions in non-IID setting. Li et al~\cite{li2019convergence}  analyzed the convergence of FedAvg under non-IID data for strongly convex functions. \cite{woodworth2020local,woodworth2020minibatch} investigated  the difference between local SGD and mini-batch SGD, in both homogeneous and heterogeneous data settings.

\section{Local SGDA}

\begin{algorithm2e}[t!]
	\DontPrintSemicolon
    \caption{\texttt{Local SGDA}}
	\label{algorithm: SGDA}
	\textbf{input:}  Synchronization gap $\tau$, Communication rounds $S$, Number of iterations $T = S\tau$,  Initial local models $\bm{x}^{(0)}_{i}$, $\bm{y}^{(0)}_{i}$  for $i\in [n]$.
	\\
	\textbf{parallel} \For{$i = 1,...,n$}{
	 
	 \For{$s = 0, \ldots, S-1$}{
	 all nodes send their local model ${\bm{x}}^{(s\tau)}_i$ and ${\bm{y}}^{(s\tau)}_i$ to server.\\
	 $ \bm{x}^{(s\tau)} = \frac{1}{n}\sum_{i=1}^n {\bm{x}}^{(s\tau)}_i$\\
 	        $ \bm{y}^{(s\tau)} = \frac{1}{n}\sum_{i=1}^n {\bm{y}}^{(s\tau)}_i$\\
 	        server sends $ \bm{x}^{(s\tau)}$, $ \bm{y}^{(s\tau)}$ to all nodes;\\
 	        each client initializes its local models: $\bm{x}_i^{(s\tau)} = \bm{x}^{(s\tau)}$ and $\bm{y}_i^{(s\tau)} = \bm{y}^{(s\tau)}$. \\
	 \For{$t = s\tau , \ldots, (s+1)\tau-1$}{
	             sample a minibatch $\xi_i^t$ from local data\\
	 	     {${\bm{x}}^{(t+1)}_i = {\bm{x}}^{(t)}_i - \eta_x \nabla_{x} f_i\left({\bm{x}}^{(t)}_i,{\bm{y}}^{(t)}_i;\xi_i^t\right)$}  \\
        { ${\bm{y}}^{(t+1)}_i =  {\bm{y}}^{(t)}_i + \eta_y \nabla_{y} f_i\left({\bm{x}}^{(t)}_i,{\bm{y}}^{(t)}_i;\xi_i^t\right)$}\\ 
	 
	 }}
	
  }
\end{algorithm2e}

In this section we formally introduce local SGDA algorithm  for solving distributed minimax problems.  The proposed algorithm can be viewed as a variant of SGDA, which  is one of the most popular primal-dual stochastic gradient algorithm to solve centralized minimax optimization problems.  Specifically, 
for solving the optimization problem in~(\ref{eq: objective}), at $t$th iteration,  SGDA performs the following updates on primal and dual variables:
\begin{equation*}
    \begin{aligned}
    {\bm{x}}^{(t+1)}  &=  {\bm{x}}^{(t)}  - \eta_x \nabla_{x}F\left({\bm{x}}^{(t)}  ,{\bm{y}}^{(t)} ;\xi^t\right)\\
    {\bm{y}}^{(t+1)} &=   {\bm{y}}^{(t)}  + \eta_y \nabla_{y} F\left({\bm{x}}^{(t)} ,{\bm{y}}^{(t)} ;\xi^t\right),
    \end{aligned}
\end{equation*}
where $\xi^t$ is minibatch sampled at $t$th iteration to compute  stochastic gradient, and  $\eta_x$ and $\eta_y$ are learning rates.

The key difficulty of  deploying SGDA in a distributed setting stems from the fact that after $t$th updating, server needs to communicate global models $\bm{x}^{(t)}$ and $\bm{y}^{(t)}$ to all nodes, so clients can locally evaluate the gradient on $\bm{x}^{(t)}$ and $\bm{y}^{(t)}$. Meanwhile local users should send their local gradients back to the server for aggregation/averaging. This suffers from heavy communication cost and could hinder the scalability of the algorithm as communication is known to be a major bottleneck that slows down the training process~\cite{alistarh2017qsgd,lin2017deep,seide20141,zhang2017zipml}.

As mentioned earlier, to mitigate the communication bottleneck, a popular idea is to update models locally via SGD, and then average them periodically~\cite{mcmahan2017communication,stich2018local}. Motivated by this, we advocate  a local primal-dual algorithm for minimax optimization as detailed in Algorithm~\ref{algorithm: SGDA}. To formally present the steps of proposed Local SGDA algorithm, consider $S$ as the rounds of communication between server and clients, and $\tau$ as the number of local updates performed by clients between two consecutive communication rounds. The algorithm proceeds for $T = S\tau$ iterations and at $t$th local iteration, the $i$th node locally performs the SGDA on its own local primal and dual variables 
\begin{align}
       &\bm{x}^{(t+1)}_i =  {\bm{x}}^{(t)}_i - \eta_x \nabla_{x} f_i\left({\bm{x}}^{(t)}_i,{\bm{y}}^{(t)}_i;\xi_i^t\right),   \nonumber \\
       &\bm{y}^{(t+1)}_i =  {\bm{y}}^{(t)}_i + \eta_y \nabla_{y} f_i\left({\bm{x}}^{(t)}_i,{\bm{y}}^{(t)}_i;\xi_i^t\right) ,\nonumber
\end{align}
for $\tau$ iterations, where $\xi_i^t$ is the minibatch sampled by $i$th client from its local data to compute local stochastic gradient at iteration $t$. At  $s$th synchronization round, the server aggregates local models $\bm{x}^{(s\tau)}_i$ and $\bm{y}^{(s\tau)}_i$, to perform the averaging: ${\bm{x}}^{(s\tau)} = \frac{1}{n} \sum_{i=1}^n \bm{x}^{(s\tau)}_i$ and ${\bm{y}}^{(s\tau)} = \frac{1}{n} \sum_{i=1}^n \bm{y}^{(s\tau)}_i$.  Then, the server  sends the averaged models back to  local nodes. We note that compared to fully synchronous distributed SGDA, which requires $T$ communication round,  in local SGDA we only require  $T/\tau$ communications.    Despite its simplicity, we are not aware of any prior result that establishes the convergence rate of local methods in minimax setting. In the following sections, we show that the proposed algorithm enjoys a fast convergence rate while significantly reducing the communication rounds by properly choosing the number of local updates $\tau$. 

\section{Strongly-Convex-Strongly-Concave Case}
 
In this section we will present the convergence analysis of local SGDA for strongly-convex-strongly-concave functions, under both homogeneous and heterogeneous data settings. In the strongly-convex-strongly-concave minimax problem, our goal is to find the \emph{saddle point} of global objective, as  defined below:
\begin{definition}
The tuple $(\bm{x}^*, \bm{y}^*)$ is said to be saddle point of convex-concave function $F (\bm{x},\bm{y})$ if  $ F(\bm{x}^*, \bm{y} ) \leq F(\bm{x}^*, \bm{y}^*) \leq   F( \bm{x}^*, \bm{y}),   \forall \bm{x} \in \mathbb{R}^{d_x}, \bm{y} \in \mathbb{R}^{d_y}.$
\end{definition}

To facilitate our analysis, we make the following standard assumptions on objective function and noise of stochastic gradients.

\begin{assumption}[Strong Convexity]\label{assumption: strong convexity}
$f_i(\bm{x},\bm{y})$ is strongly convex in $\bm{x}$, which implies there exists a $\mu > 0$ such that $\forall \bm{x}, \bm{x}'\in \mathbb{R}^{d_x},\bm{y} \in \mathbb{R}^{d_y}$ it holds that
$     f_i(\bm{x},\bm{y}) \geq f_i(\bm{x}',\bm{y}) + \langle \nabla_x f_i(\bm{x}',\bm{y}), \bm{x}'-\bm{x} \rangle + \frac{\mu}{2} \|\bm{x}-\bm{x}'\|^2$.
\end{assumption}

\begin{assumption}[Strong Concavity]\label{assumption: strong concavity}
$f_i(\bm{x},\bm{y})$ is   strongly concave in $\bm{y}$, which implies there exists a $\mu > 0$ such that $\forall \bm{x}\in \mathbb{R}^{d_x}, \bm{y},\bm{y}' \in \mathbb{R}^{d_y}$ it holds that
$ f_i(\bm{x},\textbf{y}) \leq  f_i(\bm{x},\textbf{y}') + \langle \nabla_{y} f_i(\bm{x},\textbf{y}'), \textbf{y}' - \textbf{y} \rangle 
 - \frac{\mu}{2} \|\textbf{y}-\textbf{y}'\|^2$.
\end{assumption}

\begin{assumption}[Smoothness]\label{assumption: smoothness}
There exists a $L > 0$ such that $\forall i \in [n], 
\|\nabla f_i(\bm{x}_1,\bm{y}_1 ) - \nabla f_i(\bm{x}_2, \bm{y}_2)\| \leq L \|(\bm{x}_1,\bm{y}_1 )-(\bm{x}_2, \bm{y}_2)\|, \  \forall \bm{x} \in \mathbb{R}^{d_x}, \bm{y} \in \mathbb{R}^{d_y}.$
\end{assumption}

\begin{assumption}[Bounded Variance]\label{assumption: bounded grad}
The variance of stochastic gradients  computed at each local function is bounded, i.e.,  $\forall i \in [n], \mathbb{E}[\|\nabla_x f_i (\bm{x},\bm{y};\xi) - \nabla_x f_i(\bm{x},\bm{y})\|^2]  \leq \sigma^2$ and $ \mathbb{E}[\|\nabla_y f_i (\bm{x},\bm{y};\xi) - \nabla_y f_i(\bm{x},\bm{y})\|^2]  \leq \sigma^2$.
\end{assumption}
\noindent{\textbf{Main techniques.}}~In our analysis, due to infrequent synchronization, the key  is to bound the deviation among local and global models as defined below
{\begin{align}
 \delta_{\bm{x}}^{(t)} = \frac{1}{n}\sum_{i=1}^n\left\|\bm{x}_i^{(t)}-\bm{x}^{(t)}\right\|^2,  \delta_{\bm{y}}^{(t)} = \frac{1}{n}\sum_{i=1}^n\left\|\bm{y}_i^{(t)}-\bm{y}^{(t)}\right\|^2, \label{eq:deviation bound}
\end{align}}
where $\bm{x}^{(t)} = \frac{1}{n}\sum_{i=1}^n \bm{x}^{(t)}_i$ and $\bm{y}^{(t)} = \frac{1}{n}\sum_{i=1}^n \bm{y}^{(t)}_i$ are (virtual) primal and dual global averages at iteration $t$, respectively. We note that virtual averages are introduced for analysis purposes and only computed at synchronization rounds.

Despite  minimization problems, where we already have a solid theory to bound the  deviation between global and local models~\cite{stich2018local,haddadpour2019local,haddadpour2019convergence,li2019convergence,khaled2019tighter,karimireddy2019scaffold,li2019communication,woodworth2020local,woodworth2020minibatch}, none of these guarantees apply to minimax problem, due to the unstable nature of primal-dual optimization.  Hence the key step in our analysis is to develop a relatively  tight bound for quantities introduced in~(\ref{eq:deviation bound}). In the homogeneous setting, we show that under the dynamic of primal-dual algorithm, and smooth strongly convex assumption, the deviation  $\delta_{\bm{x}}^{(t)}+\delta_{\bm{y}}^{(t)}$ can decrease as the rate of $O(\tau (1+(L-\mu)\eta)^{2\tau} \eta^2 \sigma^2)$. By properly choosing $\tau$ and $\eta$,  we can recover the rate $O(\tau \eta^2 \sigma^2)$, which matches with the existing tightest deviation bounds of local SGD~\cite{khaled2019tighter,woodworth2020local}.   In the heterogeneous setting, we prove that, by carefully controlling the step size, we can develop the deviation bound that depends on distance between the current iterate and the saddle point $(\bm{x}^*,\bm{y}^*)$: $\|\bm{x}^{(t)} - \bm{x}^*\|^2 + \|\bm{y}^{(t)} - \bm{y}^*\|^2$, plus terms that capture heterogeneity. 
\subsection{Convergence  in homogeneous  setting}
We now turn to  stating the convergence rate in homogeneous setting.
\begin{theorem} \label{Thm: IID_SCSC}
Suppose each client's objective function $f_i$ satisfy Assumptions~\ref{assumption: strong convexity},\ref{assumption: strong concavity},\ref{assumption: smoothness},\ref{assumption: bounded grad}. If we use local SGDA (Algorithm~\ref{algorithm: SGDA}) under homogeneous data setting to optimize (\ref{eq: objective}), choosing synchronization gap as $\tau =  \frac{T}{n\log T}  $,  using  learning rate $\eta_x = \eta_y= \frac{4\log T}{\mu T}$,  and by denoting $\kappa = L/\mu$, it   holds that
{\begin{align}
 &\mathbb{E}\left[\left\| \bm{x}^{(T)} -\bm{x}^{*} \right\|^2 + \left\| \bm{y}^{(T)} -\bm{y}^{*} \right\|^2\right]\nonumber\\
 &\leq \Tilde{O}\left(\frac{1}{T^2}  +\frac{ \sigma^2}{\mu^2 nT }  +    \frac{\kappa^2\sigma^2 }{\mu^2nT} + \frac{\kappa^2\sigma^2 }{\mu^2nT^2} \right).\nonumber
\end{align}} 
\end{theorem}
The proof of Theorem~\ref{Thm: IID_SCSC} is deferred to Appendix~\ref{sec:proof_scsc}. It can be observed that we obtain an $\Tilde{O}(\frac{1}{nT})$ convergence rate with only $\Tilde{O}(n)$ communication rounds. This indeed implies that we can achieve a linear speedup in terms of number of clients $n$, while significantly reducing the communication complexity from $T$ (fully synchronous SGDA) to $\Tilde{O}(n)$ in strongly-convex-strongly-concave setting. We also note that the obtained bound  matches the best known rate of local SGD for minimization problems~\cite{khaled2019tighter}, up to a logarithmic factor. The notable difference is that in~\cite{khaled2019tighter}, the communication rounds can be a constant, i.e., $O(n)$,  but in our result, we have an extra logarithmic dependency on $T$. We leave removing this log factor as a  future work.

\subsection{Convergence in heterogeneous setting}
We now turn to stating  the convergence rate of local SGDA for strongly-convex-strongly-concave functions in heterogeneous local data setting. To this end, we first need to decide a proper notion to capture the heterogeneity among local functions by introducing the  following quantities.
\begin{definition}[Heterogeneity at Optimum]\label{def: hetero}
The heterogeneity at global saddle point $(\bm{x}^*, \bm{y}^*)$  is defined as
{\begin{align}
\Delta_x &=  \frac{1}{n}\sum_{i=1}^n \|\nabla_x f_i(\bm{x}^*, \bm{y}^*)\|^2, \nonumber \\
    \Delta_y &=  \frac{1}{n}\sum_{i=1}^n \|\nabla_y f_i(\bm{x}^*, \bm{y}^*)\|^2.\nonumber    
\end{align}
}
\end{definition}
Definition~\ref{def: hetero} is a generalized notion borrowed from~\cite{khaled2019tighter}, where they firstly employ it in the analysis of local SGD. It characterizes the heterogeneity of each local function at the global optimums of primal and dual variables. The following theorem establishes the convergence rate of local SGDA in heterogeneous  settings.
\begin{theorem} \label{Thm: SCSC}
Let each client's objective $f_i$ satisfy Assumptions~\ref{assumption: strong convexity},\ref{assumption: strong concavity},\ref{assumption: smoothness},\ref{assumption: bounded grad}. If we use local SGDA (Algorithm~\ref{algorithm: SGDA}) under heterogeneous data setting to optimize (\ref{eq: objective}), choosing synchronization gap $\tau = \sqrt{T/n}$,  using decreasing learning rate $\eta_x = \eta_y = \eta_t = \frac{8}{\mu(t+a)}$, where $a=\max\big\{  2048\kappa^2\tau , 1024\sqrt{2}\tau \kappa^2, 256\kappa^2\big\}$, $\kappa = L/\mu$, then the following convergence  holds:
{\begin{align*}
  &\mathbb{E}\left[\left\| \bm{x}^{(T)} -\bm{x}^{*} \right\|^2 + \left\| \bm{y}^{(T)} -\bm{y}^{*} \right\|^2\right] 
  \leq O\left(\frac{ a^3}{T^3} \right)\nonumber\\
  &+ O\left(\frac{\sigma^2}{\mu^2 nT} \right)+  O\left(\frac{\kappa^2  \left(\Delta_{x}+\Delta_{y}\right)}{\mu nT} \right) + O\left(\frac{\kappa^2  \sigma^2}{\mu n T } \right). \nonumber
\end{align*}} 
\end{theorem}
The proof of Theorem~\ref{Thm: SCSC} is deferred to Appendix~\ref{sec:proof_scsc}. Here we obtain an $O\left(\frac{\kappa^2  \left(\Delta_{x}+\Delta_{y}\right)}{\mu nT} \right)$  rate using $\sqrt{nT}$ communication rounds, which also  enjoys the linear speedup w.r.t. the number of the nodes. This result recovers the convergence rate of local SGD or FedAvg on strongly-convex minimization problems~\cite{khaled2019tighter}. Our result does not need bounded gradient assumption, and we recover the linear dependency on function heterogeneity at global optimum, which matches the best bound for local SGD in minimization problems~\cite{khaled2019tighter}. The most analogous work to ours is~\cite{beznosikov2020local}, where it achieves an $\Tilde{O}(\frac{1}{T})$ rate with $O(n^{1/3}T^{2/3})$ communication rounds, which is worse than our result.

\section{Nonconvex-Strongly-Concave Case}
In this section we will present the convergence  of local SGDA for nonconvex-strongly-concave functions.  In this setting, since objective is no longer convex, we are unable to show the convergence to global saddle point. Thus, following the  standard machinery in nonconvex-concave analysis~\cite{lin2019gradient,thekumparampil2019efficient,rafique2018non}, we introduce the following envelope function which will prove useful in convergence analysis.
\begin{definition}We define the following envelope functions to facilitate our analysis:
{\begin{align}
     \Phi(\bm{x}) = F(\bm{x},\bm{y}^*(\bm{x})),\bm{y}^*(\bm{x}) = \arg\max_{\bm{y} \in \mathbb{R}^{d_y}} F(\bm{x}, \bm{y}) \label{eq: envelope}.
\end{align}}\vspace{-0.5cm}
\end{definition}
We consider the convergence rate to the first order stationary point of $ \Phi(\bm{x})$, as advocated in seminal nonconvex-concave minimax literature~\cite{lin2019gradient,rafique2018non,thekumparampil2019efficient}. Namely, we will show how fast $\|\nabla \Phi(\bm{x})\|$ vanishes. Our analysis here mainly considers heterogeneous setting, but it can be easily generalized to homogeneous setting as well. We will use the following quantity to measure heterogeneity in nonconvex-strongly-concave case. 
\begin{definition}[Gradient Dissimilarity]\label{def: dissimilarity}
We define the following quantities to measure the gradient dissimilarity among local  functions:
{\begin{align}
   \zeta_x = \sup_{(\bm{x},\bm{y})\in \mathbb{R}^{d_x} \times \mathbb{R}^{d_y}} \frac{1}{n} \sum_{ i=1}^n \left\|  \nabla_x f_i(\bm{x},\bm{y}) -  \nabla_x F(\bm{x},\bm{y})\right\|^2,\nonumber\\
   \zeta_y = \sup_{(\bm{x},\bm{y})\in \mathbb{R}^{d_x} \times \mathbb{R}^{d_y}} \frac{1}{n} \sum_{ i=1}^n \left\|  \nabla_y f_i(\bm{x},\bm{y}) -  \nabla_y F(\bm{x},\bm{y})\right\|^2.\nonumber
\end{align}}
 \end{definition}
 
 Definition~\ref{def: dissimilarity} is also a customary notion of heterogeneity in distributed optimization~\cite{li2019communication,woodworth2020minibatch}, and we will use it to quantify the data heterogeneity in the nonconvex-nonconcave case.  The following theorem establishes the convergence rate.

\begin{theorem} \label{Thm: NCSC}
Let each client's objective function $f_i$ satisfy Assumptions \ref{assumption: strong concavity}-\ref{assumption: bounded grad}. Running Algorithm~\ref{algorithm: SGDA} under heterogeneous data setting,  choosing $\tau = \frac{T^{1/3}}{n^{1/3}}$ and  learning rates $\eta_x = \frac{n^{1/3}}{ LT^{2/3}} $ aand $\eta_y = \frac{2}{LT^{\frac{1}{2}}}$, if we choose sufficiently large $T$ such that
{\small \begin{align}
    T \geq  \max&\left\{40^{3/2}, \frac{160^3}{n^2},\right.\nonumber\\
    &\left. \left(\frac{16 n^{4/3}\kappa^4 + \sqrt{16 n^{4/3}\kappa^8 - 12 \beta n^{1/3}/L}}{2}\right)^3 \right\}, \nonumber
\end{align} }
holds, then we have
{\small \begin{align}
&   \frac{1}{T}\sum_{t=1}^T\mathbb{E}\left[\left\| \nabla \Phi(\bm{x}^{(t)}) \right\|^2\right]   \leq O\left( \frac{\kappa^4 L^2\sigma^2}{(nT)^{1/3}} + \frac{L^2\zeta_x}{T^{2/3}}+ \frac{L^2\zeta_y}{n^{2/3}T^{1/3}}  \right),\nonumber
\end{align}}
where $\kappa = L/\mu$, $\beta = L+\kappa L$. 
\end{theorem}
The proof of Theorem~\ref{Thm: NCSC} is deferred to Appendix~\ref{sec:ncsc}. We note that when we assume local data distributions are homogeneous, the above rate still holds but the terms $\zeta_x$ and $\zeta_y$ that correspond to heterogeneity  will disappear.
Theorem~\ref{Thm: NCSC} shows that local SGDA converges in the rate of  $O\left(\frac{1}{(nT)^{1/3}}\right)$ with $O\left(n^{\frac{1}{3}}T^{\frac{2}{3}}\right)$ communication rounds. Also, local SGDA enjoys linear speedup in the number of workers $n$. The most analogue work to ours in this setting is~\cite{lin2019gradient}, where they study the convergence of \textit{centralized}  SGDA (single machine) for noncovex-strongly-concave objectives, and achieve an $O(\frac{1}{\sqrt{T}})$ convergence rate. However, their algorithm requires that the mini-batch size of stochastic gradients to be very large, i.e., $O(\frac{1}{\epsilon^2})$ to reach an $\epsilon$-stationary point, therefore, requiring more computation budget per iteration. In our case, the batch size can be a constant, which avoids expensive large batch evaluations. We also note that as pointed out in~\cite{lin2019gradient}, due to the nonsymmetric nature of the nonconvex-(strongly)-concave problem, we need different step sizes for primal and dual variables. In fact, since objective is strongly-concave in dual variable, we can choose a larger dual step size as stated in Theorem~\ref{Thm: NCSC}.

\section{Local SGDA+}

\begin{algorithm2e}[t]
	\renewcommand{\algorithmicrequire}{\textbf{Input:}}
	\DontPrintSemicolon
    \caption{\texttt{Local SGDA+}}
	\label{algorithm: SGDA+}
 
    	\textbf{Input:} Synchronization gap $\tau$, Snapshot gap $S$, Number of iterations $T$,  Initial local models $\bm{x}^{(0)}_{i}$, $\bm{y}^{(0)}_{i}$  for $i\in [n]$.
	\\
 
	\textbf{parallel} \For{$i = 1,...,n$}{
	 
	 \For{$t = 0,...,T-1$}{
  
 	 {    ${\bm{x}}^{(t+1)}_i ={\bm{x}}^{(t)}_i - \eta_x \nabla_{x}      f_i\left({\bm{x}}^{(t)}_i,{\bm{y}}^{(t)}_i;\xi_i^t\right)$  \\
            ${\bm{y}}^{(t+1)}_i =  {\bm{y}}^{(t)}_i + \eta_y \nabla_{y} f_i\left(\Tilde{\bm{x}} ,{\bm{y}}^{(t)}_i;\xi_i^t\right)$\\   }
	 \If{$t+1$ divides $\tau$}{
	        all nodes send their local model ${\bm{x}}^{(t)}_i$ and ${\bm{y}}^{(t+1)}_i$ to server.\\
           $ \bm{x}^{(t+1)} = \frac{1}{n}\sum_{i=1}^n {\bm{x}}^{(t+1)}_i$;\\
 	        $ \bm{y}^{(t+1)} = \frac{1}{n}\sum_{i=1}^n {\bm{y}}^{(t+1)}_i$;\\
 	        send $ \bm{x}^{(t+1)}$, $ \bm{y}^{(t+1)}$ to all nodes to update their local models. \\
 	        each client initializes its local models: $\bm{x}_i^{(t+1)} = \bm{x}^{(t+1)}$ and $\bm{y}_i^{(t+1)} = \bm{y}^{(t+1)}$.}
 \If{$t+1$ divides $S$}{
	        all nodes send their local model ${\bm{x}}^{(t+1)}_i$ to server.\\
           take snapshot: $ \Tilde{\bm{x}}  = \frac{1}{n}\sum_{i=1}^n {\bm{x}}^{(t+1)}_i$;\\
 	        
 	        send $ \Tilde{\bm{x}}$  to all nodes.  }
	 
	 } 
	
  }

\end{algorithm2e} 
In this section, we proceed to an even harder seting where the objective is  nonconvex in primal variable $\bm{x}$ and nonconcave in dual parameter $\bm{y}$. Nonconvex-nonconcave minimax optimization is an active research area due to the rise of GANs~\cite{goodfellow2014generative}, and a few recent studies have proposed  efficient algorithms for optimizing nonconvex-nonconcave objectives~\cite{lin2018solving,jin2019local,wang2019solving,nouiehed2019solving}. However, these algorithms are all double loop: they require solving the maximization problem to get a $\epsilon$-accurate solution, and then go back to solve minimization problem. The drawbacks will be two-fold: first, they introduce a new hyperparameter $\epsilon$, which needs to be pre-tuned; second, the implementation will be more complicated, and is not straightforward to be extended to distributed setting. In this section, we propose a variant of local SGDA, dubbed as local SGDA+, aimed at solving nonconvex-nonconcave minimax problems in distributed setting with reduced communication overhead. 

\noindent\textbf{Our proposal: snapshot iterate and stale gradients.}~Before introducing our algorithm, let us first discuss the single machine setting to illustrate our main ideas. In the vanilla single loop (S)GDA, we query the gradient based on current iterate $(\bm{x}^{(t)}, \bm{y}^{(t)})$. It posts difficulty to prove the convergence since under nonconcavity assumption, we do not know how close $\bm{y}^{(t)}$ is  to $\bm{y}^*(\bm{x}^{(t)})$ (as elaborated in~\cite{lin2019gradient}, in nonconcave case, $\bm{y}^*(\cdot)$ is not even Lipschitz). As a result, the existing methods mainly follow a double loop schema: at outer loop, we update $\bm{x}^{(t-1)}$ using SGD or its variants  to get $\bm{x}^{(t)}$, and then, we fix $\bm{x}^{(t)}$, and run few steps of stochastic gradient ascent to solve inner maximization problem: $\max_{\bm{y}\in\mathbb{R}^{d_y}} F(\bm{x}^{(t)} ,\bm{y})$ to get an $\epsilon$-accurate approximation of $\bm{y}^*(\bm{x}^{(t)})$, where $\epsilon$ is the predetermined level of accuracy. This is a successful algorithm, but due the two weaknesses we  mentioned before, we prefer a single loop algorithm is distributed setting. In order to alleviate the need for the inner loop, we propose to update $\bm{y}$ with \textbf{stale gradients} evaluated on some past \textbf{snapshot iterate} $\Tilde{\bm{x}}$. To be more specific, each local worker will perform following update:
\begin{align*}
     &{\bm{x}}^{(t+1)}_i ={\bm{x}}^{(t)}_i - \eta_x \nabla_{x}      f_i\left({\bm{x}}^{(t)}_i,{\bm{y}}^{(t)}_i;\xi_i^t\right),  \\
    &{\bm{y}}^{(t+1)}_i = {\bm{y}}^{(t)}_i + \eta_y \nabla_{y} f_i\left(\Tilde{\bm{x}} ,{\bm{y}}^{(t)}_i;\xi_i^t\right).
\end{align*}
The update for primal ${\bm{x}}^{(t)}_i$ is identical to what we did in local SGDA, however, when we update the dual model ${\bm{y}}^{(t)}_i$, instead of evaluating gradient on $\bm{x}^{(t)}_i$, we query gradient evaluated on a snapshot iterate $\Tilde{\bm{x}}$, which  will be updated every $S$ iterations. This updating scheme can guarantee that we can optimize on $\bm{y}$ for fixed $\bm{x}$ but without actually \emph{locking} the update of $\bm{x}$. This algorithm will  no longer need the inner loop hyperparameter $\epsilon$ and it is easy to be implemented in a distributed fashion. The detailed steps of local SGDA+ are provided in Algorithm~\ref{algorithm: SGDA+}. We note that by choosing a small primal learning rate, $\Tilde{\bm{x}}$ will not drift far away from current iterate $\bm{x}^{(t)}_i$, and hence its convergence is guaranteed.

\subsection{Convergence of local SGDA+}\label{sec: NCNC}
 We now establish the convergence of local SGDA+ for a class of nonconvex-nonconcave function. We consider two function class: (i)  $F(\bm{x},\bm{y})$ is nonconvex in $\bm{x}$, and satisfies PL-condition in $\bm{y}$.  (ii)  $F(\bm{x},\bm{y})$ is nonconvex in $\bm{x}$, and one-point concave in $\bm{y}$. To do so, we make the following assumptions on the objective.
\begin{assumption}[Polyak-Łojasiewicz Condition] \label{assumption: pl}
$F(\bm{x},\bm{y})$ is said to satisfy Polyak-Łojasiewicz (PL) condition in $\bm{y}$ if $\forall \bm{x} \in \mathbb{R}^{d_x}$, the following holds: $\frac{1}{2} \left\| \nabla_y F(\bm{x}, \bm{y}) \right\|^2 \geq \mu \left(F(\bm{x}, \bm{y}^*(\bm{x})) - F(\bm{x}, \bm{y})\right), \forall \bm{y}\in \mathbb{R}^{d_y}.$
\end{assumption}
 
 \begin{assumption}[Lipschitz Continuity in $\bm{x}$] \label{assumption: lip}
$F(\bm{x},\bm{y})$ is said to be $G_x$-Lipschitz in $\bm{x}$ if the following holds: $\forall \bm{x},\bm{x}'\in \mathbb{R}^{d_x}$: $\left\| F(\bm{x},\bm{y}) -  F(\bm{x}',\bm{y})\right\|  \leq G_x\left\|  \bm{x}  -   \bm{x}' \right\|$.
\end{assumption}
 
The following theorem establishes the convergence rate of local SGDA+ on nonconvex-PL objectives.
  
\begin{theorem} [Nonconvex-PL]\label{Thm: NCNC}
Let objective function $F$ satisfies Assumption~\ref{assumption: pl} and local functions satisfy Assumptions~\ref{assumption: smoothness} and \ref{assumption: lip}. Also assume $F$ is $G_x$ Lipschitz in $\bm{x}$.  Running Algorithm~\ref{algorithm: SGDA+} under heterogeneous data setting, by choosing $\tau = T^{1/3}$, $S = T^{2/3}$, $\eta_x = \frac{n^{1/3}}{LT^{2/3}}$, $\eta_y = \frac{n^{1/3}}{LT^{1/2}}$, $\tau = \frac{T^{1/3}}{n^{2/3}}$, and $S = \frac{T^{1/3}}{n^{2/3}}$, if we set
\begin{align}
     T &\geq  \max \left\{(8\kappa^2)^6, \right. \nonumber \\ 
    &\quad \left. O\left( \frac{\beta n^{1/3}}{2L} + \sqrt{ \frac{8L(L+\beta)n^{1/3}}{\mu^2}+ \frac{4L^2 n^{2/3}}{\mu}}  \right)^{3/2} \right\} \nonumber,
\end{align}
then it holds that  
{\begin{align}
    &\frac{1}{T}\sum_{t=1}^T\mathbb{E}\left[\left\| \nabla \Phi(\bm{x}^{(t)})\right\|^2\right]\nonumber\\
    &\leq O\left(\frac{\beta \sigma^2}{(nT)^{1/3}}+\frac{\kappa^2 L^2 \zeta_y}{n^{2/3}T^{1/3}} + \frac{\kappa^2 L^2\zeta_x}{ n^{2/3}T} + \frac{\kappa^2 L^2 G^2_x}{ T}\right),\nonumber
\end{align}}
where $\kappa = L/\mu$, $\beta = L+\kappa L$. 
\end{theorem}
The proof of Theorem~\ref{Thm: NCNC} is deferred to Appendix~\ref{sec:ncnc}. Again, if  we assume local functions are homogeneous, this rate also holds but the  terms $\zeta_x$ and $\zeta_y$ will disappear. Here we obtain an $O\left( \frac{1}{(nT)^{1/3}}\right)$ rate with only $O(T^{2/3})$ communication rounds, as good as what we get in nonconvex-strongly-concave case. The most analogous work is~\cite{reisizadeh2020robust}, where they prove the convergence rate of vanilla local SGDA on nonconvex-PL game. Their work shows that, the vanilla local SGDA can still converge under nonconvex-PL condition. However, their analysis does not generalize to nonconvex-one-point-concave setting, but we develop the convergence theory of local SGDA+, as we will present in the next theorem. Another similar work is~\cite{nouiehed2019solving}, where they study the single machine algorithm in nonconvex-PL setting. They propose a double loop gradient descent ascent, and achieve and $\Tilde{O}\left(\frac{1}{T^{1/2}}\right)$ convergence rate under their convergence measure, which is recognized as the first analysis for nonconvex-PL game, to our best knowledge.

Now, we proceed to an even harder case: the objective is nonconvex in $\bm{x}$ and one-point concave in $\bm{y}$. One point convexity/concavity property has been shown to hold under the dynamic of SGD on optimizing neural networks ~\cite{li2017convergence,kleinberg2018alternative,zhou2019sgd}, which has been demonstrated  both  theoretically
and empirically. In addition, some works on minimax optimization also adapt similar assumption~\cite{mertikopoulos2018optimistic,liu2019decentralized,liu2019towards,iusem2017extragradient}.  Since our objective is no longer strongly-concave or PL in $\bm{y}$, then it will be difficult to analyze the dynamic of $\Phi$ directly, because $\Phi$ is not smooth any more. Instead, we study the \emph{Moreau envelope} of $\Phi$, in order to analyze the convergence, as suggested in several recent studies~\cite{davis2019stochastic,lin2019gradient,rafique2018non}.
\begin{definition}[Moreau Envelope] A function $\Phi_{p} (\bm{x})$ is the $p$-Moreau envelope of a function $\Phi$ if $    \Phi_{p} (\bm{x}) := \min_{\bm{x}'\in \mathbb{R}^{d_x}} \left\{ \Phi  (\bm{x}') + \frac{1}{2p}\|\bm{x}'-\bm{x}\|^2\right\}$.
\end{definition}
We will use  $1/2L$-Moreau envelope of $\Phi$, following the setting in~\cite{lin2019gradient,rafique2018non}, and  state the convergence rate in terms of $\|\nabla \Phi_{1/2L} (\bm{x})\|$.

\begin{assumption}[One Point Concavity] \label{assumption: oc}
$f_i(\bm{x},\bm{y})$ is said to satisfy one point concavity in $\bm{y}$ if we fix $\bm{x}$, then $\forall \bm{y} \in \mathbb{R}^{d_y}$, the following holds: $ \langle \nabla_y f_i(\bm{x},\bm{y}), \bm{y} - \bm{y}^*(\bm{x}) \rangle \leq f_i(\bm{x},\bm{y})- f_i(\bm{x},\bm{y}^*(\bm{x}))$.
\end{assumption}

\begin{theorem} [Nonconvex-One-Point-Concave]\label{Thm: NCOC}
Let local functions satisfy Assumptions~\ref{assumption: smoothness}, \ref{assumption: lip} and \ref{assumption: oc}, and $\|\bm{y}^{(t)}\|^2 \leq \frac{D}{2}$, $\|\bm{y}^{*}(\Tilde{\bm{x}})\|^2 \leq \frac{D}{2}$ for all $t$ and $\Tilde{\bm{x}}$ during iterating. Also assume $F$ is $G_x$ Lipschitz in $\bm{x}$.   Running Algorithm~\ref{algorithm: SGDA+} under heterogeneous data setting, by choosing $\eta_x = \frac{1}{LT^{\frac{5}{6}}}$, $\eta_y = \frac{1}{4LT^{\frac{1}{2}}}$, $\tau = T^{\frac{1}{3}}/n^{\frac{1}{6}}$, $S = T^{\frac{2}{3}}$, it holds that:
 \begin{align}
    &\frac{1}{T}\sum_{t=1}^T \mathbb{E}\left[\left\|\nabla \Phi_{1/2L} (\bm{x}^{(t)})\right\|^2\right] 
    \leq  O\left(\frac{L\sigma^2 }{T^{1/6}}\right)\nonumber\\ 
    & + O\left(  \frac{L^2 \sigma^2+LG_x^2}{(nT)^{1/3}} + \frac{L^2\zeta_x}{n^{1/3}T}   + \frac{L^2\zeta_y}{(nT)^{1/3}}\right)  +O\left(\frac{D}{T^{1/6}}\right)     \nonumber.
\end{align} 
\end{theorem}
The proof of Theorem~\ref{Thm: NCOC} is deferred to Appendix~\ref{sec:ncoc}. Local SGDA+ is guaranteed to find the first order stationary point of $\Phi_{1/2L}(\bm{x})$ at the rate of $O(\frac{1}{T^{1/6}})$ with $T^{2/3}$ communication rounds. Again, if  we assume local functions are homogeneous, this rate also holds but the  terms $\zeta_x$ and $\zeta_y$ will disappear. The most similar work to ours is~\cite{lin2019gradient}, where they analyze the single machine SGDA on nonconvex-concave setting, and established a rate of $O(\frac{1}{T^{1/4}})$. In contrast,  we consider a more difficult concave setting, and their analysis technique does not apply here directly.   
 
 \begin{figure*}[t!]
	\centering
	\subfigure[$\mathsf{Synthetic}\left(0.0\right)$]{
		\centering
		\includegraphics[width=0.31\textwidth]{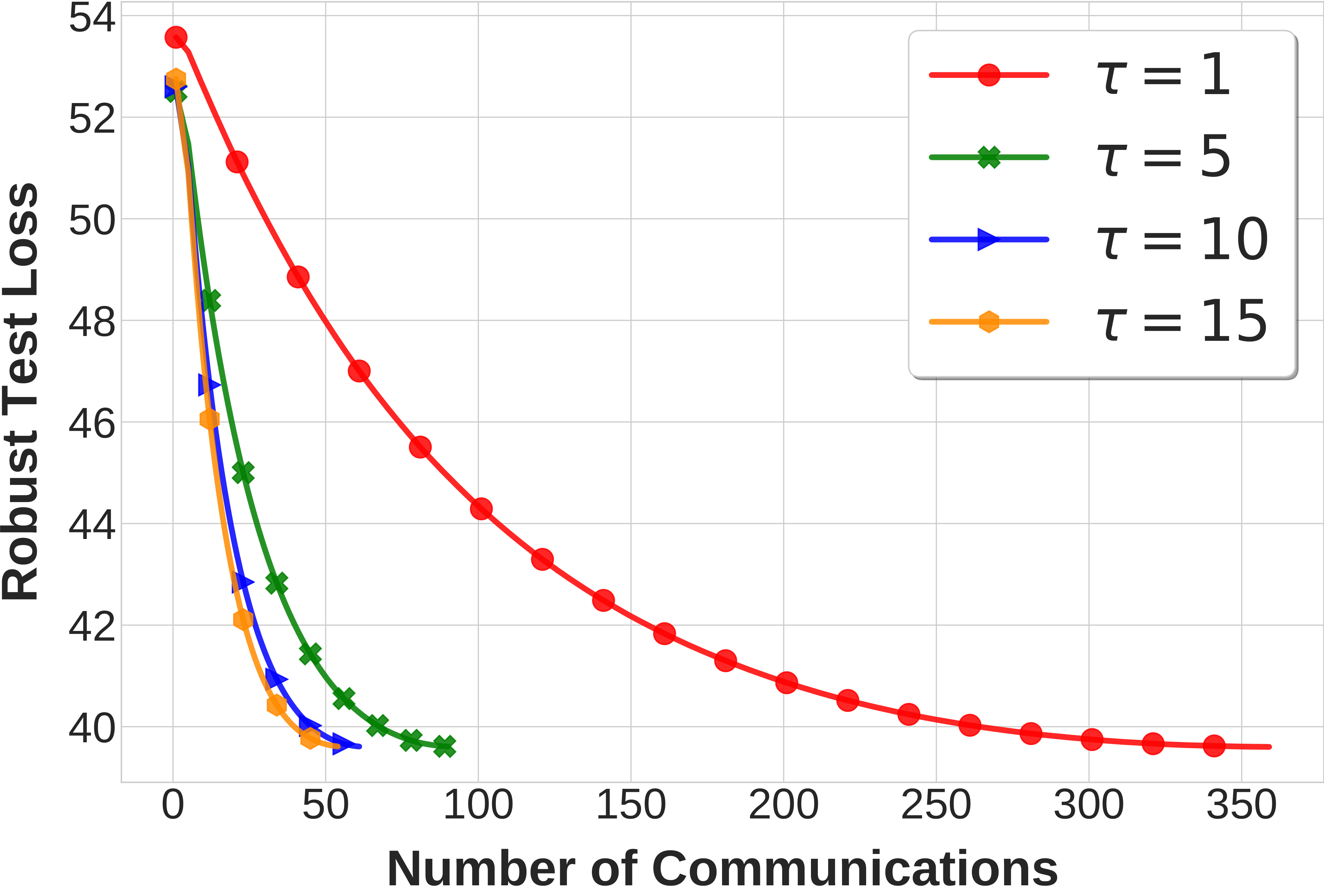}
		\label{fig:loss_synthetic0.0-0.0}
		}\hfill
	\subfigure[$\mathsf{Synthetic}\left(0.25\right)$]{
		\centering
		\includegraphics[width=0.31\textwidth]{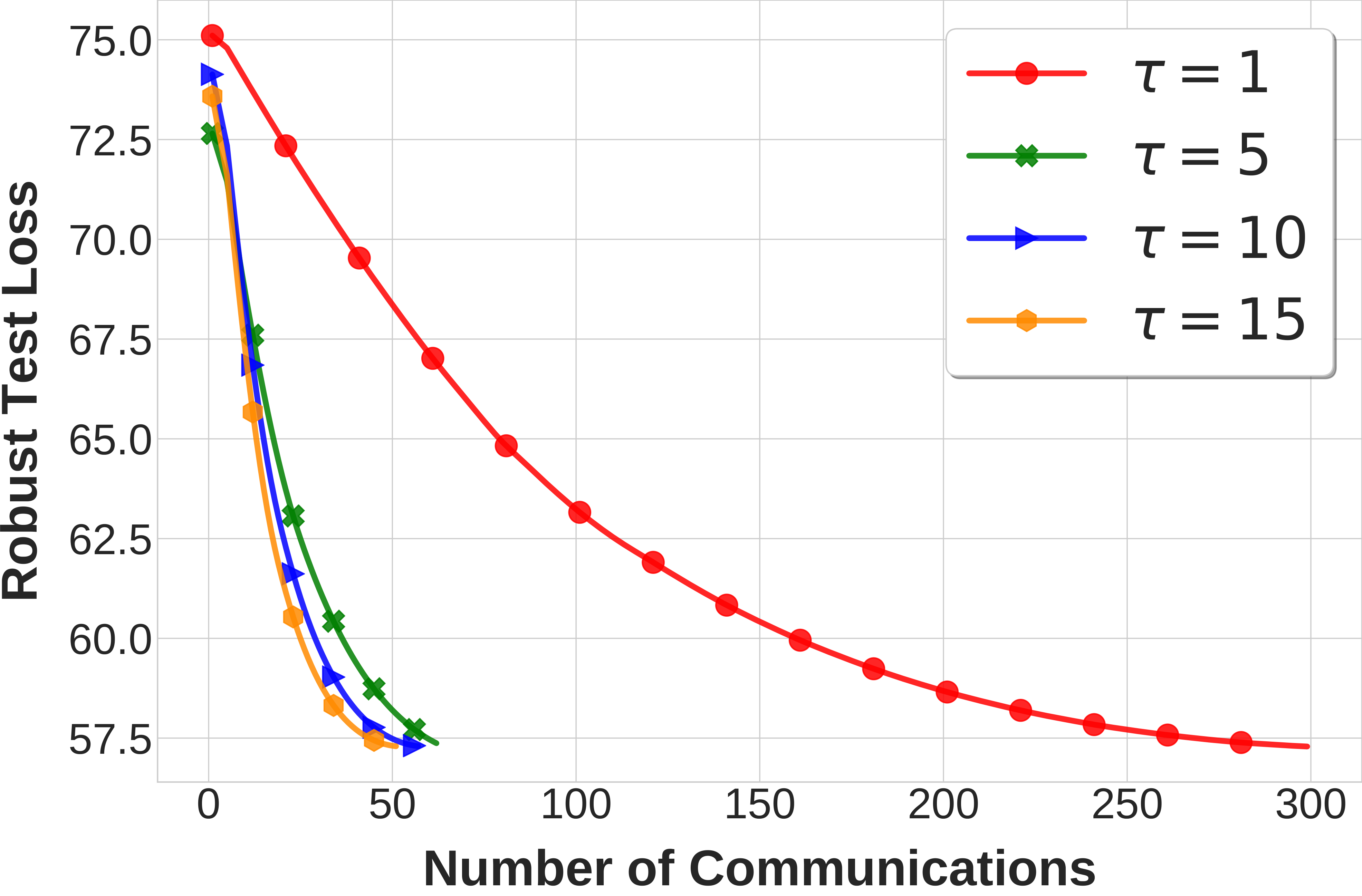}
		\label{fig:loss_synthetic0.0-0.25}
		}
		\subfigure[$\mathsf{Synthetic}\left(0.5\right)$]{
			\centering 
			\includegraphics[width=0.31\textwidth]{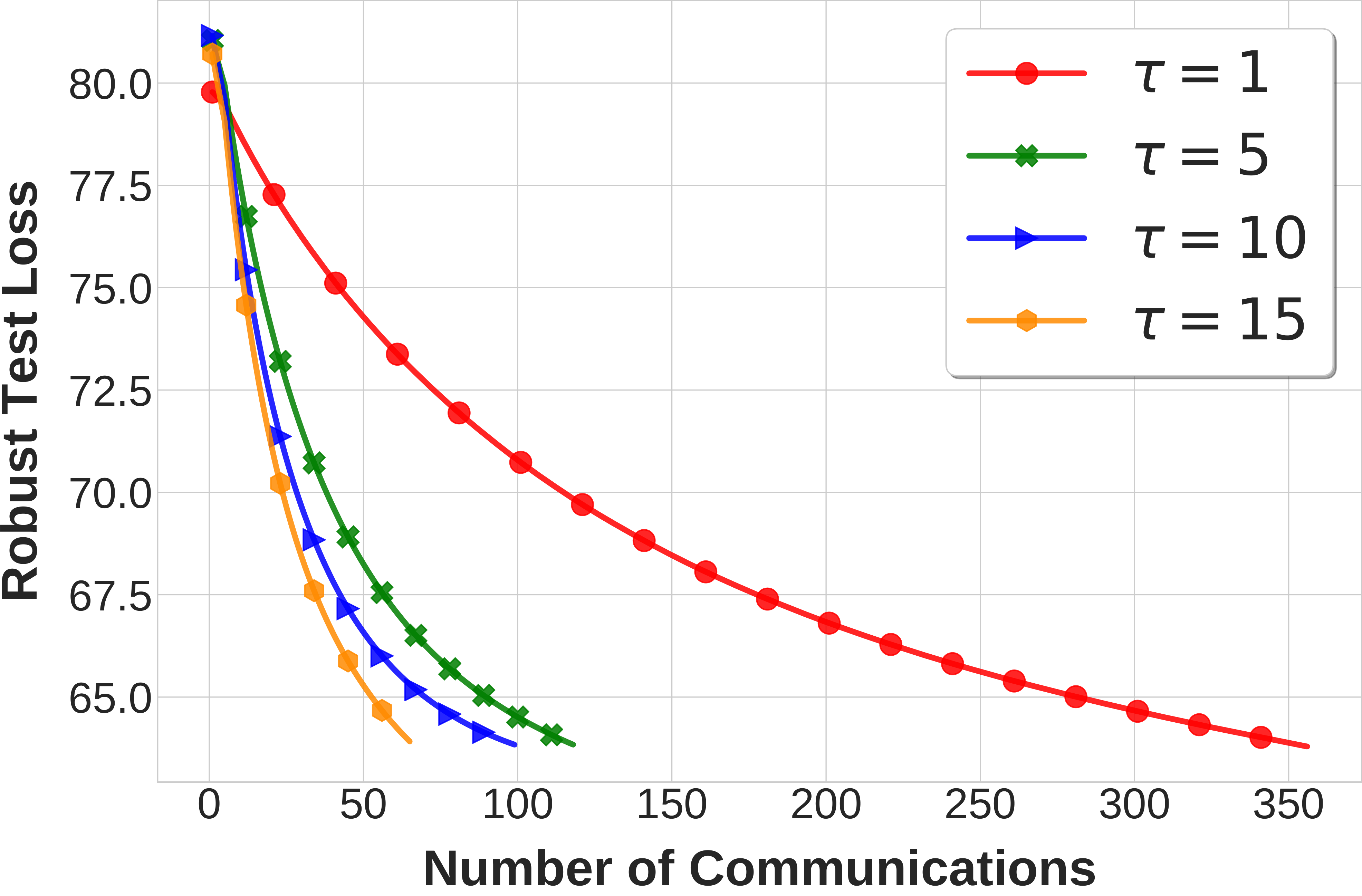}
			\label{fig:acc_synthetic0.0-0.5}
			}
	\caption[]{Linear regression with synthetic datasets using local SGDA ($\tau > 1$) comparing to SGDA ($\tau=1$). Local SGDA can achieve the same robust loss with fewer number of communication rounds than SGDA.}
	\label{fig:synthetic}
\end{figure*} 

 \section{Experiments}
In this section, we empirically examine the convergence of the proposed algorithms local SGDA and local SGDA+. We use two datasets, MNIST and a synthetic dataset and develop our code using \texttt{distributed} API of \texttt{PyTorch}. For the Algorithm~\ref{algorithm: SGDA}, to have a strongly convex-strongly concave loss function, we consider the robust linear regression problem, and for the Algorithm~\ref{algorithm: SGDA+}, to construct a nonconvex-nonconcave problem, we consider the robust neural network training, and use a $2$-layer MLP model with a cross entropy loss function. First, we explain the generation of non-iid datasets and then turn into the experimental results.

\paragraph{Datasets.}To generate a synthetic non-iid dataset, we follow the steps from~\cite{li2018federated}. In here, we only use the parameter to control the divergence between local datasets, while the true models for data generation of each node is coming from the same distribution.  Hence, for each node we generate a weight matrix $\bm{W}_i \in \mathbb{R}^{m \times 1}$ and a bias $\bm{b}\in \mathbb{R}^c$, where the output for the $i$th client is $y_i = \bm{W}_i^\top \bm{x}_i + b$. The model is generated based on a Gaussian distribution $\bm{W}_i \sim \mathcal{N}\left(0,1\right)$ and $\bm{b}_i \sim \mathcal{N}\left(0,1\right)$. The input data $\bm{x}_i \in \mathbb{R}^{m}$ has $m$ features and is drown from a Gaussian distribution $\bm{x}_i \sim \mathcal{N}\left(\bm{\mu}_i,\bm{\Sigma}\right)$, where $\bm{\mu}_i \sim \mathcal{N}\left(M_i,1\right)$ and $M_i \sim \mathcal{N}\left(0,\alpha\right)$. Also the variance $\bm{\Sigma}$ is a diagonal matrix with value of $\bm{\Sigma}_{k,k} = k^{-1.2}$. In this process, by changing $\alpha$ we can control the divergence between local input data of different nodes. We create $3$ different datasets by changing this parameter for the regression task, namely, $\mathsf{Synthetic}\left(0.0\right)$, $\mathsf{Synthetic}\left(0.25\right)$, and $\mathsf{Synthetic}\left(0.5\right)$.
For the MNIST dataset and for the classification task, we follow the same procedure in~\cite{mcmahan2017communication}, where we allocate data from only $2$ classes per node. This way, the data is distributed heterogeneously among nodes.
\begin{figure}[t!]
	\centering
	\subfigure{
		\centering
		\includegraphics[width=0.3\textwidth]{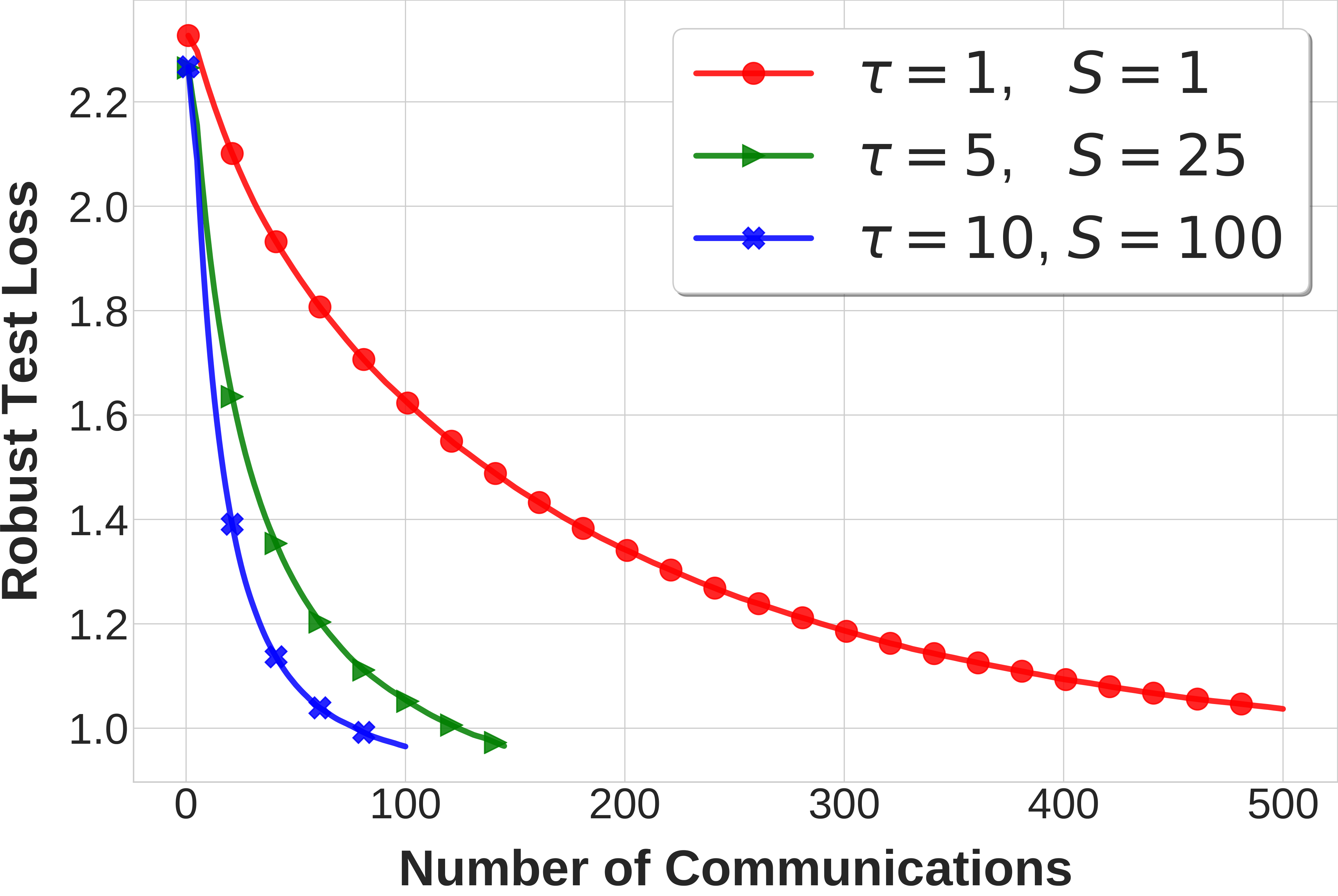}
		\label{fig:loss_mnist_global1}
		}
		\hspace{0.0cm}
		\subfigure{
			\centering 
			\includegraphics[width=0.3\textwidth]{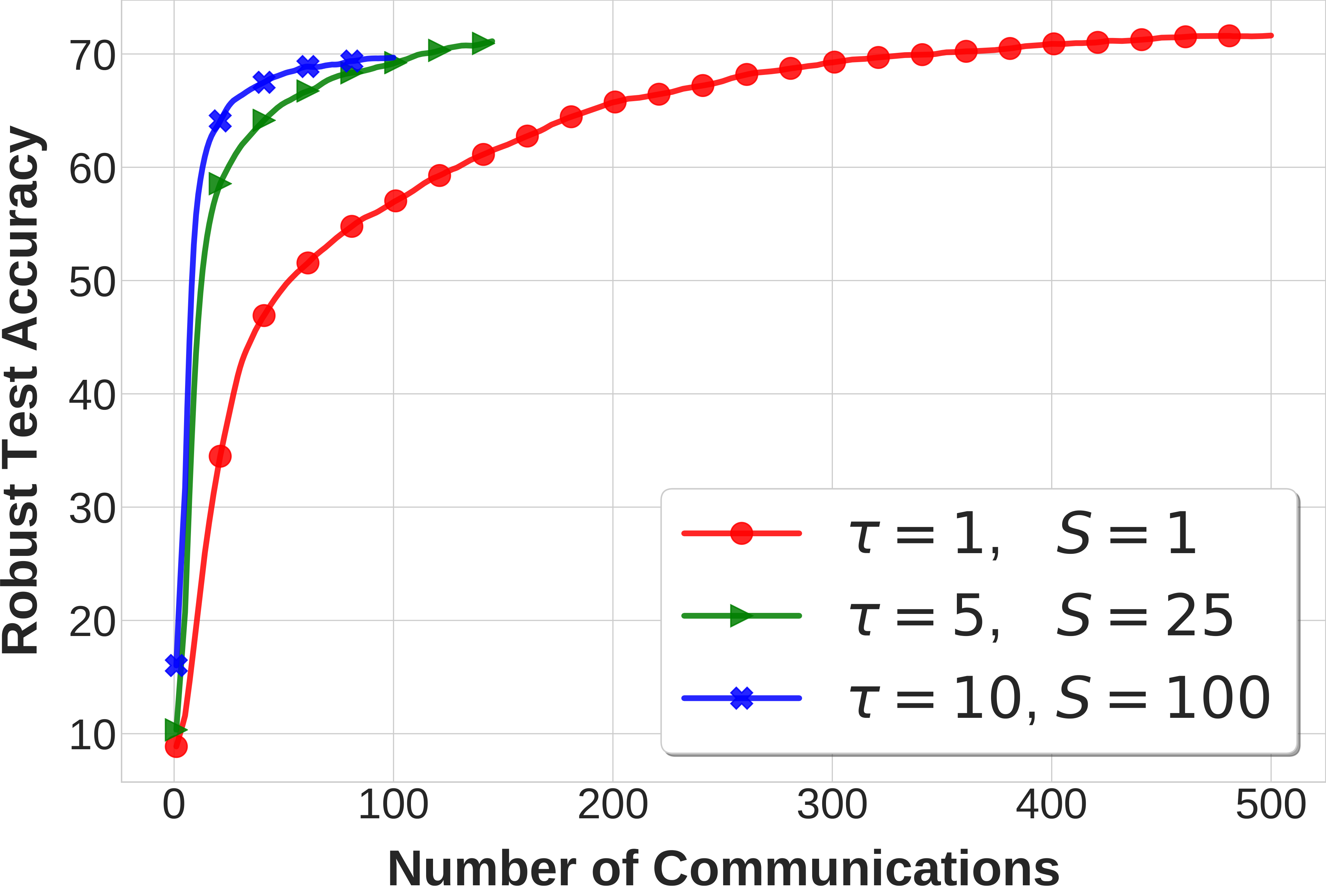}
			\label{fig:acc_mnist_global1}
			}
	
	\caption[]{Comparing local SGDA+ with normal SGDA ($\tau=1$, $S=1$) on training a $2$-layer MLP on heterogeneous MNIST dataset over $100$ nodes. Local SGDA+ acn converge to the same accuracy as SGDA with fewer rounds of communication between nodes and the server.}
	\label{fig:mnist_acc_local}
\end{figure} 
\paragraph{Robust Linear Regression.}In this experiments the model and loss function is defined as
\begin{align*}
    \min_{\bm{w}} \max_{\|\boldsymbol{\delta}\|^2\leq 1} \frac{1}{n}\sum_{i=1}^n(\bm{w}^{\top} (\bm{x}_i+\boldsymbol{\delta}) - y_i)^2 + \frac{1}{2}\|\bm{w}\|^2,
\end{align*}
For the convergence measure, we can use the robust loss. Given a model $\hat{\bm{w}}$, its robust loss is defined as
$$      \ell(\hat{\bm{w}}) = \max_{\|\boldsymbol{\delta}\|^2\leq 1} \frac{1}{n}\sum_{i=1}^n(\hat{\bm{w}}^{\top} (\bm{x}_i+\boldsymbol{\delta}) - y_i)^2 + \frac{1}{2}\|\hat{\bm{w}}\|^2,
$$ so each time to evaluate  a node's robust loss, we have to solve above maximization problem. One way to do it is to run few steps of gradient ascent to get a estimated $\hat{\delta}$.

In the first set of experiments,  we run the training procedure proposed in Algorithm~\ref{algorithm: SGDA} on synthetic datasets that introduced before. We set the input dimension to $60$ and each node has between $400$ to $500$ samples. We generate data for $100$ nodes, and drawn $20\%$ of each node's data for the test dataset to make it the average distribution among all nodes. We use the same learning rates for both dual and primal variables, and use a decaying mechanism to decrease it by $5\%$ every iteration. The initial learning rate for all the experiments is set to $0.001$. The results of this experiment is depicted in Figure~\ref{fig:synthetic}, where we compare the local SGDA ($\tau \in \{5,10,15\}$) with normal SGDA ($\tau=1$). It is clear that to achieve certain level of the robust loss, local SGDA needs significantly fewer number of communication rounds, compared to vanilla SGDA, hence it achieves communication efficiency.

\paragraph{Robust Neural Network Training.}Similar to the setting of Robust Linear Regression in~\cite{nouiehed2019solving}, here we just replace the model with a DNN  and optimize  \begin{align}
    \min_{\bm{W}} \max_{\|\boldsymbol{\delta}\|^2\leq 1} \frac{1}{n}\sum_{i=1}^n\ell(h_{\bm{W}}(\bm{x}_i+\boldsymbol{\delta}),y_i)\nonumber.
\end{align}
For this experiment, to evaluate  Algorithm~\ref{algorithm: SGDA+}, we use a $2$-layer MLP, each with $200$ neurons followed by ReLU activation  and a cross entropy loss function at the end. We divide the MNIST dataset among $100$ nodes, each with only having access to $2$ classes to introduce heterogeneity among local data shards. The test dataset is a pool of all classes, hence, it is the average dataset over all nodes. We use the same decaying learning rate scheme as the linear regression, where the initial learning rate is set to $0.01$. In this experiment, we set the snapshot gap $S=\tau^2$, as suggested in Theorem~\ref{Thm: NCNC}. The convergence measure is the robust accuracy, and we compute it similarly to robust loss as in robust linear regression. The results of these experiments are shown in Figure~\ref{fig:mnist_acc_local}, where compared to normal SGDA ($\tau=1$, $S=1$), the proposed local SGDA+ can converge faster in terms of number of communications. 

\section{Conclusions and Path Forward}
In this paper we proposed  a communication efficient  distributed method to solve minimax optimization problems and  establish  its convergence rate for strongly-convex-strongly-concave and nonconvex-strongly-concave  objectives in both homogeneous and heterogeneous data distribution settings. We also proposed a single loop variant of proposed algorithm to address nonconvex-noncancave problems that arises in learning  GANs. The present work is the first to study local SGD method in minimax setting  and leaves many interesting directions as future work. We believe some of the obtained rates can be tightened. Investigating the  achievable rates via local methods  in minimax setting also remains open. Another future work will be the exploration of faster algorithm  to match the known lower bound of first order minimax algorithm obtained in~\cite{ouyang2019lower}.

\section*{Acknowledgement}
We would like to thank Mohammad Mahdi Kamani for his help on conducting the experiments. This work has been done using the Extreme Science and Engineering Discovery Environment (XSEDE) resources, which is supported by National Science Foundation under grant number
ASC200045.
\bibliographystyle{plain}
\bibliography{references}  
\newpage
\appendix
\onecolumn
\section*{Appendix}
Here we present the omitted proofs of convergence rates. In Section~\ref{sec:proof_scsc} we give the proof of convergence  in strongly-convex-strongly-concave setting. Section~\ref{sec:ncsc} includes the proof for nonconvex-strongly-concave functions, and in Section~\ref{sec:ncnc} we present proof of local SGDA+ for nonconvex-PL objectives. Finally, in Section~\ref{sec:ncoc} we provide the proof of local SGDA+ on nonconvex-one-point-concave setting.

\section{Strongly-Convex-Strongly-Concave Setting}\label{sec:proof_scsc}

\subsection{Overview of proof techniques}
Before we dive into the proof we first sketch the proof of convergence of local SGDA  under strongly-convex-strongly-concave setting. We define the following notions to denote the (virtual) average primal and dual solution at $t$th iteration:
\begin{align}
 &\bm{x}^{(t)} = \frac{1}{n}\sum_{i=1}^n \bm{x}^{(t)}_i, \quad \bm{y}^{(t)} = \frac{1}{n}\sum_{i=1}^n \bm{y}^{(t)}_i,\nonumber
\end{align}

and the deviation between local primal and dual solutions and their corresponding  averages:
\begin{align}
 &\delta_{\bm{x}}^{(t)} = \frac{1}{n}\sum_{i=1}^n\left\|\bm{x}_i^{(t)}-\bm{x}^{(t)}\right\|^2, \quad \delta_{\bm{y}}^{(t)} = \frac{1}{n}\sum_{i=1}^n\left\|\bm{y}_i^{(t)}-\bm{y}^{(t)}\right\|^2\nonumber.
\end{align}

\paragraph{Homogeneous setting} In homogeneous setting, we first study the behavior of local SGDA for one iteration. With the help of strong convexity, concavity and smoothness we can show that:
\begin{equation*}
 \begin{aligned}
\mathbb{E}\left[ \left\|\bm{x}^{(t+1)} - \bm{x}^{*}\right\|^2 + \left\|\bm{y}^{(t+1)} - \bm{y}^{*}\right\|^2\right] & \leq \left(1-\frac{1}{2}\mu\eta\right)\left(\mathbb{E}\left[\left\|\bm{x}^{(t)} - \bm{x}^{*}\right\|^2 + \left\|\bm{y}^{(t)} - \bm{y}^{*}\right\|^2\right]\right) \nonumber \\
& \quad -2\eta \mathbb{E}\left(F(\bm{x}^{(t)},\bm{y}^*) - F(\bm{x}^{*},\bm{y}^{(t)})\right) \nonumber \\
& \quad +\frac{2\eta^2\sigma^2}{n}+ \frac{16\eta_tL^2}{\mu}\mathbb{E}\left(\delta_{\bm{x}}^{(t)}+\delta_{\bm{y}}^{(t)}\right) + 8\eta^2L^2  \mathbb{E}\left(\delta_{\bm{x}}^{(t)}+\delta_{\bm{y}}^{(t)}\right).
 \end{aligned}
\end{equation*}

Then, to bound $\delta_{\bm{x}}^{(t)}+\delta_{\bm{y}}^{(t)}$, with the help of strong convexity and smoothness, we can indeed show that it decreases in the order of $O(\tau(1+(L-\mu)\eta)^{2\tau}\eta^2\sigma^2 )$. By properly choosing $\tau$ and $\eta$, we recover the rate  $O(\tau \eta^2\sigma^2 )$ as desired.

\paragraph{Heterogeneous setting}Similarly to homogeneous setting, we first do the one iteration analysis
\begin{equation*}
 \begin{aligned}
 \mathbb{E}\left[ \left\|\bm{x}^{(t+1)} - \bm{x}^{*}\right\|^2 + \left\|\bm{y}^{(t+1)} - \bm{y}^{*}\right\|^2\right] & \leq \left(1-\frac{1}{2}\mu\eta_t \right)\left(\mathbb{E}\left[\left\|\bm{x}^{(t)} - \bm{x}^{*}\right\|^2 + \left\|\bm{y}^{(t)} - \bm{y}^{*}\right\|^2\right]\right) \nonumber \\
& \quad -2\eta_t \mathbb{E}\left(F(\bm{x}^{(t)},\bm{y}^*) - F(\bm{x}^{*},\bm{y}^{(t)})\right) \nonumber \\
& \quad +\frac{2\eta_t^2\sigma^2}{n}+ \frac{16\eta_tL^2}{\mu}\mathbb{E}\left(\delta_{\bm{x}}^{(t)}+\delta_{\bm{y}}^{(t)}\right) + 8\eta_t^2L^2 \mathbb{E}\left(\delta_{\bm{x}}^{(t)}+\delta_{\bm{y}}^{(t)}\right).
 \end{aligned}
\end{equation*}
Next we need to bound deviation $\delta_{\bm{x}}^{(t)}+\delta_{\bm{y}}^{(t)}$, which is also our main technical contribution in this section. We consider the interval of $\tau$ steps, if we choose step size to be small enough and properly choose quadratic weights $w_t = (t+a)^2$, to make sure the deviation changes slowly, we can finally prove the following statement:
\begin{equation}
 \begin{aligned}
 \sum_{t=s\tau}^{(s+1)\tau}w_t \mathbb{E}\left[\delta_{\bm{x}}^{(t)}+\delta_{\bm{y}}^{(t)} \right] &\leq       \frac{\mu}{128L^2} \sum_{j=s\tau}^{(s+1)\tau}\mu \eta_j\frac{w_{j}}{\eta_{j}}\mathbb{E}\left[\left\| \bm{x} ^{(j)}  -  \bm{x}^{*}   \right\|^2   + \left\| \bm{y}^{(j)} -\bm{y}^{*}   \right\|^2\right] \nonumber \\
& \quad   +  64 \tau^2   \sum_{j=s\tau}^{(s+1)\tau}w_j\eta_j^2\left(\Delta_{x}+\Delta_{y}\right)   + 32  \tau^2  \sum_{j=s\tau}^{(s+1)\tau}w_j\eta_j^2\sigma^2,\nonumber
\end{aligned}
\end{equation}
where we related the deviation to the gap between current iterates and saddle points, and heterogeneity at global optimum.

\subsection{Proof in homogeneous setting}
In this section we are going to present the proof in homogeneous case. Let us introduce some technical lemmas first which will help our proof.
\subsubsection{Proof of technical lemmas}
The following lemma performs one iteration analysis of local SGDA, on strongly convex function.
\begin{lemma} \label{Lemma_IID: scsc one iteration}
For local-SGDA, under Theorem~\ref{Thm: IID_SCSC}'s assumptions, the following relation holds true:
\begin{equation*}
 \begin{aligned}
\mathbb{E}\left[ \left\|\bm{x}^{(t+1)} - \bm{x}^{*}\right\|^2\right] + \mathbb{E}\left[\left\|\bm{y}^{(t+1)} - \bm{y}^{*}\right\|^2 \right]& \leq \left(1-\frac{1}{2}\mu\eta\right)\left[\mathbb{E}\left[\left\|\bm{x}^{(t)} - \bm{x}^{*}\right\|^2 \right]+ \mathbb{E}\left[\left\|\bm{y}^{(t)} - \bm{y}^{*}\right\|^2\right]\right] \nonumber \\
& \quad -2\eta \left(\mathbb{E}\left[F(\bm{x}^{(t)},\bm{y}^*) \right]- \mathbb{E}\left[F(\bm{x}^{*},\bm{y}^{(t)})\right]\right) \nonumber \\
& \quad +\frac{2\eta^2\sigma^2}{n}+ \frac{16\eta_tL^2}{\mu} \mathbb{E}\left[\delta_{\bm{x}}^{(t)}+\delta_{\bm{y}}^{(t)}\right]  + 8\eta^2L^2  \mathbb{E}\left[\delta_{\bm{x}}^{(t)}+\delta_{\bm{y}}^{(t)}\right]\nonumber,
 \end{aligned}
\end{equation*}
where $\delta_{\bm{x}}^{(t)} = \frac{1}{n}\sum_{i=1}^n\left\|\bm{x}_i^{(t)}-\bm{x}^{(t)}\right\|^2, \quad \delta_{\bm{y}}^{(t)} = \frac{1}{n}\sum_{i=1}^n\left\|\bm{y}_i^{(t)}-\bm{y}^{(t)}\right\|^2$.
\begin{proof}
According to updating rule and strong convexity we have:
\begin{align}
\mathbb{E}\left[\left\|\bm{x}^{(t+1)} - \bm{x}^{*}\right\|^2\right] &= \mathbb{E}\left[\left\|\bm{x}^{(t )} - \eta \frac{1}{n}\sum_{i=1}^n \nabla_x F(\bm{x}_i^{(t)}, \bm{y}_i^{(t)}; \xi_i^{(t)}) - \bm{x}^{*}\right\|^2\right] \nonumber\\
& \leq \mathbb{E}\left[\left\|\bm{x}^{(t)} - \bm{x}^{*}\right\|^2\right] -2\eta\mathbb{E}\left\langle   \frac{1}{n}\sum_{i=1}^n \nabla_x F(\bm{x}_i^{(t)}, \bm{y}_i^{(t)} ),  \bm{x}^{(t)} - \bm{x}^{*} \right\rangle \nonumber\\
&\quad + \frac{\eta^2\sigma^2}{n} + \eta^2\mathbb{E}\left[\left\|\frac{1}{n}\sum_{i=1}^n \nabla_x F(\bm{x}_i^{(t)}, \bm{y}_i^{(t)})\right\|^2\right]\nonumber\\
&\leq \mathbb{E}\left[\left\|\bm{x}^{(t)} - \bm{x}^{*}\right\|^2\right] -2\eta\left\langle     \nabla_x F(\bm{x}^{(t)}, \bm{y}^{(t)} ),  \bm{x}^{(t)} - \bm{x}^{*} \right\rangle \nonumber\\
&\quad -2\eta \mathbb{E}\left\langle  \frac{1}{n}\sum_{i=1}^n \nabla_x F(\bm{x}_i^{(t)},\bm{y}_i^{(t)}) - \nabla_x F(\bm{x}^{(t)},\bm{y}^{(t)}) , \bm{x}^{(t)} - \bm{x}^{*}   \right\rangle \nonumber\\
&\quad + \frac{\eta^2\sigma^2}{n} + \eta^2\mathbb{E}\left[\left\|\frac{1}{n}\sum_{i=1}^n \nabla_x F(\bm{x}_i^{(t)}, \bm{y}_i^{(t)})\right\|^2\right] \nonumber\\
&\leq \left(1- \mu\eta\right)\mathbb{E}\left[\left\|\bm{x}^{(t)} - \bm{x}^{*}\right\|^2\right] -2\eta\mathbb{E}\left( F(\bm{x}^{(t)},\bm{y}^{(t)}) -  F(\bm{x}^{*}, \bm{y}^{(t)})\right) \nonumber\\
&\quad + \eta \left(\frac{4}{\mu}\mathbb{E}\left\|\frac{1}{n}\sum_{i=1}^n \nabla_x F(\bm{x}_i^{(t)},\bm{y}_i^{(t)}) - \nabla_xF(\bm{x}^{(t)},\bm{y}^{(t)}) \right\|^2 +\frac{\mu}{4}\mathbb{E}\left\| \bm{x}^{(t)} - \bm{x}^{*} \right\|^2 \right) \nonumber \\
&\quad + \frac{\eta^2\sigma^2}{n} + \eta^2\mathbb{E}\left[\left\|\frac{1}{n}\sum_{i=1}^n \nabla_x F(\bm{x}_i^{(t)}, \bm{y}_i^{(t)})\right\|^2\right].\nonumber
\end{align}
We now proceed to bound terms $\left\|\frac{1}{n}\sum_{i=1}^n \nabla_x F(\bm{x}_i^{(t)},\bm{y}_i^{(t)}) - \nabla_xF(\bm{x}^{(t)},\bm{y}^{(t)}) \right\|^2$ and $\left\|\frac{1}{n}\sum_{i=1}^n \nabla_x F(\bm{x}_i^{(t)}, \bm{y}_i^{(t)})\right\|^2$.

By applying Jensen's inequality on $\left\|\frac{1}{n}\sum_{i=1}^n \nabla_x F(\bm{x}_i^{(t)},\bm{y}_i^{(t)}) - \nabla_xF(\bm{x}^{(t)},\bm{y}^{(t)}) \right\|^2$ we have:
\begin{align}
&\left\|\frac{1}{n}\sum_{i=1}^n \nabla_x F(\bm{x}_i^{(t)},\bm{y}_i^{(t)}) - \nabla_xF(\bm{x}^{(t)},\bm{y}^{(t)}) \right\|^2 \nonumber \\&=  \frac{1}{n}\sum_{i=1}^n\left\| \nabla_x F(\bm{x}_i^{(t)},\bm{y}_i^{(t)}) - \nabla_x F(\bm{x}^{(t)},\bm{y}^{(t)}) \right\|^2 \nonumber\\
&\leq \frac{1}{n}\sum_{i=1}^n\left(2\left\| \nabla_x F(\bm{x}_i^{(t)},\bm{y}_i^{(t)}) - \nabla_x F(\bm{x}^{(t)},\bm{y}_i^{(t)}) \right\|^2 + 2\left\| \nabla_x F(\bm{x}^{(t)},\bm{y}_i^{(t)}) - \nabla_x F(\bm{x}^{(t)},\bm{y}^{(t)}) \right\|^2\right)\nonumber\\
&\leq \frac{1}{n}\sum_{i=1}^n\left(2L^2\left\|  \bm{x}_i^{(t)}  - \bm{x}^{(t)}  \right\|^2 + 2L^2\left\|  \bm{y}_i^{(t)} - \bm{y}^{(t)}  \right\|^2\right) \nonumber\\
& \leq 2L^2 (\delta_{x}^{(t)}+\delta_{y}^{(t)}),\nonumber
\end{align}
where we use the smoothness in the second last inequality.

Then we switch to bound $\left\|\frac{1}{n}\sum_{i=1}^n \nabla_x F(\bm{x}_i^{(t)}, \bm{y}_i^{(t)})\right\|^2$ as follows:
{\begin{align}
\left\|\frac{1}{n}\sum_{i=1}^n \nabla_x F(\bm{x}_i^{(t)}, \bm{y}_i^{(t)})\right\|^2 &= \frac{1}{n}\sum_{i=1}^n\left\| \nabla_x F(\bm{x}_i^{(t)}, \bm{y}_i^{(t)})\right\|^2 \nonumber\\ &= \frac{1}{n}\sum_{i=1}^n\left\| \nabla_x F(\bm{x}_i^{(t)}, \bm{y}_i^{(t)}) -\nabla_x F(\bm{x}^{*}, \bm{y}^{*}) \right\|^2 \nonumber\\
&\leq \frac{1}{n}\sum_{i=1}^n 2\left(\left\| \nabla_x F(\bm{x}_i^{(t)}, \bm{y}_i^{(t)}) -\nabla_x F(\bm{x}^{(t)}, \bm{y}^{(t)}) \right\|^2 + \left\| \nabla_x F(\bm{x}^{(t)}, \bm{y}^{(t)}) -\nabla_x F(\bm{x}^{*}, \bm{y}^{*}) \right\|^2\right)\nonumber\\
&\leq L^2\frac{1}{n}\sum_{i=1}^n 4 \left(\left\| \bm{x}_i^{(t)} - \bm{x}^{(t)}\right\|^2 + \left\| \bm{x}^{(t)} - \bm{x}^{*}\right\|^2 + \left\|\bm{y}_i^{(t)} - \bm{y}^{(t)}\right\|^2+ \left\| \bm{y}^{(t)} - \bm{y}^{*}\right\|^2 \right).\nonumber
\end{align}}
where in the second equality we used the fact that $\nabla_x F(\bm{x}^{*}, \bm{y}^{*}) = \boldsymbol{0}$.

Putting these pieces together yields:
\begin{align}
\mathbb{E}\left[\left\|\bm{x}^{(t+1)} - \bm{x}^{*}\right\|^2\right]   &\leq \left(1-\frac{3}{4} \mu\eta\right)\mathbb{E}\left[\left\|\bm{x}^{(t)} - \bm{x}^{*}\right\|^2\right] -2\eta\mathbb{E}\left( F(\bm{x}^{(t)},\bm{y}^{(t)}) -  F(\bm{x}^{*}, \bm{y}^{(t)})\right) \nonumber\\
& \quad+   \frac{8}{\mu}\eta_tL^2 (\delta_{x}^{(t)}+\delta_{y}^{(t)}) + \frac{\eta^2\sigma^2}{n}   \nonumber \\
& \quad + 4\eta^2 L^2 \mathbb{E}\left(\delta_{x}^{(t)} + \left\| \bm{x}^{(t)} - \bm{x}^{*}\right\|^2 + \left\| \bm{y}^*  -  \bm{y}^{(t)} \right\|^2+\delta_{y}^{(t)} \right).\nonumber
\end{align}
Similarly, we can get:
\begin{align}
\mathbb{E}\left[\left\|\bm{y}^{(t+1)} - \bm{y}^{*}\right\|^2\right]  & \leq \left(1-\frac{3}{4} \mu\eta\right)\mathbb{E}\left[\left\|\bm{y}^{(t)} - \bm{y}^{*}\right\|^2\right] -2\eta\mathbb{E}\left(F(\bm{x}^{(t)}, \bm{y}^{*})- F(\bm{x}^{(t)},\bm{y}^{(t)}) \right) \nonumber\\
& \quad +   \frac{8}{\mu}\eta L^2 \mathbb{E}(\delta_{x}^{(t)}+\delta_{y}^{(t)}) + \frac{\eta^2\sigma^2}{n}   \nonumber \\
& \quad + 4\eta^2 L^2 \mathbb{E}\left(\delta_{y}^{(t)} + \left\| \bm{y}^{(t)} - \bm{y}^{*}\right\|^2 + \left\| \bm{x}^*  -  \bm{x}^{(t)} \right\|^2+ \delta_{x}^{(t)} \right).\nonumber
\end{align}
Adding above two inequalities up  yields:
\begin{equation}
 \begin{aligned}
 \mathbb{E}\left[\left\|\bm{x}^{(t+1)} - \bm{x}^{*}\right\|^2\right] + \mathbb{E}\left[\left\|\bm{y}^{(t+1)} - \bm{y}^{*}\right\|^2\right]  
 & \leq  \left(1-\frac{3}{4} \mu\eta\right)\left(\mathbb{E}\left[\left\|\bm{x}^{(t)} - \bm{x}^{*}\right\|^2\right]+\mathbb{E}\left[\left\|\bm{y}^{(t)} - \bm{y}^{*}\right\|^2\right] \right)  \nonumber\\
& \quad -2\eta\mathbb{E}\left( F(\bm{x}^{(t)},\bm{y}^{*}) -  F(\bm{x}^{*}, \bm{y}^{(t)})\right)  +   \frac{16}{\mu} \eta L^2 \mathbb{E}(\delta_{x}^{(t)}+\delta_{y}^{(t)}) + \frac{2\eta^2\sigma^2}{n}   \nonumber \\
& \quad  + 8\eta^2 L^2 \left(\mathbb{E}\left[\delta_{x}^{(t)}+\delta_{y}^{(t)}\right] + \left(\mathbb{E}\left[\left\|\bm{x}^{(t)} - \bm{x}^{*}\right\|^2\right]+\mathbb{E}\left[\left\|\bm{y}^{(t)} - \bm{y}^{*}\right\|^2\right] \right)  \right).\nonumber
 \end{aligned}
\end{equation}
Since $\eta \leq \frac{\sqrt{\mu}}{4\sqrt{2} L  }$,  we have $8\eta^2 L^2   \leq \frac{\mu\eta}{4}$, then we can conclude:
\begin{equation}
 \begin{aligned}
\mathbb{E}\left[\left\|\bm{x}^{(t+1)} - \bm{x}^{*}\right\|^2\right] + \mathbb{E}\left[\left\|\bm{y}^{(t+1)} - \bm{y}^{*}\right\|^2\right]  &\leq  \left(1-\frac{1}{2} \mu\eta\right)\left(\mathbb{E}\left[\left\|\bm{x}^{(t)} - \bm{x}^{*}\right\|^2\right]+\mathbb{E}\left[\left\|\bm{y}^{(t)} - \bm{y}^{*}\right\|^2\right] \right)  \nonumber\\
& \quad -2\eta\mathbb{E}\left( F(\bm{x}^{(t)},\bm{y}^{*}) -  F(\bm{x}^{*}, \bm{y}^{(t)})\right)  +   \frac{16}{\mu}\eta L^2 \mathbb{E}(\delta_{x}^{(t)}+\delta_{y}^{(t)}) + \frac{2\eta^2\sigma^2}{n}   \nonumber \\
& \quad + 8\eta^2 L^2 \mathbb{E}\left(\delta_{x}^{(t)}+\delta_{y}^{(t)}\right)  .\nonumber
 \end{aligned}
\end{equation}
\end{proof}
\end{lemma}

The next lemma characterizes the local model deviation during the dynamics of local SGDA.
\begin{lemma}\label{Lemma_IID: deviation}
For local-SGDA, under Theorem~\ref{Thm: IID_SCSC}'s assumptions, the following relation holds true for any $i,j\in [n]$:
\begin{align}
    \mathbb{E}\left[\|\bm{x}_i^{(t)} - \bm{x}_j^{(t)}\|^2\right] + \mathbb{E}\left[\|\bm{y}_i^{(t)} - \bm{y}_j^{(t)}\|^2 \right]\leq \tau (1+(L-\mu)\eta)^{2\tau} 8\eta^2 \sigma^2.\nonumber
\end{align}
\begin{proof}
Let $i,j \in [n]$, and define $\varepsilon_{\sigma,x}^i = \nabla_x F(\bm{x}_i^{(t)},\bm{y}_i^{(t)} ) -  \nabla_x F(\bm{x}_i^{(t)},\bm{y}_i^{(t)};\xi_i^{(t)})$, $\varepsilon_{\sigma,y}^i = \nabla_y F(\bm{x}_i^{(t)},\bm{y}_i^{(t)} ) -  \nabla_y F(\bm{x}_i^{(t)},\bm{y}_i^{(t)};\xi_i^{(t)})$. Then according to the updating rule, we have:
\begin{align}
    \bm{x}_i^{(t+1)} - \bm{x}_j^{(t+1)} &= \bm{x}_i^{(t)} - \eta \nabla_x F(\bm{x}_i^{(t)},\bm{y}_i^{(t)};\xi_i^{(t)}) - \bm{x}_j^{(t)} + \eta \nabla_x F(\bm{x}_j^{(t)},\bm{y}_j^{(t)};\xi_j^{(t)}) \nonumber \\
    &= \bm{x}_i^{(t)} - \bm{x}_j^{(t)} - \eta \left(\nabla_x F(\bm{x}_i^{(t)},\bm{y}_i^{(t)} ) -  \nabla_x F(\bm{x}_j^{(t)},\bm{y}_j^{(t)} ) \right) \nonumber\\
    &\quad + \eta \left(\nabla_x F(\bm{x}_i^{(t)},\bm{y}_i^{(t)} ) -  \nabla_x F(\bm{x}_i^{(t)},\bm{y}_i^{(t)};\xi_i^{(t)}) \right)+ \eta \left(\nabla_x F(\bm{x}_j^{(t)},\bm{y}_j^{(t)};\xi_j^{(t)})-\nabla_x F(\bm{x}_j^{(t)},\bm{y}_j^{(t)} ) \right) \nonumber\\
    &= \bm{x}_i^{(t)} - \bm{x}_j^{(t)} - \eta \left(\nabla_x F(\bm{x}_i^{(t)},\bm{y}_i^{(t)} ) -  \nabla_x F(\bm{x}_j^{(t)},\bm{y}_i^{(t)} ) \right) - \eta  \left(\nabla_x F(\bm{x}_j^{(t)},\bm{y}_i^{(t)} )  -  \nabla_x F(\bm{x}_j^{(t)},\bm{y}_j^{(t)} ) \right) \nonumber\\
    & \quad + \eta  \varepsilon_{\sigma,x}^i- \eta \varepsilon_{\sigma,x}^j\nonumber\\
    &= (1- \eta_t \mathbf{H}_1)\left( \bm{x}_i^{(t)} - \bm{x}_j^{(t)}\right) - \eta \mathbf{H}_2\left( \bm{y}_i^{(t)}-\bm{y}_j^{(t)}  \right)  + \eta  \varepsilon_{\sigma,x}^i- \eta \varepsilon_{\sigma,x}^j\nonumber,
\end{align}
where we used the  $\mu$-strong-convexity and $L$-smoothness assumptions, that imply  $\mu \mathbf{I}\preccurlyeq \mathbf{H}_1\preccurlyeq L \mathbf{I}$ and $\mu \mathbf{I}\preccurlyeq \mathbf{H}_2\preccurlyeq L \mathbf{I}$. We similarly continue to bound $\bm{y}_i^{(t+1)} - \bm{y}_j^{(t+1)}$:
\begin{align}
    \bm{y}_i^{(t+1)} - \bm{y}_j^{(t+1)} &= \bm{y}_i^{(t)} +\eta  \nabla_y F(\bm{x}_i^{(t)},\bm{y}_i^{(t)};\xi_i^{(t)}) - \bm{y}_j^{(t)} - \eta  \nabla_y F(\bm{x}_j^{(t)},\bm{y}_j^{(t)};\xi_j^{(t)}) \nonumber \\
    &= \bm{y}_i^{(t)} - \bm{y}_j^{(t)} + \eta  \left(\nabla_y F(\bm{x}_i^{(t)},\bm{y}_i^{(t)} ) -  \nabla_y F(\bm{x}_j^{(t)},\bm{y}_j^{(t)} ) \right) \nonumber\\
    &\quad -\eta  \left(\nabla_y F(\bm{x}_i^{(t)},\bm{y}_i^{(t)} ) -  \nabla_y F(\bm{x}_i^{(t)},\bm{y}_i^{(t)};\xi_i^{(t)}) \right)- \eta  \left(\nabla_y F(\bm{x}_j^{(t)},\bm{y}_j^{(t)};\xi_j^{(t)})-\nabla_y F(\bm{x}_j^{(t)},\bm{y}_j^{(t)} ) \right) \nonumber\\
    &= \bm{y}_i^{(t)} - \bm{y}_j^{(t)} + \eta  \left(\nabla_y F(\bm{x}_i^{(t)},\bm{y}_i^{(t)} ) -  \nabla_y F(\bm{x}_i^{(t)},\bm{y}_j^{(t)} ) \right) + \eta  \left(\nabla_y F(\bm{x}_i^{(t)},\bm{y}_j^{(t)} )  -  \nabla_y F(\bm{x}_j^{(t)},\bm{y}_j^{(t)} )\right) \nonumber\\
    &\quad - \eta  \varepsilon_{\sigma,y}^i+\eta \varepsilon_{\sigma,y}^j\nonumber\\
    &= (1 - \eta \mathbf{H}_3)\left( \bm{y}_i^{(t)} - \bm{y}_j^{(t)}\right) - \eta \mathbf{H}_4\left( \bm{x}_i^{(t)}-\bm{x}_j^{(t)}  \right)  - \eta  \varepsilon_{\sigma,y}^i+\eta \varepsilon_{\sigma,y}^j\nonumber,
\end{align}
where $\mu \mathbf{I}\preccurlyeq \mathbf{H}_3\preccurlyeq L \mathbf{I}$ and $\mu \mathbf{I}\preccurlyeq \mathbf{H}_4\preccurlyeq L \mathbf{I}$.

Let $\varepsilon_{x}^{t}= \bm{x}_i^{(t)} - \bm{x}_j^{(t)}$, $\varepsilon_{y}^{t}= \bm{y}_i^{(t)} - \bm{y}_j^{(t)}$. Writing the above inequalities into compact matrix form, we have:
\begin{align}
  \begin{bmatrix}
 \varepsilon_{x}^{t+1} \\
 \varepsilon_{y}^{t+1} 
\end{bmatrix} = \mathcal{A}^t \begin{bmatrix}
\varepsilon_{x}^{t} \\
 \varepsilon_{y}^{t} 
\end{bmatrix}  + \begin{bmatrix}
\eta  \mathbf{I},&0\\
 0,&\eta  \mathbf{I}
\end{bmatrix}\begin{bmatrix}
 \varepsilon_{\sigma,x}^i- \varepsilon_{\sigma,x}^j \\
 \varepsilon_{\sigma,y}^j- \varepsilon_{\sigma,y}^i
\end{bmatrix}, \label{eq: IID_SCSC_L2_0}
\end{align}
where:
\begin{align}
  \mathcal{A}^t = \begin{bmatrix}
 (1- \eta \mathbf{H}_1),& - \eta \mathbf{H}_2\\
- \eta \mathbf{H}_4, &(1 - \eta \mathbf{H}_3)
\end{bmatrix} .
\end{align}
Taking squared norm and expectation over (\ref{eq: IID_SCSC_L2_0}) yields:
\begin{align}
 \mathbb{E}\left[\left\| \begin{bmatrix}
 \varepsilon_{x}^{t+1} \\
 \varepsilon_{y}^{t+1} 
\end{bmatrix}\right\|^2\right] &=\mathbb{E}\left[\left\|  \mathcal{A}^t \begin{bmatrix}
\varepsilon_{x}^{t} \\
 \varepsilon_{y}^{t} 
\end{bmatrix}\right\|^2\right]  + \mathbb{E}\left[\left\| \begin{bmatrix}
\eta \mathbf{I},&0\\
 0,&\eta  \mathbf{I}
\end{bmatrix}\begin{bmatrix}
 \varepsilon_{\sigma,x}^i- \varepsilon_{\sigma,x}^j  \\
\varepsilon_{\sigma,y}^j- \varepsilon_{\sigma,y}^i
\end{bmatrix}  \right\|^2\right] \nonumber\\
&\leq \mathbb{E}\left[\left\|  \mathcal{A}^t\right\|^2\right] \mathbb{E}\left[\left\|\begin{bmatrix}
\varepsilon_{x}^{t} \\
 \varepsilon_{y}^{t} 
\end{bmatrix}\right\|^2\right]+8\eta^2\sigma^2. \label{eq: IID_SCSC_L2_1}
\end{align}
Now let us examine the upper bound of $\left\|  \mathcal{A}^t\right\|^2$. According to~\cite{yuan2020federated} (Lemma G.1), we have:
\begin{align*}
 \| \mathcal{A}^t\| = \left\|\begin{bmatrix}
 (1- \eta \mathbf{H}_1),& - \eta \mathbf{H}_2\\
- \eta \mathbf{H}_4, &(1 - \eta \mathbf{H}_3)
\end{bmatrix}\right\| \leq \max\{\|1- \eta \mathbf{H}_1\|, \|1- \eta \mathbf{H}_3\|\} +  \max\{\|\eta \mathbf{H}_2\|,\|\eta \mathbf{H}_4\|\} = 1 +(L-\mu)\eta.
\end{align*}
So $\left\|  \mathcal{A}^t\right\|^2 \leq (1 +(L-\mu)\eta)^2$. Letting $t_0$ denote the latest synchronization stage, and plugging $\left\|  \mathcal{A}^t\right\|^2 \leq (1 +(L-\mu)\eta)^2$ back to (\ref{eq: IID_SCSC_L2_1}) we have:
\begin{align*}
 \mathbb{E}\left[\left\| \begin{bmatrix}
 \varepsilon_{x}^{t+1} \\
 \varepsilon_{y}^{t+1} 
\end{bmatrix}\right\|^2\right] 
&\leq (1 +(L-\mu)\eta)^2 \mathbb{E}\left[\left\|\begin{bmatrix}
\varepsilon_{x}^{t} \\
 \varepsilon_{y}^{t} 
\end{bmatrix}\right\|^2\right]+8\eta^2\sigma^2\\
&\leq \sum_{t'=0}^{t-t_0} (1 +(L-\mu)\eta)^{2t'} 8\eta^2\sigma^2\\
&\leq \tau (1 +(L-\mu)\eta)^{2\tau} 8\eta^2\sigma^2,
\end{align*}
where we use the fact $\left\|\begin{bmatrix}
\varepsilon_{x}^{t_0} \\
 \varepsilon_{y}^{t_0} 
\end{bmatrix}\right\|^2 = 0$ at second inequality.

\end{proof}

\end{lemma}
 
 \subsubsection{Proof of Theorem~\ref{Thm: IID_SCSC}}\label{sec: proof-thm-scsc-iid}
Now we can proceed to the proof of Theorem~\ref{Thm: IID_SCSC}.
 \begin{proof}
 According to Lemma~\ref{Lemma_IID: scsc one iteration} we have:

 \begin{align}
 \mathbb{E}\left[\left\|\bm{x}^{(t+1)} - \bm{x}^{*}\right\|^2\right] + \mathbb{E}\left[\left\|\bm{y}^{(t+1)} - \bm{y}^{*}\right\|^2\right] & \leq \left(1-\frac{1}{2}\mu\eta\right)\left(\mathbb{E}\left[\left\|\bm{x}^{(t)} - \bm{x}^{*}\right\|^2\right] + \mathbb{E}\left[\left\|\bm{y}^{(t)} - \bm{y}^{*}\right\|^2\right]\right) \nonumber \\
& \quad -2\eta \left(\mathbb{E}\left[F(\bm{x}^{(t)},\bm{y}^*)\right] - \mathbb{E}\left[F(\bm{x}^{*},\bm{y}^{(t)})\right]\right) \nonumber \\
& \quad +\frac{2\eta^2\sigma^2}{n}+ \frac{16\eta_tL^2}{\mu}\left(\mathbb{E}\left[\delta_{\bm{x}}^{(t)}+\delta_{\bm{y}}^{(t)}\right]\right) + 8\eta^2L^2  \left(\mathbb{E}\left[\delta_{\bm{x}}^{(t)}+\delta_{\bm{y}}^{(t)}\right]\right).\label{eq:proof-thm-scsc-iid_1}
 \end{align}

Notice that $F(\bm{x}^{(t)},\bm{y}^*) - F(\bm{x}^{*},\bm{y}^{(t)})  = F(\bm{x}^{(t)},\bm{y}^*) - F(\bm{x}^{*},\bm{y}^*)+ F(\bm{x}^{*},\bm{y}^*)-F(\bm{x}^{*},\bm{y}^{(t)}) \geq 0$, we can omit this term. We plug Lemma~\ref{Lemma_IID: deviation} into (\ref{eq:proof-thm-scsc-iid_1}) to get: 
 \begin{align}
\mathbb{E}\left[\left\|\bm{x}^{(t+1)} - \bm{x}^{*}\right\|^2\right] + \mathbb{E}\left[\left\|\bm{y}^{(t+1)} - \bm{y}^{*}\right\|^2\right]& \leq \left(1-\frac{1}{2}\mu\eta\right)\left(\mathbb{E}\left[\left\|\bm{x}^{(t)} - \bm{x}^{*}\right\|^2\right] + \mathbb{E}\left[\left\|\bm{y}^{(t)} - \bm{y}^{*}\right\|^2\right]\right) \nonumber \\ 
& \quad +\frac{2\eta^2\sigma^2}{n}+ \left(\frac{16\eta L^2}{\mu}+8\eta^2L^2   \right)\left(\tau (1+(L-\mu)\eta)^{2\tau} 8\eta^2 \sigma^2\right) \nonumber.
 \end{align}
 Unrolling the recursion yields:
  \begin{align}
 \mathbb{E}\left[\left\|\bm{x}^{(T)} - \bm{x}^{*}\right\|^2 + \left\|\bm{y}^{(T)} - \bm{y}^{*}\right\|^2\right] & \leq \left(1-\frac{1}{2}\mu\eta\right)^T \left(\mathbb{E}\left[\left\|\bm{x}^{(0)} - \bm{x}^{*}\right\|^2  + \left\|\bm{y}^{(0)} - \bm{y}^{*}\right\|^2\right]\right) \nonumber \\ 
& \quad +\frac{2\eta\sigma^2}{\mu n}+ \left(\frac{32L^2}{\mu^2}+\frac{16\eta L^2   }{\mu}\right)\left(\tau (1+(L-\mu)\eta)^{2\tau} 8\eta^2 \sigma^2\right) \nonumber.
 \end{align}
Plugging in $\tau =  \frac{T}{n\log T}  $ and $ \eta = \frac{4\log T}{\mu T}$, we have:
 \begin{align}
&\mathbb{E}\left[\left\|\bm{x}^{(T)} - \bm{x}^{*}\right\|^2 + \left\|\bm{y}^{(T)} - \bm{y}^{*}\right\|^2\right]\nonumber\\
 & \leq \left(1-\frac{2\log T}{T} \right)^T\left(\mathbb{E}\left[\left\|\bm{x}^{(0)} - \bm{x}^{*}\right\|^2  + \left\|\bm{y}^{(0)} - \bm{y}^{*}\right\|^2\right]\right) \nonumber \\ 
& \quad +\frac{8\log T \sigma^2}{\mu^2nT}+ \left(\frac{32L^2}{\mu^2}+16\frac{4\log T}{\mu^2 T} L^2  )\right)\left( \frac{T}{n\log T}\left(1+(L-\mu)\frac{4\log T}{\mu T}\right)^{2 \frac{T}{n\log T}}  \frac{128\log^2 T}{\mu^2 T^2} \sigma^2\right) \nonumber\\
 & \leq \exp(-\log T^2)\left(\mathbb{E}\left[\left\|\bm{x}^{(0)} - \bm{x}^{*}\right\|^2  + \left\|\bm{y}^{(0)} - \bm{y}^{*}\right\|^2\right]\right) \nonumber \\ 
& \quad +\frac{8\log T \sigma^2}{\mu^2nT}+ \left(\frac{32L^2}{\mu^2}+16\frac{4\log T}{\mu^2 T} L^2   \right)\left( \frac{T}{n\log T}\left(1+(L-\mu)\frac{4\log T}{\mu T}\right)^{2 \frac{T}{n\log T}}  \frac{128\log^2 T}{\mu^2 T^2} \sigma^2\right) \nonumber.
 \end{align}
Notice that:
\begin{align}
    \left(1+(L-\mu)\frac{4\log T}{\mu T}\right)^{  \frac{2T}{n\log T}} =  \left(1+(L-\mu)\frac{4\log T}{\mu T}\right)^{ \frac{\mu T}{4(L-\mu)\log T}   \frac{2T}{n\log T} \frac{4(L-\mu)\log T}{\mu T}} \leq \exp\left( \frac{8(L-\mu) }{\mu n} \right)\nonumber.
\end{align}
So we can conclude the proof:
 {\begin{align}
\mathbb{E} &\left[\left\|\bm{x}^{(T)} - \bm{x}^{*}\right\|^2 + \left\|\bm{y}^{(T)} - \bm{y}^{*}\right\|^2\right]  \nonumber\\
&\quad\leq \frac{\mathbb{E}\left[\left\|\bm{x}^{(0)} - \bm{x}^{*}\right\|^2  + \left\|\bm{y}^{(0)} - \bm{y}^{*}\right\|^2\right]}{T^2} \nonumber \\ 
& \quad+\frac{8\log T \sigma^2}{\mu^2nT}+ \left(\frac{32L^2}{\mu^2}+16\frac{4\log T}{\mu^2 T} L^2   \right)\left( \frac{T}{n\log T} \exp\left( \frac{8(L-\mu) }{\mu n} \right) \frac{128\log^2 T}{\mu^2 T^2} \sigma^2\right) \nonumber\\
&\quad\leq \Tilde{O}\left(\frac{1}{T^2}  +\frac{ \sigma^2}{\mu^2 nT }  +    \frac{ \kappa^2\sigma^2 }{\mu^2nT} + \frac{ \kappa^2\sigma^2 }{\mu^2nT^2} \right).\nonumber
 \end{align}} 
as stated where we used $\Tilde{O}(\cdot)$ in last inequality to keep key parameters. 
 \end{proof}

\subsection{Proof in heterogeneous setting}\label{sec: proof-thm-scsc-non-iid}
 In this section we are going to present the proof in heterogeneous case. Let us introduce some technical lemmas first which will help our proof.

\subsubsection{Proof of technical lemmas} \label{Sec: SCSC-Non-iid-lemmas}
The following lemma performs one iteration analysis:
\begin{lemma} \label{lemma: scsc non-iid one iteration}
For local-SGDA, under Theorem~\ref{Thm: SCSC}'s assumptions, the following relation holds true:
\begin{equation}
 \begin{aligned}
\mathbb{E}\left[\left\|\bm{x}^{(t+1)} - \bm{x}^{*}\right\|^2\right] + \mathbb{E}\left[\left\|\bm{y}^{(t+1)} - \bm{y}^{*}\right\|^2\right]  &\leq  \left(1-\frac{1}{2} \mu\eta_t\right)\left(\mathbb{E}\left[\left\|\bm{x}^{(t)} - \bm{x}^{*}\right\|^2\right]+\mathbb{E}\left[\left\|\bm{y}^{(t)} - \bm{y}^{*}\right\|^2\right] \right)  \nonumber\\
& \quad -2\eta_t\left( F(\bm{x}^{(t)},\bm{y}^{*}) -  F(\bm{x}^{*}, \bm{y}^{(t)})\right)  +   \frac{16}{\mu}\eta_tL^2 (\delta_{x}^{(t)}+\delta_{y}^{(t)}) + \frac{2\eta_t^2\sigma^2}{n}   \nonumber \\
& \quad + 8\eta_t^2 L^2 \left(\delta_{x}^{(t)}+\delta_{y}^{(t)}\right)  .\nonumber
 \end{aligned}
\end{equation}
\begin{proof}
According to updating rule and strong convexity:
\begin{align}
\mathbb{E}\left[\left\|\bm{x}^{(t+1)} - \bm{x}^{*}\right\|^2\right] &= \mathbb{E}\left[\left\|\bm{x}^{(t )} - \eta_t \frac{1}{n}\sum_{i=1}^n \nabla_x f_i(\bm{x}_i^{(t)}, \bm{y}_i^{(t)}; \xi_i^{(t)}) - \bm{x}^{*}\right\|^2\right] \nonumber\\
& \leq \mathbb{E}\left[\left\|\bm{x}^{(t)} - \bm{x}^{*}\right\|^2\right] -2\eta_t\left\langle   \frac{1}{n}\sum_{i=1}^n \nabla_x f_i(\bm{x}_i^{(t)}, \bm{y}_i^{(t)} ),  \bm{x}^{(t)} - \bm{x}^{*} \right\rangle \nonumber\\
&+ \frac{\eta_t^2\sigma^2}{n} + \eta_t^2\mathbb{E}\left[\left\|\frac{1}{n}\sum_{i=1}^n \nabla_x f_i(\bm{x}_i^{(t)}, \bm{y}_i^{(t)})\right\|^2\right]\nonumber\\
&\leq \mathbb{E}\left[\left\|\bm{x}^{(t)} - \bm{x}^{*}\right\|^2\right] -2\eta_t\left\langle     \nabla_x F(\bm{x}^{(t)}, \bm{y}^{(t)} ),  \bm{x}^{(t)} - \bm{x}^{*} \right\rangle \nonumber\\
&-2\eta_t \left\langle  \frac{1}{n}\sum_{i=1}^n \nabla_x f_i(\bm{x}_i^{(t)},\bm{y}_i^{(t)}) - \nabla_xF(\bm{x}^{(t)},\bm{y}^{(t)}) , \bm{x}^{(t)} - \bm{x}^{*}   \right\rangle \nonumber\\
&+ \frac{\eta_t^2\sigma^2}{n} + \eta_t^2\mathbb{E}\left[\left\|\frac{1}{n}\sum_{i=1}^n \nabla_x f_i(\bm{x}_i^{(t)}, \bm{y}_i^{(t)})\right\|^2\right] \nonumber\\
&\leq \left(1- \mu\eta_t\right)\mathbb{E}\left[\left\|\bm{x}^{(t)} - \bm{x}^{*}\right\|^2\right] -2\eta_t\left( F(\bm{x}^{(t)},\bm{y}^{(t)}) -  F(\bm{x}^{*}, \bm{y}^{(t)})\right) \nonumber\\
&+ \eta_t \mathbb{E}\left(\frac{4}{\mu}\left\|\frac{1}{n}\sum_{i=1}^n \nabla_x f_i(\bm{x}_i^{(t)},\bm{y}_i^{(t)}) - \nabla_xF(\bm{x}^{(t)},\bm{y}^{(t)}) \right\|^2 +\frac{\mu}{4}\left\| \bm{x}^{(t)} - \bm{x}^{*} \right\|^2 \right) \nonumber \\
&+ \frac{\eta_t^2\sigma^2}{n} + \eta_t^2\mathbb{E}\left[\left\|\frac{1}{n}\sum_{i=1}^n \nabla_x f_i(\bm{x}_i^{(t)}, \bm{y}_i^{(t)})\right\|^2\right].\nonumber
\end{align}
Now we are going to bound terms $\left\|\frac{1}{n}\sum_{i=1}^n \nabla_x f_i(\bm{x}_i^{(t)},\bm{y}_i^{(t)}) - \nabla_xF(\bm{x}^{(t)},\bm{y}^{(t)}) \right\|^2$ and $\left\|\frac{1}{n}\sum_{i=1}^n \nabla_x f_i(\bm{x}_i^{(t)}, \bm{y}_i^{(t)})\right\|^2$ .

By applying Jensen's inequality on $\left\|\frac{1}{n}\sum_{i=1}^n \nabla_x f_i(\bm{x}_i^{(t)},\bm{y}_i^{(t)}) - \nabla_xF(\bm{x}^{(t)},\bm{y}^{(t)}) \right\|^2$ we have:
\begin{align}
&\left\|\frac{1}{n}\sum_{i=1}^n \nabla_x f_i(\bm{x}_i^{(t)},\bm{y}_i^{(t)}) - \nabla_xF(\bm{x}^{(t)},\bm{y}^{(t)}) \right\|^2 =  \frac{1}{n}\sum_{i=1}^n\left\| \nabla_x f_i(\bm{x}_i^{(t)},\bm{y}_i^{(t)}) - \nabla_x f_i(\bm{x}^{(t)},\bm{y}^{(t)}) \right\|^2 \nonumber\\
&\leq \frac{1}{n}\sum_{i=1}^n\left(2\left\| \nabla_x f_i(\bm{x}_i^{(t)},\bm{y}_i^{(t)}) - \nabla_x f_i(\bm{x}^{(t)},\bm{y}_i^{(t)}) \right\|^2 + 2\left\| \nabla_x f_i(\bm{x}^{(t)},\bm{y}_i^{(t)}) - \nabla_x f_i(\bm{x}^{(t)},\bm{y}^{(t)}) \right\|^2\right)\nonumber\\
&\leq \frac{1}{n}\sum_{i=1}^n\left(2L^2\left\|  \bm{x}_i^{(t)}  - \bm{x}^{(t)}  \right\|^2 + 2L^2\left\|  \bm{y}_i^{(t)} - \bm{y}^{(t)}  \right\|^2\right) \nonumber\\
& \leq 2L^2 (\delta_{x}^{(t)}+\delta_{y}^{(t)}),\nonumber
\end{align}
where we use the smoothness in the second last inequality.

Then we switch to bound $\left\|\frac{1}{n}\sum_{i=1}^n \nabla_x f_i(\bm{x}_i^{(t)}, \bm{y}_i^{(t)})\right\|^2$:
\begin{align}
\left\|\frac{1}{n}\sum_{i=1}^n \nabla_x f_i(\bm{x}_i^{(t)}, \bm{y}_i^{(t)})\right\|^2 &= \frac{1}{n}\sum_{i=1}^n\left\| \nabla_x f_i(\bm{x}_i^{(t)}, \bm{y}_i^{(t)})\right\|^2\nonumber\\
&= 2\left\|\frac{1}{n}\sum_{i=1}^n \nabla_x f_i(\bm{x}_i^{(t)}, \bm{y}_i^{(t)}) -\nabla_x F(\bm{x}^{(t)}, \bm{y}^{(t)}) \right\|^2 + 2\left\| \nabla_x F(\bm{x}^{(t)}, \bm{y}^{(t)}) - \nabla_x F( \bm{x}^* , \bm{y}^*) \right\|^2 \nonumber\\ 
&\leq L^2\frac{1}{n}\sum_{i=1}^n 4 \left(\left\| \bm{x}_i^{(t)} - \bm{x}^{(t)}\right\|^2 + \left\| \bm{x}^{(t)} - \bm{x}^{*}\right\|^2 + \left\|\bm{y}_i^{(t)} - \bm{y}^{(t)}\right\|^2+ \left\| \bm{y}^{(t)} - \bm{y}^{*}\right\|^2 \right) .\nonumber
\end{align}
 
Putting these pieces together yields:
\begin{align}
\mathbb{E}\left[\left\|\bm{x}^{(t+1)} - \bm{x}^{*}\right\|^2\right]   &\leq \left(1-\frac{3}{4} \mu\eta_t\right)\mathbb{E}\left[\left\|\bm{x}^{(t)} - \bm{x}^{*}\right\|^2\right] -2\eta_t\left( \mathbb{E}\left[F(\bm{x}^{(t)},\bm{y}^{(t)}) -  F(\bm{x}^{*}, \bm{y}^{(t)})\right]\right) \nonumber\\
& \quad +   \frac{8}{\mu}\eta_tL^2 \mathbb{E}(\delta_{x}^{(t)}+\delta_{y}^{(t)}) + \frac{\eta_t^2\sigma^2}{n}   \nonumber \\
& \quad + 4\eta_t^2 L^2 \mathbb{E}\left(\delta_{x}^{(t)} + \left\| \bm{x}^{(t)} - \bm{x}^{*}\right\|^2 +  \left\| \bm{y}^*  -  \bm{y}^{(t)} \right\|^2+  \delta_{y}^{(t)} \right).\nonumber
\end{align}
Similarly, we can get:
\begin{align}
\mathbb{E}\left[\left\|\bm{y}^{(t+1)} - \bm{y}^{*}\right\|^2\right]   &\leq \left(1-\frac{3}{4} \mu\eta_t\right)\mathbb{E}\left[\left\|\bm{y}^{(t)} - \bm{y}^{*}\right\|^2\right] -2\eta_t\mathbb{E}\left(F(\bm{x}^{(t)}, \bm{y}^{*})- F(\bm{x}^{(t)},\bm{y}^{(t)}) \right) \nonumber\\
& \quad +   \frac{8}{\mu}\eta_tL^2 \mathbb{E}(\delta_{x}^{(t)}+\delta_{y}^{(t)}) + \frac{\eta_t^2\sigma^2}{n}   \nonumber \\
& \quad + 4\eta_t^2 L^2 \mathbb{E}\left(\delta_{y}^{(t)} + \left\| \bm{y}^{(t)} - \bm{y}^{*}\right\|^2 +  \left\| \bm{x}^*  -  \bm{x}^{(t)} \right\|^2+  \delta_{x}^{(t)} \right).\nonumber
\end{align}
Combining the above two inequalities   yields:
\begin{equation}
 \begin{aligned}
 &\mathbb{E}\left[\left\|\bm{x}^{(t+1)} - \bm{x}^{*}\right\|^2\right] + \mathbb{E}\left[\left\|\bm{y}^{(t+1)} - \bm{y}^{*}\right\|^2\right]  \nonumber\\
 & \leq  \left(1-\frac{3}{4} \mu\eta_t\right)\left(\mathbb{E}\left[\left\|\bm{x}^{(t)} - \bm{x}^{*}\right\|^2\right]+\mathbb{E}\left[\left\|\bm{y}^{(t)} - \bm{y}^{*}\right\|^2\right] \right)  \nonumber\\
& \quad -2\eta_t\mathbb{E}\left( F(\bm{x}^{(t)},\bm{y}^{*}) -  F(\bm{x}^{*}, \bm{y}^{(t)})\right)  +   \frac{16}{\mu}\eta_tL^2 (\delta_{x}^{(t)}+\delta_{y}^{(t)}) + \frac{2\eta_t^2\sigma^2}{n}   \nonumber \\
& \quad  + 8\eta_t^2 L^2 \left( \mathbb{E}\left[\delta_{x}^{(t)}+\delta_{y}^{(t)} \right] +  \left(\mathbb{E}\left[\left\|\bm{x}^{(t)} - \bm{x}^{*}\right\|^2\right]+\mathbb{E}\left[\left\|\bm{y}^{(t)} - \bm{y}^{*}\right\|^2\right] \right)  \right).\nonumber
 \end{aligned}
\end{equation}
Since $\eta_t = \frac{8}{\mu(t+a)}$ and $a=\max\{  2048\kappa^2\tau , 1024\sqrt{2}\tau  \kappa^2, 256\kappa^2\}$, so we have $8\eta_t^2 L^2   \leq \frac{\mu\eta_t}{4}$, then we can conclude:
\begin{equation}
 \begin{aligned}
\mathbb{E}\left[\left\|\bm{x}^{(t+1)} - \bm{x}^{*}\right\|^2\right] + \mathbb{E}\left[\left\|\bm{y}^{(t+1)} - \bm{y}^{*}\right\|^2\right]  &\leq  \left(1-\frac{1}{2} \mu\eta_t\right)\left(\mathbb{E}\left[\left\|\bm{x}^{(t)} - \bm{x}^{*}\right\|^2\right]+\mathbb{E}\left[\left\|\bm{y}^{(t)} - \bm{y}^{*}\right\|^2\right] \right)  \nonumber\\
& \quad -2\eta_t\left( \mathbb{E}\left[F(\bm{x}^{(t)},\bm{y}^{*}) -  F(\bm{x}^{*}, \bm{y}^{(t)})\right]\right)  +   \frac{16}{\mu}\eta_tL^2 (\mathbb{E}\left[\delta_{x}^{(t)}+\delta_{y}^{(t)}\right]) + \frac{2\eta_t^2\sigma^2}{n}   \nonumber \\
& \quad + 8\eta_t^2 L^2 \left(\mathbb{E}\left[\delta_{x}^{(t)}+\delta_{y}^{(t)}\right]\right)  .\nonumber
 \end{aligned}
\end{equation}
\end{proof}
\end{lemma}

The next lemma upper bounds the weighted accumulative local model deviations between two communication rounds in strongly convex setting under  heterogeneous data assumption.
\begin{lemma}\label{lemma:scsc deviation1}
 For local-SGDA, under Theorem~\ref{Thm: SCSC}'s assumption, by letting $w_t = (t+a)^2$,  the following inequality holds:
\begin{equation}
 \begin{aligned}
 \sum_{t=s\tau}^{(s+1)\tau}w_t (\mathbb{E}\left[\delta_{\bm{x}}^{(t)}+\delta_{\bm{y}}^{(t)}\right] ) &\leq       \frac{\mu}{64L^2} \sum_{j=s\tau}^{(s+1)\tau}\mu \eta_j\frac{w_{j}}{\eta_{j}}\left(\mathbb{E}\left[\left\| \bm{x} ^{(j)}  -  \bm{x}^{*}   \right\|^2   + \left\| \bm{y}^{(j)} -\bm{y}^{*}   \right\|^2\right]\right) \nonumber \\
& \quad   +  64 \tau^2   \sum_{j=s\tau}^{(s+1)\tau}w_j\eta_j^2\left(\Delta_{x}+\Delta_{y}\right)   + 32  \tau^2  \sum_{j=s\tau}^{(s+1)\tau}w_j\eta_j^2\sigma^2.\nonumber
\end{aligned}
\end{equation}
where $\delta_{\bm{x}}^{(t)} = \frac{1}{n}\sum_{i=1}^n\left\|\bm{x}_i^{(t)}-\bm{x}^{(t)}\right\|^2, \quad \delta_{\bm{y}}^{(t)} = \frac{1}{n}\sum_{i=1}^n\left\|\bm{y}_i^{(t)}-\bm{y}^{(t)}\right\|^2$.
\begin{proof}
Assume that $s\tau \leq t\leq (s+1)\tau$. According to the updating rule, we have:
\begin{equation}
 \begin{aligned}
\delta_{\bm{x}}^{(t)} &= \frac{1}{n}\sum_{i=1}^n\left\|\bm{x}_i^{(t)}-\bm{x}^{(t)}\right\|^2\nonumber\\
& = \frac{1}{n}\sum_{i=1}^n\left\|\bm{x}^{(s\tau)} -  \sum_{j=s\tau}^{t} \eta_j \nabla_x f_i\left(\bm{x}_i^{(j)},\bm{y}_i^{(j)};\xi_i^{(j)} \right) -\left(\bm{x}^{(s\tau)} - \frac{1}{n}\sum_{k=1}^n\sum_{j=s\tau}^{t} \eta_j \nabla_x f_k\left(\bm{x}_i^{(k)},\bm{y}_i^{(k)};\xi_i^{(k)} \right)\right) \right\|^2  \nonumber\\
& = \frac{1}{n}\sum_{i=1}^n\left\|  \sum_{j=s\tau}^{t-1} \eta_j\nabla_x f_i\left(\bm{x}_i^{(j)},\bm{y}_i^{(j)};\xi_i^{(j)} \right)  - \frac{1}{n}\sum_{k=1}^n\sum_{j=s\tau}^{t-1} \eta_j\nabla_x f_k\left(\bm{x}_i^{(k)},\bm{y}_i^{(k)};\xi_i^{(k)} \right)  \right\|^2  \nonumber\\
& \leq \frac{1}{n}\sum_{i=1}^n\left\|  \sum_{j=s\tau}^{t-1}\eta_j \nabla_x f_i\left(\bm{x}_i^{(j)},\bm{y}_i^{(j)};\xi_i^{(j)} \right)   \right\|^2  \nonumber\\
& \leq \frac{1}{n}\sum_{i=1}^n\tau\sum_{j=s\tau}^{(s+1)\tau} \eta_j^2\left(2\left\|   \nabla_x f_i\left(\bm{x}_i^{(j)},\bm{y}_i^{(j)} \right)   \right\|^2 + 2\sigma^2\right).\nonumber
\end{aligned}
\end{equation}
By applying Jensen's inequality to $\left\|   \nabla_x f_i\left(\bm{x}_i^{(j)},\bm{y}_i^{(j)} \right)   \right\|^2$:
\begin{align}
&\left\|   \nabla_x f_i\left(\bm{x}_i^{(j)},\bm{y}_i^{(j)} \right)   \right\|^2 \leq 4\left\|   \nabla_x f_i\left(\bm{x}_i^{(j)},\bm{y}_i^{(j)} \right)  - \nabla_x f_i\left(\bm{x}^{(j)},\bm{y}^{(j)} \right)  \right\|^2 + 4\left\|   \nabla_x f_i\left(\bm{x} ^{(j)},\bm{y} ^{(j)} \right)  - \nabla_x f_i\left(\bm{x}^{*},\bm{y}^{(j)} \right)  \right\|^2 \nonumber\\
&+ 4\left\|    \nabla_x f_i\left(\bm{x}^{*},\bm{y}^{(j)} \right)   - \nabla_x f_i\left(\bm{x}^{*},\bm{y}^{*} \right)  \right\|^2 + 4\left\| \nabla_x f_i\left(\bm{x}^{*},\bm{y}^{*} \right)  \right\|^2  \nonumber\\
& \leq 8L^2\left(\left\| \bm{x}_i^{(j)} - \bm{x}^{(j)}  \right\|^2 + \left\| \bm{y}_i^{(j)} - \bm{y}^{(j)}  \right\|^2\right) + 4L^2\left\| \bm{x} ^{(j)}  -  \bm{x}^{*}   \right\|^2 \nonumber\\
& + 4L^2\left\| \bm{y}^{(j)} -\bm{y}^{*}   \right\|^2 + 4\left\| \nabla_x f_i\left(\bm{x}^{*},\bm{y}^{*} \right)  \right\|^2.\nonumber
\end{align}
Plugging back and taking expectation yields:
\begin{align}
\mathbb{E}\left[\delta_{\bm{x}}^{(t)} \right]  &\leq \frac{1}{n}\sum_{i=1}^n\tau\sum_{j=s\tau}^{(s+1)\tau} \nonumber\\
&\times \eta_j^2\left(16L^2\left(\mathbb{E}\left[\left\| \bm{x}_i^{(j)} - \bm{x}^{(j)}  \right\|^2 + \left\| \bm{y}_i^{(j)} - \bm{y}^{(j)}  \right\|^2\right] \right) + \mathbb{E}\left[8L^2\left\| \bm{x} ^{(j)}  -  \bm{x}^{*}   \right\|^2   + 8L^2\left\| \bm{y}^{(j)} -\bm{y}^{*}   \right\|^2\right] \right) \nonumber \\
& + \frac{1}{n}\sum_{i=1}^n\tau\sum_{j=s\tau}^{(s+1)\tau} \left(\eta_j^2 8 \left\| \nabla_x f_i\left(\bm{x}^{*},\bm{y}^{*} \right)  \right\|^2  + 2\sigma^2\right)\nonumber\\
& \leq \tau\sum_{j=s\tau}^{(s+1)\tau} \eta_j^2\left(16L^2\left(\delta_{\bm{x}}^{(j)} + \delta_{\bm{y}}^{(j)} \right) + 8L^2\left(\mathbb{E}\left[\left\| \bm{x} ^{(j)}  -  \bm{x}^{*}   \right\|^2   + \left\| \bm{y}^{(j)} -\bm{y}^{*}   \right\|^2\right]\right)\right) \nonumber\\
& +  8\tau  \sum_{j=s\tau}^{(s+1)\tau}\eta_j^2 \Delta_{x}   + 2 \tau  \sum_{j=s\tau}^{(s+1)\tau}\eta_j^2\sigma^2.  \nonumber  
\end{align}
Then multiplying $w_t$ on both sides and summing from $t = s\tau$ to $(s+1)\tau$ yields:
\begin{align}
\sum_{t=s\tau}^{(s+1)\tau}w_t\mathbb{E}\left[ \delta_{\bm{x}}^{(t)}\right]    & \leq \sum_{j=s\tau}^{(s+1)\tau}w_t  \tau\sum_{j=s\tau}^{(s+1)\tau} \eta_j^2\left(16L^2\left(\mathbb{E}\left[\delta_{\bm{x}}^{(j)} + \delta_{\bm{y}}^{(j)} \right]\right) + 8L^2\left(\mathbb{E}\left[\left\| \bm{x} ^{(j)}  -  \bm{x}^{*}   \right\|^2   + \left\| \bm{y}^{(j)} -\bm{y}^{*}   \right\|^2\right]\right)\right)\nonumber \\
&  +  8\sum_{t=s\tau}^{(s+1)\tau}w_t\tau   \sum_{j=s\tau}^{(s+1)\tau}\eta_j^2\Delta_{x}   + 2\sum_{t=s\tau}^{(s+1)\tau}w_t\tau  \sum_{j=s\tau}^{(s+1)\tau}\eta_j^2\sigma^2.  \nonumber  
\end{align}
Notice that $w_t = (t+a)^2$ and $a\geq \tau$, so $w_t<w_{(s+1)\tau}\leq 4 w_j$, $\forall t,j$ such that $ s\tau \leq t,j\leq (s+1)\tau$. So we have:
\begin{align}
\sum_{t=s\tau}^{(s+1)\tau}w_t \mathbb{E}\left[ \delta_{\bm{x}}^{(t)}\right]      & \leq     \tau^2 \sum_{j=s\tau}^{(s+1)\tau}4w_j \eta_j^2\left(16L^2\left(\mathbb{E}\left[\delta_{\bm{x}}^{(j)} + \delta_{\bm{y}}^{(j)}\right]  \right) + 8L^2\left(\mathbb{E}\left[\left\| \bm{x} ^{(j)}  -  \bm{x}^{*}   \right\|^2   + \left\| \bm{y}^{(j)} -\bm{y}^{*}   \right\|^2\right] \right)\right)\nonumber \\
&   +  32 \tau^2   \sum_{j=s\tau}^{(s+1)\tau}w_j\eta_j^2\Delta_{x}   + 8  \tau^2  \sum_{j=s\tau}^{(s+1)\tau}w_j\eta_j^2\sigma^2.  \nonumber  
\end{align}
Since $\eta_t = \frac{8}{\mu(t+a)}$ and $a=\max\{  2048\kappa^2\tau , 1024\sqrt{2}\tau  \kappa^2, 256\kappa^2\}$, we have the following facts:
\begin{align}
&\eta_t<\eta_{s\tau}\leq 2 \eta_j, \quad \forall t,j \ \text{such that} \ s\tau \leq t,j\leq (s+1)\tau,\nonumber\\
&256\eta_t^2\tau^2 L^2 \leq \frac{1}{4},\nonumber\\
&128\eta_t^2\tau^2 L^2 \leq \frac{\mu^2}{256L^2}.\nonumber
\end{align}
Hence:
\begin{align}
\sum_{t=s\tau}^{(s+1)\tau}w_t \mathbb{E}\left[\delta_{\bm{x}}^{(t)}\right]   &  \leq     4\eta_t^2\tau^2 \sum_{j=s\tau}^{(s+1)\tau}4w_j \left(16L^2\left(\mathbb{E}\left[\delta_{\bm{x}}^{(j)} + \delta_{\bm{y}}^{(j)}\right] \right) + 8L^2\left(\mathbb{E}\left[\left\| \bm{x} ^{(j)}  -  \bm{x}^{*}   \right\|^2   + \left\| \bm{y}^{(j)} -\bm{y}^{*}   \right\|^2\right]\right)\right)\nonumber \\
&   \quad +  32 \tau^2   \sum_{j=s\tau}^{(s+1)\tau}w_j\eta_j^2\Delta_{x}   + 8  \tau^2  \sum_{j=s\tau}^{(s+1)\tau}w_j\eta_j^2\sigma^2.  \nonumber  \\
 &  \leq     \frac{1}{4}  \sum_{j=s\tau}^{(s+1)\tau} w_j \  \left(\mathbb{E}\left[\delta_{\bm{x}}^{(j)} + \delta_{\bm{y}}^{(j)}\right] \right) + 128\eta_t^2\tau^2 L^2 \sum_{j=s\tau}^{(s+1)\tau}w_j\left(\mathbb{E}\left[\left\| \bm{x} ^{(j)}  -  \bm{x}^{*}   \right\|^2   + \left\| \bm{y}^{(j)} -\bm{y}^{*}   \right\|^2\right]\right) \nonumber \\
& \quad +  32 \tau^2   \sum_{j=s\tau}^{(s+1)\tau}w_j\eta_j^2\Delta_{x}   + 8  \tau^2  \sum_{j=s\tau}^{(s+1)\tau}w_j\eta_j^2\sigma^2\nonumber \\
 &  \leq     \frac{1}{4}  \sum_{j=s\tau}^{(s+1)\tau} w_j \  \left(\mathbb{E}\left[\delta_{\bm{x}}^{(j)} + \delta_{\bm{y}}^{(j)}\right] \right) +  \frac{\mu^2}{256L^2} \sum_{j=s\tau}^{(s+1)\tau}w_j\left(\mathbb{E}\left[\left\| \bm{x} ^{(j)}  -  \bm{x}^{*}   \right\|^2   + \left\| \bm{y}^{(j)} -\bm{y}^{*}   \right\|^2\right]\right) \nonumber \\
&  \quad +  32 \tau^2   \sum_{j=s\tau}^{(s+1)\tau}w_j\eta_j^2\Delta_{x}   + 8  \tau^2  \sum_{j=s\tau}^{(s+1)\tau}w_j\eta_j^2\sigma^2 \nonumber\\
   &  \leq     \frac{1}{4 }  \sum_{j=s\tau}^{(s+1)\tau} w_j \  \left(\mathbb{E}\left[\delta_{\bm{x}}^{(j)} + \delta_{\bm{y}}^{(j)}\right] \right) +  \frac{\mu}{256L^2} \sum_{j=s\tau}^{(s+1)\tau}\mu \eta_j\frac{w_{j}}{\eta_{j}}\left(\mathbb{E}\left[\left\| \bm{x} ^{(j)}  -  \bm{x}^{*}   \right\|^2   + \left\| \bm{y}^{(j)} -\bm{y}^{*}   \right\|^2\right]\right) \nonumber \\
& \quad +  32 \tau^2   \sum_{j=s\tau}^{(s+1)\tau}w_j\eta_j^2\Delta_{x}   + 8  \tau^2  \sum_{j=s\tau}^{(s+1)\tau}w_j\eta_j^2\sigma^2.\nonumber
\end{align}
Similarly, we get:
\begin{align}
\sum_{t=s\tau}^{(s+1)\tau}w_t \mathbb{E}[\delta_{\bm{y}}^{(t)}]  &  \leq     \frac{1}{4 }  \sum_{j=s\tau}^{(s+1)\tau} w_j \  \left(\mathbb{E}\left[\delta_{\bm{x}}^{(j)} + \delta_{\bm{y}}^{(j)}\right]\right) +  \frac{\mu}{256L^2} \sum_{j=s\tau}^{(s+1)\tau}\mu \eta_j\frac{w_{j}}{\eta_{j}}\left(\mathbb{E}\left[\left\| \bm{x} ^{(j)}  -  \bm{x}^{*}   \right\|^2   + \left\| \bm{y}^{(j)} -\bm{y}^{*}   \right\|^2\right]\right) \nonumber \\
& \quad +  32 \tau^2   \sum_{j=s\tau}^{(s+1)\tau}w_j\eta_j^2\Delta_{y}   + 8  \tau^2  \sum_{j=s\tau}^{(s+1)\tau}w_j\eta_j^2\sigma^2.\nonumber
\end{align}
Adding the two inequalities up gives:
\begin{align}
\sum_{t=s\tau}^{(s+1)\tau}w_t \left(\mathbb{E}\left[\delta_{\bm{x}}^{(t)}+\delta_{\bm{y}}^{(t)} \right]\right)   &\leq   \frac{1}{2 }  \sum_{j=s\tau}^{(s+1)\tau} w_j \  \left(\mathbb{E}\left[\delta_{\bm{x}}^{(j)} + \delta_{\bm{y}}^{(j)}\right] \right) +  \frac{\mu}{128L^2} \sum_{j=s\tau}^{(s+1)\tau}\mu \eta_j\frac{w_{j}}{\eta_{j}}\left(\mathbb{E}\left[\left\| \bm{x} ^{(j)}  -  \bm{x}^{*}   \right\|^2   + \left\| \bm{y}^{(j)} -\bm{y}^{*}   \right\|^2\right]\right) \nonumber \\
&+  32 \tau^2   \sum_{j=s\tau}^{(s+1)\tau}w_j\eta_j^2 \left(\Delta_{x}+\Delta_{y}\right)   + 16  \tau^2  \sum_{j=s\tau}^{(s+1)\tau}w_j\eta_j^2\sigma^2  \nonumber\\
  \Longleftrightarrow \frac{1}{2}\sum_{t=s\tau}^{(s+1)\tau}w_t \left(\mathbb{E}\left[\delta_{\bm{x}}^{(t)}+\delta_{\bm{y}}^{(t)} \right]\right) &\leq       \frac{\mu}{128L^2} \sum_{j=s\tau}^{(s+1)\tau}\mu \eta_j\frac{w_{j}}{\eta_{j}}\left(\mathbb{E}\left[\left\| \bm{x} ^{(j)}  -  \bm{x}^{*}   \right\|^2   + \left\| \bm{y}^{(j)} -\bm{y}^{*}   \right\|^2\right]\right) \nonumber \\
& \quad  +  32 \tau^2   \sum_{j=s\tau}^{(s+1)\tau}w_j\eta_j^2\left(\Delta_{x}+\Delta_{y}\right)   + 16  \tau^2  \sum_{j=s\tau}^{(s+1)\tau}w_j\eta_j^2\sigma^2 \nonumber\\
  \Longleftrightarrow  \sum_{t=s\tau}^{(s+1)\tau}w_t \left(\mathbb{E}\left[\delta_{\bm{x}}^{(t)}+\delta_{\bm{y}}^{(t)} \right] \right) &\leq       \frac{\mu}{64L^2} \sum_{j=s\tau}^{(s+1)\tau}\mu \eta_j\frac{w_{j}}{\eta_{j}}\left(\mathbb{E}\left[\left\| \bm{x} ^{(j)}  -  \bm{x}^{*}   \right\|^2   + \left\| \bm{y}^{(j)} -\bm{y}^{*}   \right\|^2\right]\right) \nonumber \\
&  \quad +  64 \tau^2   \sum_{j=s\tau}^{(s+1)\tau}w_j\eta_j^2\left(\Delta_{x}+\Delta_{y}\right)   + 32  \tau^2  \sum_{j=s\tau}^{(s+1)\tau}w_j\eta_j^2\sigma^2.\nonumber
\end{align}
\end{proof}
\end{lemma}

The following lemma also gives the upper bound for weighted local model deviations, but the weights multiplied in front of $\mathbb{E}\left[\delta_{\bm{x}}^{(t)}+\delta_{\bm{y}}^{(t)} \right] $ is different from Lemma~\ref{lemma:scsc deviation1}. 
\begin{lemma}\label{lemma:scsc deviation2}
For local-SGDA, under Theorem~\ref{Thm: SCSC}'s assumption, by letting $w_t = (t+a)^2$,  the following  holds:
\begin{align}
 \sum_{t=s\tau}^{(s+1)\tau}w_t\eta_t \left(\mathbb{E}\left[\delta_{\bm{x}}^{(t)}+\delta_{\bm{y}}^{(t)} \right] \right) &\leq \frac{1}{64L^2} \sum_{j=s\tau}^{(s+1)\tau}\mu \eta_j\frac{w_{j}}{\eta_{j}}\left(\mathbb{E}\left[\left\| \bm{x} ^{(j)}  -  \bm{x}^{*}   \right\|^2   + \left\| \bm{y}^{(j)} -\bm{y}^{*}   \right\|^2\right]\right) \nonumber \\
& \quad +  128 \tau^2   \sum_{j=s\tau}^{(s+1)\tau}w_j\eta_j^3\left(\Delta_{x}+\Delta_{y}\right)   + 64  \tau^2  \sum_{j=s\tau}^{(s+1)\tau}w_j\eta_j^3\sigma^2.\nonumber
\end{align}

\begin{proof}
 
According to Lemma~\ref{lemma:scsc deviation1}, we have:
\begin{align}
\sum_{t=s\tau}^{(s+1)\tau}w_t \eta_t \mathbb{E}\left[\delta_{\bm{x}}^{(t)} \right] &   \leq \sum_{t=s\tau}^{(s+1)\tau}w_t  \tau \eta_t\sum_{j=s\tau}^{(s+1)\tau} \eta_j^2\left(16L^2\left(\mathbb{E}\left[\delta_{\bm{x}}^{(j)} + \delta_{\bm{y}}^{(j)}\right] \right) + 8L^2\left(\mathbb{E}\left[\left\| \bm{x} ^{(j)}  -  \bm{x}^{*}   \right\|^2   + \left\| \bm{y}^{(j)} -\bm{y}^{*}   \right\|^2\right]\right)\right) \nonumber\\
& \quad +  8\sum_{t=s\tau}^{(s+1)\tau}w_t\tau   \sum_{j=s\tau}^{(s+1)\tau}\eta_j^2\Delta_{x}   + 2\sum_{t=s\tau}^{(s+1)\tau}w_t\tau  \sum_{j=s\tau}^{(s+1)\tau}\eta_j^2\sigma^2.  \nonumber \\
\end{align}
Notice that $w_t = (t+a)^2$ and $a\geq \tau$, so $w_t<w_{(s+1)\tau}\leq 4 w_j$, $\forall t,j$ such that $ s\tau \leq t,j\leq (s+1)\tau$. So we have:
\begin{align}
\sum_{t=s\tau}^{(s+1)\tau}w_t \eta_t \mathbb{E}\left[\delta_{\bm{x}}^{(t)} \right]   &  \leq     \tau^2\eta_t \sum_{j=s\tau}^{(s+1)\tau}4w_j \eta_j^2\left(16L^2\left(\mathbb{E}\left[\delta_{\bm{x}}^{(j)} + \delta_{\bm{y}}^{(j)}\right] \right) + 8L^2\left(\mathbb{E}\left[\left\| \bm{x} ^{(j)}  -  \bm{x}^{*}   \right\|^2   + \left\| \bm{y}^{(j)} -\bm{y}^{*}   \right\|^2\right]\right)\right)\nonumber  \\
& \quad +  32 \tau^2 \eta_t  \sum_{j=s\tau}^{(s+1)\tau}w_j\eta_j^2\Delta_{x}   + 8  \tau^2 \eta_t \sum_{j=s\tau}^{(s+1)\tau}w_j\eta_j^2\sigma^2.  \nonumber \\
& \leq     \tau^2  \sum_{j=s\tau}^{(s+1)\tau}4w_j \eta_j^2\left(16L^2\left(\mathbb{E}\left[\delta_{\bm{x}}^{(j)} + \delta_{\bm{y}}^{(j)}\right] \right) + 8L^2\left(\mathbb{E}\left[\left\| \bm{x} ^{(j)}  -  \bm{x}^{*}   \right\|^2   + \left\| \bm{y}^{(j)} -\bm{y}^{*}   \right\|^2\right]\right)\right) \label{eq: Lemma 2 Eq 1}\\
&  \quad +  32 \tau^2 \eta_t  \sum_{j=s\tau}^{(s+1)\tau}w_j\eta_j^2\Delta_{x}   + 8  \tau^2 \eta_t \sum_{j=s\tau}^{(s+1)\tau}w_j\eta_j^2\sigma^2,  \nonumber 
\end{align}
where we omit a $\eta_t$ in~(\ref{eq: Lemma 2 Eq 1}) since $\eta_t \leq 1$.

Since $\eta_t = \frac{8}{\mu(t+a)}$ and $a=\max\{  2048\kappa^2\tau , 1024\sqrt{2}\tau  \kappa^2, 256\kappa^2\}$, we have the following facts:
\begin{align}
&\eta_t<\eta_{s\tau}\leq 2 \eta_j, \quad \forall t,j \ \text{such that} \ s\tau \leq t,j\leq (s+1)\tau,\nonumber\\
&256\eta_t^2\tau^2 \leq \frac{1}{4},\nonumber\\
&128\eta_t^2\tau^2 L^2 \leq \frac{\mu}{256L^2 }.\nonumber
\end{align}
Hence:
\begin{align}
\sum_{t=s\tau}^{(s+1)\tau}w_t\eta_t \mathbb{E}\left[\delta_{\bm{x}}^{(t)} \right]   &  \leq     4\eta_t^2\tau^2 \sum_{j=s\tau}^{(s+1)\tau}4w_j \left(16L^2\left(\mathbb{E}\left[\delta_{\bm{x}}^{(j)} + \delta_{\bm{y}}^{(j)}\right] \right) + 8L^2\left(\mathbb{E}\left[\left\| \bm{x} ^{(j)}  -  \bm{x}^{*}   \right\|^2   + \left\| \bm{y}^{(j)} -\bm{y}^{*}   \right\|^2\right]\right)\right) \nonumber\\
& \quad +  64 \tau^2   \sum_{j=s\tau}^{(s+1)\tau}w_j\eta_j^3\Delta_{x}   + 16 \tau^2  \sum_{j=s\tau}^{(s+1)\tau}w_j\eta_j^3\sigma^2.  \nonumber  \\
 & \leq     \frac{1}{4}  \sum_{j=s\tau}^{(s+1)\tau} w_j \  \left(\mathbb{E}\left[\delta_{\bm{x}}^{(j)} + \delta_{\bm{y}}^{(j)}\right] \right) + 128\eta_t^2\tau^2 L^2 \sum_{j=s\tau}^{(s+1)\tau}w_j\left(\mathbb{E}\left[\left\| \bm{x} ^{(j)}  -  \bm{x}^{*}   \right\|^2   + \left\| \bm{y}^{(j)} -\bm{y}^{*}   \right\|^2\right]\right) \nonumber \\
& \quad  +  64 \tau^2   \sum_{j=s\tau}^{(s+1)\tau}w_j\eta_j^3\Delta_{x}   + 16  \tau^2  \sum_{j=s\tau}^{(s+1)\tau}w_j\eta_j^2\sigma^2 \nonumber\\
 &\leq     \frac{1}{4 }  \sum_{j=s\tau}^{(s+1)\tau} w_j \  \left(\mathbb{E}\left[\delta_{\bm{x}}^{(j)} + \delta_{\bm{y}}^{(j)}\right] \right) +  \frac{\mu}{256L^2 } \sum_{j=s\tau}^{(s+1)\tau}w_j\left(\mathbb{E}\left[\left\| \bm{x} ^{(j)}  -  \bm{x}^{*}   \right\|^2   + \left\| \bm{y}^{(j)} -\bm{y}^{*}   \right\|^2\right]\right) \nonumber \\
& \quad +  64 \tau^2   \sum_{j=s\tau}^{(s+1)\tau}w_j\eta_j^3\Delta_{x}   + 16  \tau^2  \sum_{j=s\tau}^{(s+1)\tau}w_j\eta_j^3\sigma^2\nonumber\\
& \leq     \frac{1}{4 }  \sum_{j=s\tau}^{(s+1)\tau} w_j \  \left(\mathbb{E}\left[\delta_{\bm{x}}^{(j)} + \delta_{\bm{y}}^{(j)}\right] \right) +  \frac{1}{256L^2} \sum_{j=s\tau}^{(s+1)\tau}\mu \eta_j\frac{w_{j}}{\eta_{j}}\left(\mathbb{E}\left[\left\| \bm{x} ^{(j)}  -  \bm{x}^{*}   \right\|^2   + \left\| \bm{y}^{(j)} -\bm{y}^{*}   \right\|^2\right]\right) \nonumber \\
&  \quad +  64 \tau^2   \sum_{j=s\tau}^{(s+1)\tau}w_j\eta_j^3\Delta_{x}   + 16  \tau^2  \sum_{j=s\tau}^{(s+1)\tau}w_j\eta_j^3\sigma^2.\nonumber
\end{align}
Similarly, we get:
\begin{align}
\sum_{t=s\tau}^{(s+1)\tau}w_t\eta_t \mathbb{E}\left[\delta_{\bm{y}}^{(t)} \right]   & \leq     \frac{1}{4 }  \sum_{j=s\tau}^{(s+1)\tau} w_j \  \left(\mathbb{E}\left[\delta_{\bm{x}}^{(j)} + \delta_{\bm{y}}^{(j)}\right] \right) + \frac{1}{256L^2} \sum_{j=s\tau}^{(s+1)\tau}\mu \eta_j\frac{w_{j}}{\eta_{j}}\left(\mathbb{E}\left[\left\| \bm{x} ^{(j)}  -  \bm{x}^{*}   \right\|^2   + \left\| \bm{y}^{(j)} -\bm{y}^{*}   \right\|^2\right]\right) \nonumber \\
& \quad +  64 \tau^2   \sum_{j=s\tau}^{(s+1)\tau}w_j\eta_j^3\Delta_{y}   + 16 \tau^2  \sum_{j=s\tau}^{(s+1)\tau}w_j\eta_j^3\sigma^2.\nonumber
\end{align}
Combining the two inequalities yields:
{\begin{align}
\sum_{t=s\tau}^{(s+1)\tau}w_t \eta_t(\mathbb{E}\left[\delta_{\bm{x}}^{(t)}+\delta_{\bm{y}}^{(t)}\right] )  & \leq     \frac{1}{2 }  \sum_{j=s\tau}^{(s+1)\tau} w_j \  \left(\mathbb{E}\left[\delta_{\bm{x}}^{(j)} + \delta_{\bm{y}}^{(j)}\right] \right) +  \frac{1}{128L^2 } \sum_{j=s\tau}^{(s+1)\tau}\mu \eta_j\frac{w_{j}}{\eta_{j}}\left(\mathbb{E}\left[\left\| \bm{x} ^{(j)}  -  \bm{x}^{*}   \right\|^2   + \left\| \bm{y}^{(j)} -\bm{y}^{*}   \right\|^2\right]\right) \nonumber \\
& \quad +  64 \tau^2   \sum_{j=s\tau}^{(s+1)\tau}w_j\eta_j^3 \left(\Delta_{x}+\Delta_{y}\right)   + 32 \tau^2  \sum_{j=s\tau}^{(s+1)\tau}w_j\eta_j^3\sigma^2\nonumber\\
  \Longleftrightarrow \frac{1}{2}\sum_{t=s\tau}^{(s+1)\tau}w_t\eta_t (\mathbb{E}\left[\delta_{\bm{x}}^{(t)}+\delta_{\bm{y}}^{(t)}\right]  ) &\leq        \frac{1}{128L^2} \sum_{j=s\tau}^{(s+1)\tau}\mu \eta_j\frac{w_{j}}{\eta_{j}}\left(\mathbb{E}\left[\left\| \bm{x} ^{(j)}  -  \bm{x}^{*}   \right\|^2   + \left\| \bm{y}^{(j)} -\bm{y}^{*}   \right\|^2\right]\right) \nonumber \\
& \quad+  64 \tau^2   \sum_{j=s\tau}^{(s+1)\tau}w_j\eta_j^3\left(\Delta_{x}+\Delta_{y}\right)   + 32  \tau^2  \sum_{j=s\tau}^{(s+1)\tau}w_j\eta_j^3\sigma^2\nonumber\\
  \Longleftrightarrow  \sum_{t=s\tau}^{(s+1)\tau}w_t\eta_t (\mathbb{E}\left[\delta_{\bm{x}}^{(t)}+\delta_{\bm{y}}^{(t)}\right]  ) &\leq       \frac{1}{64L^2 } \sum_{j=s\tau}^{(s+1)\tau}\mu \eta_j\frac{w_{j}}{\eta_{j}}\left(\mathbb{E}\left[\left\| \bm{x} ^{(j)}  -  \bm{x}^{*}   \right\|^2   + \left\| \bm{y}^{(j)} -\bm{y}^{*}   \right\|^2\right]\right) \nonumber \\
& \quad +  128 \tau^2   \sum_{j=s\tau}^{(s+1)\tau}w_j\eta_j^3\left(\Delta_{x}+\Delta_{y}\right)   + 64  \tau^2  \sum_{j=s\tau}^{(s+1)\tau}w_j\eta_j^3\sigma^2.\nonumber
\end{align}}
\end{proof}
\end{lemma}

\subsubsection{Proof of Theorem~\ref{Thm: SCSC}}
Now we are going to proof Theorem~\ref{Thm: SCSC}.
\begin{proof}
 According to Lemma~\ref{lemma: scsc non-iid one iteration} we have:
\begin{equation}
 \begin{aligned}
\mathbb{E}\left[\left\|\bm{x}^{(t+1)} - \bm{x}^{*}\right\|^2\right] + \mathbb{E}\left[\left\|\bm{y}^{(t+1)} - \bm{y}^{*}\right\|^2\right]  &\leq  \left(1-\frac{1}{2} \mu\eta_t\right)\left(\mathbb{E}\left[\left\|\bm{x}^{(t)} - \bm{x}^{*}\right\|^2\right]+\mathbb{E}\left[\left\|\bm{y}^{(t)} - \bm{y}^{*}\right\|^2\right] \right)  \nonumber\\
& \quad -2\eta_t\mathbb{E}\left( F(\bm{x}^{(t)},\bm{y}^{*}) -  F(\bm{x}^{*}, \bm{y}^{(t)})\right)  +   \frac{16}{\mu}\eta_tL^2 \mathbb{E}(\delta_{x}^{(t)}+\delta_{y}^{(t)}) + \frac{2\eta_t^2\sigma^2}{n}   \nonumber \\
& \quad + 8\eta_t^2 L^2 \mathbb{E}\left(\delta_{x}^{(t)}+\delta_{y}^{(t)}\right)  .\nonumber
 \end{aligned}
\end{equation}
Then, letting $w_t = (t+a)^2$ and multiplying $\frac{w_t}{\eta_t}$ on both sides, and summing up from $t=1$ to $T$:
\begin{align}
&\sum_{s=0}^{S-1}\sum_{t=s\tau}^{(s+1)\tau} \frac{w_t}{\eta_t}\mathbb{E}\left(\left\|\bm{x}^{(t+1)} - \bm{x}^{*}\right\|^2 + \left\|\bm{y}^{(t+1)} - \bm{y}^{*}\right\|^2 \right)\nonumber\\
&\leq \sum_{s=0}^{S-1}\sum_{t=s\tau}^{(s+1)\tau}\left(1-\frac{1}{2}\mu\eta_t \right)\frac{w_t}{\eta_t}\mathbb{E}\left(\left\|\bm{x}^{(t)} - \bm{x}^{*}\right\|^2 + \left\|\bm{y}^{(t)} - \bm{y}^{*}\right\|^2\right)\nonumber\\
& \quad -2\sum_{s=0}^{S-1}\sum_{t=s\tau}^{(s+1)\tau} w_t \mathbb{E}\left(F(\bm{x}^{(t)},\bm{y}^*) - F(\bm{x}^{*},\bm{y}^{(t)})\right)+\sum_{s=0}^{S-1}\sum_{t=s\tau}^{(s+1)\tau} \frac{2w_t\eta_t\sigma^2}{n} \nonumber \nonumber\\
& \quad + \underbrace{\frac{16L^2}{\mu}\sum_{s=0}^{S-1}\sum_{t=s\tau}^{(s+1)\tau} w_t\mathbb{E}\left(\delta_{\bm{x}}^{(t)}+\delta_{\bm{y}}^{(t)}\right)}_{T_1} + \underbrace{8L^2  \sum_{s=0}^{S-1}\sum_{t=s\tau}^{(s+1)\tau} w_t\eta_t\mathbb{E}\left(\delta_{\bm{x}}^{(t)}+\delta_{\bm{y}}^{(t)}\right)}_{T_2}.\label{eq: Thm 1 Eq 1}
\end{align}
Then we use  Lemmas~\ref{lemma:scsc deviation1} and~\ref{lemma:scsc deviation2} in $T_1$ and $T_2$ to get:
\begin{align}
T_1 &=  \sum_{s=0}^{S-1}\sum_{t=s\tau}^{(s+1)\tau}\frac{1}{8} \mu \eta_t\frac{w_{t}}{\eta_{t}}\mathbb{E}\left(\left\| \bm{x} ^{(t)}  -  \bm{x}^{*}   \right\|^2   + \left\| \bm{y}^{(t)} -\bm{y}^{*}   \right\|^2\right) \nonumber \\
&  \quad +  \frac{1024\tau^2L^2}{\mu}     \sum_{s=0}^{S-1}\sum_{t=s\tau}^{(s+1)\tau}w_t\eta_t^2\left(\Delta_{x}+\Delta_{y}\right)   + \frac{512\tau^2L^2}{\mu}\sum_{s=0}^{S-1} \sum_{t=s\tau}^{(s+1)\tau}w_t\eta_t^2\sigma^2\nonumber
\end{align}
\begin{align}
T_2 &=  \sum_{s=0}^{S-1}\sum_{t=s\tau}^{(s+1)\tau}\frac{1}{8} \mu \eta_t\frac{w_{t}}{\eta_{t}}\mathbb{E}\left(\left\| \bm{x} ^{(t)}  -  \bm{x}^{*}   \right\|^2   + \left\| \bm{y}^{(t)} -\bm{y}^{*}   \right\|^2\right) \nonumber \\
& \quad  +  1024L^2 \tau^2     \sum_{s=0}^{S-1}\sum_{t=s\tau}^{(s+1)\tau}w_t\eta_t^3\left(\Delta_{x}+\Delta_{y}\right)   + 512L^2 \tau^2\sum_{s=0}^{S-1} \sum_{t=s\tau}^{(s+1)\tau}w_t\eta_t^3\sigma^2.\nonumber
\end{align}
Plugging $T_1$ and $T_2$ back into (\ref{eq: Thm 1 Eq 1}) yields:
{\begin{align}
\sum_{s=0}^{S-1}\sum_{t=s\tau}^{(s+1)\tau} \frac{w_t}{\eta_t}\mathbb{E}\left(\left\|\bm{x}^{(t+1)} - \bm{x}^{*}\right\|^2 + \left\|\bm{y}^{(t+1)} - \bm{y}^{*}\right\|^2 \right) & \leq \sum_{s=0}^{S-1}\sum_{t=s\tau}^{(s+1)\tau}\left(1-\frac{1}{4}\mu\eta_t \right)\frac{w_t}{\eta_t}\mathbb{E}\left(\left\|\bm{x}^{(t)} - \bm{x}^{*}\right\|^2 + \left\|\bm{y}^{(t)} - \bm{y}^{*}\right\|^2\right)\nonumber\\
& \quad -2\sum_{s=0}^{S-1}\sum_{t=s\tau}^{(s+1)\tau} w_t \mathbb{E}\left(F(\bm{x}^{(t)},\bm{y}^*) - F(\bm{x}^{*},\bm{y}^{(t)})\right)+\sum_{s=0}^{S-1}\sum_{t=s\tau}^{(s+1)\tau} \frac{2w_t\eta_t\sigma^2}{n} \nonumber \\
& \quad + \left(\frac{1024\tau^2L^2}{\mu}+  1024L^2 \tau^2    \right) \left(\Delta_{x}+\Delta_{y}\right)\sum_{s=0}^{S-1}\sum_{t=s\tau}^{(s+1)\tau}w_t\left(\eta_t^2+\eta_t^3\right)\nonumber\\
& \quad + \left(\frac{512\tau^2L^2}{\mu}+  512L^2 \tau^2    \right) \sigma^2 \sum_{s=0}^{S-1}\sum_{t=s\tau}^{(s+1)\tau}w_t\left(\eta_t^2+\eta_t^3\right).\nonumber
\end{align}}
Using the fact that $\left(1-\frac{1}{4}\mu\eta_t \right)\frac{w_t}{\eta_t} \leq \frac{w_{t-1}}{\eta_{t-1}}$, we can cancel up the terms:
\begin{align}
& \frac{w_T}{\eta_T}\mathbb{E}\left(\left\|\bm{x}^{(T+1)} - \bm{x}^{*}\right\|^2 + \left\|\bm{y}^{(T+1)} - \bm{y}^{*}\right\|^2 \right)\nonumber\\
&  \leq  \frac{w_{0}}{\eta_{0}}\left(\left\|\bm{x}^{(1)} - \bm{x}^{*}\right\|^2 + \left\|\bm{y}^{(1)} - \bm{y}^{*}\right\|^2\right)   \nonumber\\ 
&  \quad +  \left(\frac{1024\tau^2L^2}{\mu}+  1023L^2 \tau^2    \right) \left(\Delta_{x}+\Delta_{y}\right)\sum_{s=0}^{S-1}\sum_{t=s\tau}^{(s+1)\tau}w_t\left(\eta_t^2+\eta_t^3\right)\nonumber\\
& \quad +   \left(\frac{512\tau^2L^2}{\mu}+ 512L^2 \tau^2   \right) \sigma^2 \sum_{s=0}^{S-1}\sum_{t=s\tau}^{(s+1)\tau}w_t\left(\eta_t^2+\eta_t^3\right)+  \sum_{s=0}^{S-1}\sum_{t=s\tau}^{(s+1)\tau} \frac{2w_t\eta_t\sigma^2}{n}\nonumber.
\end{align}
Dividing both side by $\frac{w_T}{\eta_T}$ yields:
\begin{align} 
&\mathbb{E}\left[\left\|\bm{x}^{(T+1)} - \bm{x}^{*}\right\|^2 + \left\|\bm{y}^{(T+1)} - \bm{y}^{*}\right\|^2\right]\nonumber\\
&\leq \frac{8}{ \mu(T+a)^3}  \frac{w_{0}}{\eta_{0}}\left(\left\|\bm{x}^{(1)} - \bm{x}^{*}\right\|^2 + \left\|\bm{y}^{(1)} - \bm{y}^{*}\right\|^2\right)\nonumber\\
& \quad + \frac{8}{ \mu(T+a)^3} \left(\frac{1024\tau^2L^2}{\mu}+  1024L^2\tau^2    \right) \left(\Delta_{x}+\Delta_{y}\right) \left( \frac{64T}{\mu^2} + \frac{\Theta \left(\ln T\right)}{\mu^3}\right)\nonumber\\
& \quad + \frac{8}{ \mu(T+a)^3} \left(\frac{512\tau^2L^2}{\mu}+  512L^2 \tau^2    \right) \sigma^2 \left( \frac{64T}{\mu^2} + \frac{\Theta \left(\ln T\right)}{\mu^3}\right)+ \frac{8}{ \mu(T+a)^2}  \frac{16T\sigma^2}{\mu n} \nonumber\\
& \leq O\left(\frac{ a^3}{T^3} \right)   +  O\left(\frac{\kappa^2  \tau^2\left(\Delta_{x}+\Delta_{y}\right)}{\mu T^2} \right) + O\left(\frac{\kappa^2  \tau^2\sigma^2}{\mu T^2} \right) + O\left(\frac{\sigma^2}{\mu^2 nT} \right).\nonumber
\end{align}
Plugging in $\tau = \sqrt{T/n}$  concludes the proof.
\end{proof}

\section{Proof of Nonconvex-Strongly-Concave Case}\label{sec:ncsc}
\subsection{Overview of proofs}
Now we proceed to the proof of convergence rate in nonconvex-strongly-concave setting. Recall that in this case we study the envelope function $\Phi(\cdot)$ and $\bm{y}^{*}(\cdot)$. The following proposition establishes the smoothness property of these auxiliary functions.
\begin{proposition}[Lin et al~\cite{lin2019gradient}]
If a function $f(\bm{x},\cdot)$ is $\mu$-strongly concave and $L$ smooth, then $\Phi(\bm{x})$ is $\beta=\kappa L + L$ smooth and $\bm{y}^{*}(\bm{x})$ is $ \kappa $-Lipschitz where $\kappa = L/\mu$.
\end{proposition}

Since $\Phi$ is $\beta$-smooth, then the starting point is to conduct the standard analysis scheme for nonconvex smooth function on one iteration as follows:
\begin{align}
\mathbb{E}\left[\Phi (\bm{x}^{(t+1)} )\right] - \mathbb{E}\left[\Phi(\bm{x}^{(t)})\right] & \leq -\frac{\eta}{2} \mathbb{E}\left[\left\| \nabla \Phi(\bm{x}^{(t)}) \right\|^2\right] - \left(\eta_x - 3\beta \eta_x^2 \right)  \mathbb{E}\left[\left\| \frac{1}{n} \sum_{i=1}^n \nabla_x f_i(\bm{x}^{(t)}, \bm{y}^{(t)}) \right\|^2\right] \nonumber\\
&\quad + \left(2 \eta  +3\beta\eta_x^2\right)L^2 \mathbb{E}\left[\left(\delta_{\bm{x}}^{(t)} + \delta_{\bm{y}}^{(t)} \right)\right]
+\frac{\eta_x L^2}{2}  \mathbb{E}\left[\left\| \bm{y}^*(\bm{x}^{(t)}) - \bm{y}^{(t)}\right\|^2\right] + \frac{3\beta \eta_x^2 \sigma^2}{2n}.\nonumber
\end{align}
We can see the convergence depends on $\delta_{\bm{x}}^{(t)} + \delta_{\bm{y}}^{(t)}$, and a new term: $\left\| \bm{y}^*(\bm{x}^{(t)}) - \bm{y}^{(t)}\right\|^2$. The bound we derived for $\delta_{\bm{x}}^{(t)} + \delta_{\bm{y}}^{(t)}$ is no longer suitable here since in nonconvex objective, convergence to global saddle point is NP-hard. Instead, we derive the following deviation bound with the help of \emph{gradient dissimilarity}:
\begin{align}
 \frac{1}{T}\sum_{t=1}^{T} \mathbb{E}\left(\delta_{\bm{x}}^{(t)} + \delta_{\bm{y}}^{(t)} \right) \leq 10\tau^2 (\eta_x^2 + \eta_y^2) \left( \sigma^2 + \frac{\sigma^2}{n}  \right)+10\tau^2\eta_x^2\zeta_x+ 10\tau^2 \eta_y^2 \zeta_y .\nonumber
\end{align} 

Another thing is to bound the gap of current dual iterate and optimal dual variable: $\left\| \bm{y}^*(\bm{x}^{(t)}) - \bm{y}^{(t)}\right\|^2$. \cite{lin2019gradient} has established the convergence of it, but they use a fairly large dual step size $O(1/L)$. However, in the local descent method, due to the issue of local model drifting, we are forced to stick with a small step size. Thus, as our main contribution in this part, we established the convergence of $\left\| \bm{y}^*(\bm{x}^{(t)}) - \bm{y}^{(t)}\right\|^2$ using a smaller dual step size:

{\begin{align}
      \frac{1}{T}\sum_{t=1}^{T} \mathbb{E}\left[\left\|  \bm{y}^{(t)} - \bm{y}^* (\bm{x}^{(t)})\right\|^2 \right]  & \leq \frac{2C\kappa}{T}   \mathbb{E}\left[\left\| \bm{y}^{(0)} -  \bm{y}^* (\bm{x}^{(0)}) \right\|^2 \right]+ O\left(  \frac{C\eta_y^2\sigma^2}{n}\right) \nonumber\\
     & \quad  + \frac{1}{T}\sum_{t=1}^T O\left(C\left( \eta_y   +  \eta_y^2 \right) + C^2 \eta_x^2 \right)  \mathbb{E}\left[  \delta_{\bm{x}}^{(t)} + \delta_{\bm{y}}^{(t)} \right]  \nonumber \\ &\quad + \frac{1}{T}\sum_{t=1}^T O \left(C^2  \eta_x^2\mathbb{E}\left[ \left\|\frac{1}{n}\sum_{i=1}^n \nabla_x f_i(\bm{x} ^{(t)},\bm{y} ^{(t)})\right\|^2  \right]\right),\nonumber 
\end{align}} 
where $C = \frac{2}{\eta_y L}$. $C$ could be large if we choose $\eta_y$ to be small, and will thus negatively affect convergence rate, which means we trade some rate for communication efficiency.

Putting these piece together, and letting $\eta_x$ and $\eta_y$ to be sufficiently small, we can cancel up the term $\mathbb{E}\left[\left\| \frac{1}{n} \sum_{i=1}^n \nabla_x f_i(\bm{x}^{(t)}, \bm{y}^{(t)}) \right\|^2\right]$ and establish the convergence rate. 

\subsection{Proof of technical lemmas}
Before proceeding to the main proof of theorem, let us introduce a few  useful intermediate results. The following lemma shows the analysis for one iteration of local SGDA, on nonconvex-strongly-concave function.
\begin{lemma}\label{lm4}
For local-SGDA, under the assumptions in Theorem~\ref{Thm: NCSC}, the following statement holds:
{\begin{align}
\mathbb{E}\left[\Phi (\bm{x}^{(t+1)} )\right] - \mathbb{E}\left[\Phi(\bm{x}^{(t)})\right] &\leq -\frac{\eta}{2} \mathbb{E}\left[\left\| \nabla \Phi(\bm{x}^{(t)}) \right\|^2\right] - \left(\eta_x - 3\beta \eta_x^2 \right)  \mathbb{E}\left[\left\| \frac{1}{n} \sum_{i=1}^n \nabla_x f_i(\bm{x}^{(t)}, \bm{y}^{(t)}) \right\|^2\right] \nonumber\\
&\quad + \left(2 \eta  +3\beta\eta_x^2\right)L^2 \mathbb{E}\left[\left(\delta_{\bm{x}}^{(t)} + \delta_{\bm{y}}^{(t)} \right)\right]
+\frac{\eta_x L^2}{2}  \mathbb{E}\left[\left\| \bm{y}^* (\bm{x}^{(t)}) - \bm{y}^{(t)}\right\|^2\right] + \frac{3}{2n}\beta \eta_x^2 \sigma^2,\nonumber
\end{align}}
where $\beta=L+ \kappa L$, and $\delta_{\bm{x}}^{(t)} = \frac{1}{n}\sum_{i=1}^n\left\|\bm{x}_i^{(t)}-\bm{x}^{(t)}\right\|^2, \quad \delta_{\bm{y}}^{(t)} = \frac{1}{n}\sum_{i=1}^n\left\|\bm{y}_i^{(t)}-\bm{y}^{(t)}\right\|^2$.
\begin{proof}
 According to~\cite{lin2019gradient}, $\Phi(\cdot)$ is $\beta=L+ \kappa L$-smooth, together with updating rule, so we have:
 {\begin{align}
     \Phi(\bm{x}^{(t+1)}) &\leq  \Phi(\bm{x}^{(t)}) + \left\langle \nabla \Phi(\bm{x}^{(t)}), \bm{x}^{(t+1)}-\bm{x}^{(t)} \right\rangle + \frac{\beta}{2}\left\| \bm{x}^{(t+1)}-\bm{x}^{(t)} \right\|^2\nonumber\\
    & \leq  \Phi(\bm{x}^{(t)}) - \eta_x\left\langle \nabla \Phi(\bm{x}^{(t)}), \frac{1}{n} \sum_{i=1}^n \nabla_x f_i(\bm{x}_i^{(t)}, \bm{y}_i^{(t)};\xi_i^t) \right\rangle + \frac{\beta}{2}\eta^2\left\| \frac{1}{n} \sum_{i=1}^n \nabla_x f_i(\bm{x}_i^{(t)}, \bm{y}_i^{(t)};\xi_i^t) \right\|^2.\nonumber
 \end{align}}
 Taking expectation on both sides yields:
  {\begin{align}
      \mathbb{E}\left[\Phi(\bm{x}^{(t+1)})\right] &\leq  \mathbb{E}\left[\Phi(\bm{x}^{(t)})\right] -  \eta_x\left\langle \nabla \Phi(\bm{x}^{(t)}),\frac{1}{n} \sum_{i=1}^n \nabla_x f_i(\bm{x}_i^{(t)}, \bm{y}_i^{(t)} ) \right\rangle + \frac{\beta}{2}\eta_x^2\mathbb{E}\left[\left\| \frac{1}{n} \sum_{i=1}^n \nabla_x f_i(\bm{x}_i^{(t)}, \bm{y}_i^{(t)};\xi_i^t) \right\|^2\right] \nonumber \\
    &  \leq  \mathbb{E}\left[\Phi(\bm{x}^{(t)})\right] - \eta_x\left\langle \nabla \Phi(\bm{x}^{(t)}), \frac{1}{n} \sum_{i=1}^n \nabla_x f_i(\bm{x}^{(t)}, \bm{y}^{(t)} ) \right\rangle+  \frac{\beta}{2}\eta_x^2\mathbb{E}\left[\left\| \frac{1}{n} \sum_{i=1}^n \nabla_x f_i(\bm{x}_i^{(t)}, \bm{y}_i^{(t)};\xi_i^t) \right\|^2\right] \nonumber\\
    & \quad - \eta_x\left\langle \nabla \Phi(\bm{x}^{(t)}), \frac{1}{n} \sum_{i=1}^n \nabla_x f_i(\bm{x}_i^{(t)}, \bm{y}_i^{(t)} ) - \frac{1}{n} \sum_{i=1}^n \nabla_x f_i(\bm{x}^{(t)}, \bm{y}^{(t)} )\right\rangle \nonumber.
 \end{align}}
 Using the identity $\langle \bm{a},\bm{b}\rangle = -\frac{1}{2}\|\bm{a} - \bm{b}\|^2 + \frac{1}{2}\|\bm{a}\|^2 + \frac{1}{2}\|\bm{b}\|^2 $, we have:
   {\begin{align}
     & \mathbb{E}\left[\Phi(\bm{x}^{(t+1)})\right]-\mathbb{E}\left[\Phi(\bm{x}^{(t)})\right]  \nonumber\\
     & \leq   - \frac{\eta_x}{2}\mathbb{E}\left[\left\| \nabla \Phi(\bm{x}^{(t)})\right\|^2\right] - \frac{\eta_x}{2}\mathbb{E}\left[\left\|\frac{1}{n} \sum_{i=1}^n \nabla_x f_i(\bm{x}^{(t)}, \bm{y}^{(t)} ) \right\|^2\right] + \frac{\eta_x}{2}\mathbb{E}\left[\left\| \nabla \Phi(\bm{x}^{(t)}) - \frac{1}{n} \sum_{i=1}^n \nabla_x f_i(\bm{x}^{(t)}, \bm{y}^{(t)} ) \right\|^2\right] \nonumber\\
    & \quad + \frac{\eta_x}{2}\left(\frac{1}{2}\mathbb{E}\left[\left\| \nabla \Phi(\bm{x}^{(t)})\right\|^2\right] +2\mathbb{E}\left[\left\|\frac{1}{n} \sum_{i=1}^n \nabla_x f_i(\bm{x}_i^{(t)}, \bm{y}_i^{(t)} ) - \frac{1}{n} \sum_{i=1}^n \nabla_x f_i(\bm{x}^{(t)}, \bm{y}^{(t)} )\right\|^2\right]\right)\nonumber\\
    & \quad +  \frac{\beta}{2}\eta_x^2\mathbb{E}\left[\left\| \frac{1}{n} \sum_{i=1}^n \nabla_x f_i(\bm{x}_i^{(t)}, \bm{y}_i^{(t)};\xi_i^t) \right\|^2\right]\nonumber\\
    & \leq   - \frac{\eta_x}{4}\mathbb{E}\left[\left\| \nabla \Phi(\bm{x}^{(t)})\right\|^2\right] - \frac{\eta_x}{2}\mathbb{E}\left[\left\|\frac{1}{n} \sum_{i=1}^n \nabla_x f_i(\bm{x}^{(t)}, \bm{y}^{(t)} ) \right\|^2\right] + \frac{\eta_x L^2}{2}\mathbb{E}\left[\left\| \bm{y}^*(\bm{x}^{(t)}) -\bm{y}^{(t)}   \right\|^2\right] \nonumber\\
    &  \quad +  \eta_x L^2  \frac{1}{n} \sum_{i=1}^n  \mathbb{E}\left[2\left\|  \bm{x}_i^{(t)}  -   \bm{x}^{(t)}   \right\|^2 + 2\left\|  \bm{y}_i^{(t)}  -  \bm{y}^{(t)}  \right\|^2\right] \nonumber\\
    &  \quad +   \frac{\beta}{2} \eta_x^2\mathbb{E}\left[3\left\| \frac{1}{n} \sum_{i=1}^n \nabla_x f_i(\bm{x}^{(t)}, \bm{y}^{(t)} ) \right\|^2+ 3\left\|\frac{1}{n} \sum_{i=1}^n \nabla_x f_i(\bm{x}_i^{(t)}, \bm{y}_i^{(t)} )- \frac{1}{n} \sum_{i=1}^n \nabla_x f_i(\bm{x}^{(t)}, \bm{y}^{(t)} ) \right\|^2 + 3\sigma^2\right]\nonumber\\
     & \leq   - \frac{\eta_x}{4}\mathbb{E}\left[\left\| \nabla \Phi(\bm{x}^{(t)})\right\|^2\right] - \left(\frac{\eta}{2}-\frac{3\beta}{2} \eta^2\right)\mathbb{E}\left[\left\|\frac{1}{n} \sum_{i=1}^n \nabla_x f_i(\bm{x}^{(t)}, \bm{y}^{(t)} ) \right\|^2\right] + \frac{\eta_x L^2}{2}\mathbb{E}\left[\left\| \bm{y}^*(\bm{x}^{(t)}) -\bm{y}^{(t)}   \right\|^2\right] \nonumber\\
    &  \quad  +  (2\eta_x+3\beta\eta_x^2) L^2  \mathbb{E}\left[ \delta_{\bm{x}}^{(t)} + \delta_{\bm{y}}^{(t)}\right]  +   \frac{3\beta}{2n} \eta_x^2 \sigma^2 .\nonumber
 \end{align}}
\end{proof}
\end{lemma}

The following lemma characterizes the local model deviation bound for nonconvex-strongly-concave function.
 \begin{lemma}\label{lm5}

For local-SGDA, under assumptions of Theorem~\ref{Thm: NCSC}, the following statement holds true:
\begin{align} \frac{1}{T}\sum_{t=1}^{T}\frac{1}{n}\sum_{i=1}^n\mathbb{E}\left[\left\|\bm{x}^{(t)}-\bm{x}^{(t)}_i \right\|^2 \right]+\mathbb{E}\left[\left\|\bm{y}^{(t)}-\bm{y}^{(t)}_i \right\|^2 \right]   \leq  10\tau^2 (\eta_x^2 + \eta_y^2) \left( \sigma^2 + \frac{\sigma^2}{n}  \right)+10\tau^2\eta_x^2\zeta_x+ 10\tau^2 \eta_y^2 \zeta_y  .\nonumber
\end{align}
\begin{proof}
We start to prove the first statement here.
 For the simplicity of notations, we define $\delta^t = \mathbb{E}[\delta^t_x + \delta^t_y] = \frac{1}{n}\sum_{i=1}^n\mathbb{E}\left[\left\|\bm{x}^{(t)}-\bm{x}^{(t)}_i \right\|^2\right] +\mathbb{E}\left[\left\|\bm{y}^{(t)}-\bm{y}^{(t)}_i \right\|^2\right]  $. Assume $s\tau+1 \leq t \leq (s+1)\tau$. Notice that:

 \begin{align}
 \delta^t &= \frac{1}{n}\sum_{i=1}^n\mathbb{E}\left[\left\| \bm{x}^{s\tau} - \sum_{j=s\tau}^{(s+1)\tau}\frac{\eta_x}{n}\sum_{k=1}^n \nabla_x f_k(\bm{x}_k^{(j)},\bm{y}_k^{(j)};\xi_k^j)  -\left(\bm{x}^{s\tau}  -\sum_{j=s\tau}^{(s+1)\tau}\eta_x \nabla_x f_i(\bm{x}_i^{(j)},\bm{y}_i^{(j)};\xi_i^j) \right) \right\|^2\right]\nonumber\\
 & \quad +\frac{1}{n}\sum_{i=1}^n\mathbb{E}\left[\left\| \bm{y}^{s\tau} - \sum_{j=s\tau}^{(s+1)\tau}\frac{\eta_y}{n}\sum_{k=1}^n \nabla_y f_k(\bm{x}_k^{(j)} ,\bm{y}_k^{(j)};\xi_k^j)  -\left(\bm{y}^{s\tau}  -\sum_{j=s\tau}^{(s+1)\tau}\eta_y \nabla_y f_i(\bm{x}_i^{(j)},\bm{y}_i^{(j)};\xi_i^j) \right) \right\|^2\right]\nonumber\\
 & =\tau \sum_{j=s\tau}^{(s+1)\tau}\frac{\eta_x^2}{n}\sum_{i=1}^n\mathbb{E}\left[\left\|   \frac{1}{n}\sum_{k=1}^n \nabla_x f_k(\bm{x}_k^{(j)},\bm{y}_k^{(j)};\xi_k^j)  -  \nabla_x f_i(\bm{x}_i^{(j)},\bm{y}_i^{(j)};\xi_i^j)  \right\|^2\right]\nonumber\\
 & \quad + \tau\sum_{j=s\tau}^{(s+1)\tau}\frac{\eta_y}{n}\sum_{i=1}^n\mathbb{E}\left[\left\|   \frac{1}{n}\sum_{k=1}^n \nabla_y f_k(\bm{x}_k^{(j)},\bm{y}_k^{(j)};\xi_k^j)  -  \nabla_y f_i(\bm{x}_i^{(j)},\bm{y}_i^{(j)};\xi_i^j)  \right\|^2\right]\nonumber\\
  &  = \tau \sum_{j=s\tau}^{(s+1)\tau}\frac{\eta_x^2}{n} \sum_{i=1}^n\mathbb{E}\left[\left\|   \frac{1}{n}\sum_{k=1}^n \nabla_x f_k(\bm{x}_k^{(j)},\bm{y}_k^{(j)};\xi_k^j)  - \nabla_x f_k(\bm{x}_k^{(j)},\bm{y}_k^{(j)} ) +\nabla_x f_k(\bm{x}_k^{(j)},\bm{y}_k^{(j)} ) -\nabla_x f_k(\bm{x}^{(j)},\bm{y}^{(j)} )  \right.\right. \nonumber\\
  & \quad \left.\left. + \nabla_x f_k(\bm{x}^{(j)},\bm{y}^{(j)} ) -\nabla_x f_i(\bm{x}^{(j)},\bm{y}^{(j)} )+\nabla_x f_i(\bm{x}^{(j)},\bm{y}^{(j)} ) -\nabla_x f_i(\bm{x}_i^{(j)},\bm{y}_i^{(j)} ) +\nabla_x f_i(\bm{x}_i^{(j)},\bm{y}_i^{(j)} ) - \nabla_x f_i(\bm{x}_i^{(j)},\bm{y}_i^{(j)};\xi_i^t )  \right\|^2\right] \nonumber\\
  &\quad +\tau\sum_{j=s\tau}^{(s+1)\tau}\frac{\eta_y^2}{n}  \sum_{i=1}^n\mathbb{E}\left[\left\|   \frac{1}{n}\sum_{k=1}^n \nabla_y f_k(\bm{x}_k^{(j)},\bm{y}_k^{(j)};\xi_k^j)  - \nabla_y f_k(\bm{x}_k^{(j)},\bm{y}_k^{(j)} ) +\nabla_y f_k(\bm{x}_k^{(j)},\bm{y}_k^{(j)} ) -\nabla_y f_k(\bm{x}^{(j)},\bm{y}^{(j)} )  \right.\right. \nonumber\\
  &\quad  \left.\left. + \nabla_y f_k(\bm{x}^{(j)},\bm{y}^{(j)} )-\nabla_y f_i(\bm{x}^{(j)},\bm{y}^{(j)} )+\nabla_y f_i(\bm{x}^{(j)},\bm{y}^{(j)} ) -\nabla_y f_i(\bm{x}^{(j)}_i,\bm{y}_i^{(j)} ) +\nabla_y f_i(\bm{x}^{(j)}_i,\bm{y}_i^{(j)} ) - \nabla_y f_i(\bm{x}^{(j)}_i,\bm{y}_i^{(j)};\xi_i^j )  \right\|^2\right] \nonumber\\
  &  \leq \sum_{j=s\tau}^{(s+1)\tau}5\eta_x^2 \left( \sigma^2 + \frac{\sigma^2}{n} +2L^2\delta^j + \zeta_x \right) + 5\eta_y^2 \left( \sigma^2 + \frac{\sigma^2}{n} +2L^2\delta^j + \zeta_y \right). \nonumber  
 \end{align}
 Summing over $t$ from $s\tau$ to $(s+1)\tau$ yields:
 \begin{align}
  \sum_{t=s\tau}^{(s+1)\tau}  \delta^t &\leq   \sum_{t=s\tau}^{(s+1)\tau}\sum_{j=s\tau}^{(s+1)\tau}5\tau\eta_x^2 \left( \sigma^2 + \frac{\sigma^2}{n} +2L^2\delta^j + \zeta_x \right) + 5\tau\eta_y^2 \left( \sigma^2 + \frac{\sigma^2}{n} +2L^2\delta^j + \zeta_y \right)\nonumber\\
 & \leq  10L^2   \tau^2    (\eta_x^2 + \eta_y^2)\sum_{j=s\tau}^{(s+1)\tau} \delta^j + 5\tau^2 (\eta_x^2 + \eta_y^2) \left( \sigma^2 + \frac{\sigma^2}{n}  \right)+5 \tau^2\eta_x^2\zeta_x+ 5 \tau^2 \eta_y^2 \zeta_y.  
 \end{align}
 Since $\tau = \frac{T^{1/3}}{n^{1/3}}, \eta_x = \frac{n^{1/3}}{LT^{2/3}}, \eta_y = \frac{2}{LT^{1/2}}$ and $T \geq \max\left\{\frac{160^3}{n^2}, 40^{3/2}\right\}$, then $10L^2   \tau^2    (\eta_x^2 + \eta_y^2)\leq \frac{1}{2}$, by re-arranging the terms we have:
  \begin{align}
  \sum_{t=s\tau+1}^{(s+1)\tau}  \delta^t  \leq   10\tau^3 (\eta_x^2 + \eta_y^2) \left( \sigma^2 + \frac{\sigma^2}{n}  \right)+10\tau^3\eta_x^2\zeta_x+ 10\tau^3 \eta_y^2 \zeta_y .\nonumber
 \end{align}
 Summing over $s$ from $0$ to $T/\tau - 1$, and dividing both sides by $T$ can conclude the proof of the first statement:
  \begin{align}
 & \frac{1}{T}\sum_{t=1}^{T}  \delta^t   \leq   10\tau^2 (\eta_x^2 + \eta_y^2) \left( \sigma^2 + \frac{\sigma^2}{n}  \right)+10\tau^2\eta_x^2\zeta_x+ 10\tau^2 \eta_y^2 \zeta_y  .\nonumber
 \end{align} 
\end{proof}
\end{lemma}

The next lemma establishes an upper bound on the  dual optimality gap.
\begin{lemma}\label{lm6}
For local-SGDA, if we choose $\eta_y = \frac{2}{CL}$, then under assumptions of Theorem~\ref{Thm: NCSC}, the gap between $\bm{y}^{t}$ and $\bm{y}^*(\bm{x}^{(t)})$ can be bounded as follows:
{\begin{align}
      \frac{1}{T}\sum_{t=1}^{T} \mathbb{E}\left[\left\|  \bm{y}^{(t)} - \bm{y}^* (\bm{x}^{(t)})\right\|^2 \right]   &\leq \frac{2C\kappa}{T}   \mathbb{E}\left[\left\| \bm{y}^{(0)} -  \bm{y}^* (\bm{x}^{(0)}) \right\|^2 \right]+ 2C\kappa\left(1+\frac{1}{2(C\kappa-1)}\right)\frac{4\eta_y^2\sigma^2}{n} \nonumber\\
     & \quad + \frac{1}{T}\sum_{t=1}^T2C\kappa\left(1+\frac{1}{2(C\kappa-1)}\right)  \left(\frac{4\eta_y L^2}{\mu} + 8\eta_y^2L^2\right)\mathbb{E}\left[  \delta_{\bm{x}}^{(t)} + \delta_{\bm{y}}^{(t)} \right]\nonumber\\
     & \quad + \frac{1}{T}\sum_{t=1}^T4C^2 \kappa^4 \eta_x^2\mathbb{E}\left[ 3\left\|\frac{1}{n}\sum_{i=1}^n \nabla_x f_i(\bm{x} ^{(t)},\bm{y} ^{(t)})\right\|^2 +6 L^2(\delta_{\bm{x}}^{(t)}+ \delta_{\bm{y}}^{(t)})  + \frac{3\sigma^2 }{n}\right].  
\end{align}} 
where $\delta_{\bm{x}}^{(t)} = \frac{1}{n}\sum_{i=1}^n\left\|\bm{x}_i^{(t)}-\bm{x}^{(t)}\right\|^2$ and $  \delta_{\bm{y}}^{(t)} = \frac{1}{n}\sum_{i=1}^n\left\|\bm{y}_i^{(t)}-\bm{y}^{(t)}\right\|^2$.
\begin{proof}
 According to arithmetic and geometric inequality and Cauchy's inequality: $\|\bm{a}+\bm{b}\|^2 \leq  \|\bm{a}\|^2+2\|\bm{a}\| \|\bm{b}\|+\|\bm{b}\|^2 \leq  \left(1+\frac{1}{q}\right)\|\bm{a}\|^2 +\left(1+q\right)\|\bm{b}\|^2$,  we have:
 \begin{align}
     \mathbb{E}\left[\left\| \bm{y}^* (\bm{x}^{(t)}) - \bm{y}^{(t)}\right\|^2\right]  \leq \left(1+\frac{1}{2(C\kappa-1)}\right) \mathbb{E}\left[\left\| \bm{y}^* (\bm{x}^{(t-1)}) - \bm{y}^{(t)}\right\|^2 \right] + \left(1+2(C\kappa-1)\right)\mathbb{E}\left[ \left\| \bm{y}^* (\bm{x}^{(t)}) -  \bm{y}^* (\bm{x}^{(t-1)})\right\|^2\right] .\nonumber
 \end{align}
 Then we are going to bound $\left\| \bm{y}^* (\bm{x}^{(t-1)}) - \bm{y}^{(t)}\right\|^2 $ and $ \left\| \bm{y}^* (\bm{x}^{(t)}) -  \bm{y}^* (\bm{x}^{(t-1)})\right\|^2$ separately.
 
 First, according to updating rule for $\bm{y}$ and strong concavity, we have:
 {\begin{align}
    &  \mathbb{E}\left[\left\| \bm{y}^{(t)} -  \bm{y}^* (\bm{x}^{(t-1)}) \right\|^2 \right]\nonumber\\
    &  = \mathbb{E}\left[\left\| \bm{y}^{(t-1)} +\eta_y \frac{1}{n}\sum_{i=1}^n \nabla_y f_i(\bm{x}_i^{(t-1)}, \bm{y}_i^{(t-1)}; \xi_i^t) -  \bm{y}^* (\bm{x}^{(t-1)}) \right\|^2\right] \nonumber\\
     &  \leq \mathbb{E}\left[\left\| \bm{y}^{(t-1)} -  \bm{y}^* (\bm{x}^{(t-1)}) \right\|^2 \right]  + \eta_y^2 \mathbb{E}\left[\left\| \frac{1}{n}\sum_{i=1}^n \nabla_y f_i(\bm{x}_i^{(t-1)}, \bm{y}_i^{(t-1)}; \xi_i^t)  \right\|^2 \right]\nonumber\\
     & \quad  + 2\eta_y\mathbb{E}\left[ \left\langle  \frac{1}{n}\sum_{i=1}^n \nabla_y f_i(\bm{x}_i^{(t-1)}, \bm{y}_i^{(t-1)} ),  \bm{y}^{(t-1)} -  \bm{y}^* (\bm{x}^{(t-1)}) \right\rangle \right]\nonumber\\
     & \leq \mathbb{E}\left[\left\| \bm{y}^{(t-1)} -  \bm{y}^* (\bm{x}^{(t-1)}) \right\|^2 \right] \nonumber\\
    & \quad + \eta_y^2  \left(4\underbrace{\mathbb{E}\left[\left\|  \nabla_y F(\bm{x}^{(t-1)}, \bm{y}^* (\bm{x}^{(t-1)}) )  \right\|^2 \right]}_{=0}+4\mathbb{E}\left[\left\|  \nabla_y F(\bm{x}^{(t-1)},  \bm{y}^{(t-1)} ) -  \nabla_y F(\bm{x}^{(t-1)}, \bm{y}^* (\bm{x}^{(t-1)}) )\right\|^2 \right]  \right) \nonumber\\
    & \quad + \eta_y^2  \frac{1}{n}\sum_{i=1}^n\left(4\mathbb{E}\left[\left\|   \nabla_y f_i(\bm{x}^{(t-1)},  \bm{y}^{(t-1)}  ) -  \nabla_y f_i(\bm{x}_i^{(t-1)},  \bm{y}_i^{(t-1)}  ) \right\|^2 \right]+4 \frac{\sigma^2}{n}  \right) \nonumber\\
     & \quad + 2\eta_y\mathbb{E}\left[ \left\langle  \frac{1}{n}\sum_{i=1}^n \nabla_y f_i(\bm{x}^{(t-1)}, \bm{y}^{(t-1)} ),  \bm{y}^{(t-1)} -  \bm{y}^* (\bm{x}^{(t-1)}) \right\rangle \right]\nonumber\\
       & \quad + 2\eta_y\mathbb{E}\left[ \left\langle  \frac{1}{n}\sum_{i=1}^n \nabla_y f_i(\bm{x}_i^{(t-1)}, \bm{y}_i^{(t-1)} )-\nabla_x f_i(\bm{x}^{(t-1)}, \bm{y}^{(t-1)} ),  \bm{y}^{(t-1)} -  \bm{y}^* (\bm{x}^{(t-1)}) \right\rangle \right]\nonumber\\
        & \leq \left(1-\mu\eta_y \right) \mathbb{E}\left[\left\| \bm{y}^{(t-1)} -  \bm{y}^* (\bm{x}^{(t-1)}) \right\|^2 \right]  + 4\eta_y^2   \frac{\sigma^2}{n} + \frac{\mu\eta_y}{2} \mathbb{E}\left[\left\| \bm{y}^{(t-1)} -  \bm{y}^* (\bm{x}^{(t-1)}) \right\|^2\right]  \nonumber\\
     & \quad + 2\underbrace{(\eta_y-4\eta_y^2L)}_{\geq 0}\underbrace{\mathbb{E}\left[    F(\bm{x}^{(t-1)}, \bm{y}^{(t-1)} ) - F(\bm{x}^{(t-1)}, \bm{y}^* (\bm{x}^{(t-1)})) \right]}_{\leq 0} \nonumber\\
       & \quad  +  \left(\frac{2\eta_y}{\mu}+4\eta_y^2\right)\mathbb{E}\left[ \frac{1}{n}\sum_{i=1}^n\left\|  \nabla_y f_i(\bm{x}_i^{(t-1)}, \bm{y}_i^{(t-1)} )-\nabla_y f_i(\bm{x}^{(t-1)}, \bm{y}^{(t-1)} )\right\|^2\right]\nonumber  \\
           &  \leq \left(1-\frac{\mu\eta_y}{2} \right) \mathbb{E}\left[\left\| \bm{y}^{(t-1)} -  \bm{y}^* (\bm{x}^{(t-1)}) \right\|^2 \right]  +    \frac{4\eta_y^2\sigma^2}{n}+  \left(\frac{4\eta_y L^2}{\mu} + 8\eta_y^2L^2\right)\mathbb{E}\left[  \delta_{\bm{x}}^{(t-1)} + \delta_{\bm{y}}^{(t-1)} \right]. \label{eq: lm7 1} 
 \end{align}}
 Then, for the term $ \left\| \bm{y}^* (\bm{x}^{(t)}) -  \bm{y}^* (\bm{x}^{(t-1)})\right\|^2 $, since $\bm{y}^*(\cdot)$ is $\kappa$-Lipschitz, we have:
 {\begin{align}
     &\mathbb{E}\left[ \left\| \bm{y}^* (\bm{x}^{(t)}) -  \bm{y}^* (\bm{x}^{(t-1)})\right\|^2\right]  \leq \kappa^2 \mathbb{E}\left[\|\bm{x}^{(t)} - \bm{x}^{(t-1)}\|^2 \right]\nonumber\\
    &  = \kappa^2 \eta_x^2 \mathbb{E}\left[\left\|\frac{1}{n}\sum_{i=1}^n \nabla_x f_i(\bm{x}_i^{(t-1)},\bm{y}_i^{(t-1)};\xi_i^t)\right\|^2 \right]\nonumber\\
     & \leq \kappa^2 \eta_x^2\mathbb{E}\left[ 3\left\|\frac{1}{n}\sum_{i=1}^n \nabla_x f_i(\bm{x} ^{(t-1)},\bm{y} ^{(t-1)})\right\|^2 +3 \frac{1}{n}\sum_{i=1}^n\left\| \nabla_x f_i(\bm{x}_i^{(t-1)},\bm{y}_i^{(t-1)}) - \nabla_x f_i(\bm{x} ^{(t-1)},\bm{y} ^{(t-1)})\right\|^2  + \frac{3\sigma^2}{n} \right] \nonumber\\
      & \leq \kappa^2 \eta_x^2\mathbb{E}\left[ 3\left\|\frac{1}{n}\sum_{i=1}^n \nabla_x f_i(\bm{x} ^{(t-1)},\bm{y} ^{(t-1)})\right\|^2 +6 L^2(\delta_{\bm{x}}^{(t-1)}+ \delta_{\bm{y}}^{(t-1)})  + \frac{3\sigma^2}{n} \right]. \label{eq: lm7 2} \nonumber\\
\end{align} }
Recall that we choose $\eta_y = \frac{2}{CL}$, $C >  0$. 
Combining (\ref{eq: lm7 1}) and (\ref{eq: lm7 2}) yields:
\begin{align}
      &\mathbb{E}\left[\left\| \bm{y}^* (\bm{x}^{(t)}) - \bm{y}^{(t)}\right\|^2\right] \nonumber\\ 
     &  \leq \left(1+\frac{1}{2(C\kappa-1)}\right) \left( \left(1-\frac{\mu\eta_y}{2} \right) \mathbb{E}\left[\left\| \bm{y}^{(t-1)} -  \bm{y}^* (\bm{x}^{(t-1)}) \right\|^2 \right]  +    \frac{4\eta_y^2\sigma^2}{n}+  \left(\frac{4\eta_y L^2}{\mu} + 8\eta_y^2L^2\right)\mathbb{E}\left[  \delta_{\bm{x}}^{(t-1)} + \delta_{\bm{y}}^{(t-1)} \right]\right)\nonumber\\
     &  \quad + \left(1+2(C\kappa-1)\right)\kappa^2 \eta_x^2\mathbb{E}\left[ 3\left\|\frac{1}{n}\sum_{i=1}^n \nabla_x f_i(\bm{x} ^{(t-1)},\bm{y} ^{(t-1)})\right\|^2 +6 L^2(\delta_{\bm{x}}^{(t-1)}+ \delta_{\bm{y}}^{(t-1)})  + \frac{3\sigma^2}{n} \right]\nonumber\\
      &  \leq   \left(1+\frac{1}{2(C\kappa-1)}\right) \left(1-\frac{1}{C\kappa} \right) \mathbb{E}\left[\left\| \bm{y}^{(t-1)} -  \bm{y}^* (\bm{x}^{(t-1)})\right\|^2 \right]\nonumber\\
      & \quad  + \left(1+\frac{1}{2(C\kappa-1)}\right) \left(\frac{4\eta_y^2\sigma^2}{n}+  \left(\frac{4\eta_y L^2}{\mu} + 8\eta_y^2L^2\right)\mathbb{E}\left[  \delta_{\bm{x}}^{(t-1)} + \delta_{\bm{y}}^{(t-1)} \right] \right)\nonumber\\
     & \quad + \left(1+2(C\kappa-1)\right)\kappa^2 \eta_x^2\mathbb{E}\left[ 3\left\|\frac{1}{n}\sum_{i=1}^n \nabla_x f_i(\bm{x} ^{(t-1)},\bm{y} ^{(t-1)})\right\|^2 +6 L^2(\delta_{\bm{x}}^{(t-1)}+ \delta_{\bm{y}}^{(t-1)})  + \frac{3\sigma^2}{n} \right].\nonumber
\end{align}
Using the fact $\left(1+\frac{1}{2(C\kappa-1)}\right) \left(1-\frac{1}{C\kappa} \right) = \left(1-\frac{1}{2C\kappa }\right) $, and unrolling the recursion yields:
\begin{align}
     & \mathbb{E}\left[\left\| \bm{y}^* (\bm{x}^{(t)}) - \bm{y}^{(t)}\right\|^2\right] \nonumber\\ 
      & \leq  \left(1-\frac{1}{2C\kappa }\right)  \mathbb{E}\left[\left\| \bm{y}^{(t-1)} -  \bm{y}^* (\bm{x}^{(t-1)}) \right\|^2 \right]  + \left(1+\frac{1}{2(C\kappa-1)}\right) \left(\frac{4\eta_y^2\sigma^2}{n}+  \left(\frac{4\eta_y L^2}{\mu} + 8\eta_y^2L^2\right)\mathbb{E}\left[  \delta_{\bm{x}}^{(t-1)} + \delta_{\bm{y}}^{(t-1)} \right]\right)\nonumber\\
     &  \quad + \left(1+2(C\kappa-1)\right)\kappa^2 \eta_x^2\mathbb{E}\left[ 3\left\|\frac{1}{n}\sum_{i=1}^n \nabla_x f_i(\bm{x} ^{(t-1)},\bm{y} ^{(t-1)})\right\|^2 +6 L^2(\delta_{\bm{x}}^{(t-1)}+ \delta_{\bm{y}}^{(t-1)})  + \frac{3\sigma^2}{n} \right]\nonumber\\
     &  \leq  \left(1-\frac{1}{2C\kappa }\right)^t  \mathbb{E}\left[\left\| \bm{y}^{(0)} -  \bm{y}^* (\bm{x}^{(0)}) \right\|^2 \right]\nonumber\\
     & \quad + \sum_{j=1}^t\left(1-\frac{1}{2C\kappa }\right)^{t-j}\left(1+\frac{1}{2(C\kappa-1)}\right) \left( \frac{4\eta_y^2\sigma^2}{n}+  \left(\frac{4\eta_y L^2}{\mu} + 8\eta_y^2L^2\right)\mathbb{E}\left[  \delta_{\bm{x}}^{(t-1)} + \delta_{\bm{y}}^{(t-1)} \right] \right)\nonumber\\
     & \quad + \sum_{j=1}^t\left(1-\frac{1}{2C\kappa }\right)^{t-j}\left(1+2(C\kappa-1)\right)\kappa^2 \eta_x^2\mathbb{E}\left[ 3\left\|\frac{1}{n}\sum_{i=1}^n \nabla_x f_i(\bm{x} ^{(j-1)},\bm{y} ^{(t-1)})\right\|^2 +6 L^2(\delta_x^{j-1}+ \delta_y^{j-1})  + \frac{3\sigma^2}{n} \right]\nonumber\\
      &  \leq  \left(1-\frac{1}{2C\kappa }\right)^t   \mathbb{E}\left[\left\| \bm{y}^{(0)} -  \bm{y}^* (\bm{x}^{(0)}) \right\|^2 \right] + 2C\kappa\left(1+\frac{1}{2(C\kappa-1)}\right)   \left(\frac{4\eta_y^2\sigma^2}{n} \right) \nonumber\\
     & \quad + \sum_{j=1}^t\left(1-\frac{1}{2C\kappa }\right)^{t-j}\left(1+\frac{1}{2(C\kappa-1)}\right) \left(    \left(\frac{4\eta_y L^2}{\mu} + 8\eta_y^2L^2\right)\mathbb{E}\left[  \delta_{\bm{x}}^{(t-1)} + \delta_{\bm{y}}^{(t-1)} \right] \right)\nonumber\\
     & \quad + \sum_{j=1}^t\left(1-\frac{1}{2C\kappa }\right)^{t-j}\left(1+2(C\kappa-1)\right)\kappa^2 \eta_x^2\mathbb{E}\left[ 3\left\|\frac{1}{n}\sum_{i=1}^n \nabla_x f_i(\bm{x} ^{(j-1)},\bm{y} ^{(j-1)})\right\|^2 +6 L^2(\delta_x^{j-1}+ \delta_y^{j-1})  + \frac{3\sigma^2}{n} \right].\nonumber 
\end{align}
Summing from $t$ = $1$ to $T$, and dividing by $T$ yields:
\begin{align}
     & \frac{1}{T}\sum_{t=1}^T\mathbb{E}\left[\left\| \bm{y}^* (\bm{x}^{(t)}) - \bm{y}^{(t)}\right\|^2\right] \nonumber\\ 
      &  \leq \frac{1}{T}\sum_{t=1}^T \left(1-\frac{1}{2C\kappa }\right)^t   \mathbb{E}\left[\left\| \bm{y}^{(0)} -  \bm{y}^* (\bm{x}^{(0)}) \right\|^2 \right]+ 2C\kappa\left(1+\frac{1}{2(C\kappa-1)}\right) \frac{4\eta_y^2\sigma^2}{n} \nonumber\\
     &   \quad + \frac{1}{T}\sum_{t=1}^T\sum_{j=1}^t\left(1-\frac{1}{2C\kappa }\right)^{t-j}\left(1+\frac{1}{2(C\kappa-1)}\right ) \left(\frac{4\eta_y L^2}{\mu} + 8\eta_y^2L^2\right)\mathbb{E}\left[  \delta_{\bm{x}}^{(t-1)} + \delta_{\bm{y}}^{(t-1)} \right] \nonumber\\
     & \quad + \frac{1}{T}\sum_{t=1}^T\sum_{j=1}^t\left(1-\frac{1}{2C\kappa }\right)^{t-j}\left(1+2(C\kappa-1)\right)\kappa^2 \eta_x^2\mathbb{E}\left[ 3\left\|\frac{1}{n}\sum_{i=1}^n \nabla_x f_i(\bm{x} ^{(j-1)},\bm{y} ^{(t-1)})\right\|^2 +6 L^2(\delta_x^{j-1}+ \delta_y^{j-1})  + \frac{3\sigma^2}{n} \right].\nonumber\\
     &  \leq \frac{2C\kappa}{T}   \mathbb{E}\left[\left\| \bm{y}^{(0)} -  \bm{y}^* (\bm{x}^{(0)}) \right\|^2 \right]+ 2C\kappa\left(1+\frac{1}{2(C\kappa-1)}\right)\frac{4\eta_y^2\sigma^2}{n} \nonumber \\
     &  \quad + \frac{1}{T}\sum_{t=0}^T2C\kappa\left(1+\frac{1}{2(C\kappa-1)}\right)  \left(\frac{4\eta_y L^2}{\mu} + 8\eta_y^2L^2\right)\mathbb{E}\left[  \delta_{\bm{x}}^{(t)} + \delta_{\bm{y}}^{(t)} \right]\nonumber\\
     & \quad + \frac{1}{T}\sum_{t=0}^T4C^2 \kappa^4 \eta_x^2\mathbb{E}\left[ 3\left\|\frac{1}{n}\sum_{i=1}^n \nabla_x f_i(\bm{x} ^{(t)},\bm{y} ^{(t)})\right\|^2 +6 L^2(\delta_{\bm{x}}^{(t)}+ \delta_{\bm{y}}^{(t)})  + \frac{3\sigma^2 }{n}\right].\nonumber 
\end{align}
 
\end{proof}

\end{lemma}

\subsection{Proof of Theorem~\ref{Thm: NCSC}}\label{sec:proof_thm_ncsc}
Now we provide the proof of Theorem~\ref{Thm: NCSC}.
In Lemma~\ref{lm4}, summing over $t$ = $1$ to $T$ and divding both sides by $T$ yields:
{\begin{align}
 &\frac{1}{T}\left(\mathbb{E}\left[\Phi (\bm{x}^{(T+1)} )\right] - \mathbb{E}\left[\Phi(\bm{x}^{(0)})\right]\right) \nonumber \\ &\leq -\frac{\eta_x}{2} \frac{1}{T}\sum_{t=1}^T\mathbb{E}\left[\left\| \nabla \Phi(\bm{x}^{(t)}) \right\|^2\right] - \left(\eta_x - 3\beta \eta_x^2 \right) \frac{1}{T}\sum_{t=1}^T \mathbb{E}\left[\left\| \frac{1}{n} \sum_{i=1}^n \nabla_x f_i(\bm{x}^{(t)}, \bm{y}^{(t)}) \right\|^2\right] \nonumber\\
& \quad + \left(2 \eta_x  +3\beta\eta_x^2\right)L^2\frac{1}{T}\sum_{t=1}^T \mathbb{E}\left[\left(\delta_{\bm{x}}^{(t)} + \delta_{\bm{y}}^{(t)} \right)\right]
+\frac{\eta_x L^2}{2} \frac{1}{T}\sum_{t=1}^T \mathbb{E}\left[\left\| \bm{y}^* (\bm{x}^{(t)}) - \bm{y}^{(t)}\right\|^2\right] + \frac{3}{2}\beta \eta_x^2 \frac{\sigma^2}{n}.\nonumber
\end{align}}
For the simplicity of the notation, we let $\Re = \frac{1}{T}\sum_{t=1}^T \mathbb{E}\left[\left\| \frac{1}{n} \sum_{i=1}^n \nabla_x f_i(\bm{x}^{(t)}, \bm{y}^{(t)}) \right\|^2\right]$. Re-arranging the terms and plugging in Lemma~\ref{lm5} and Lemma~\ref{lm6} gives:
{\begin{align}
& \frac{1}{T}\sum_{t=1}^T\mathbb{E}\left[\left\| \nabla \Phi(\bm{x}^{(t)}) \right\|^2\right] \nonumber  \\
& \leq  \frac{2}{\eta_x T}\mathbb{E}\left[\Phi(\bm{x}^{(0)})\right]- 2\left(1 - 3\beta \eta_x \right) \Re \nonumber\\
&  \quad + 2\left(2   +3\beta\eta_x\right)L^2\frac{1}{T}\sum_{t=1}^T \mathbb{E}\left[\left(\delta_{\bm{x}}^{(t)} + \delta_{\bm{y}}^{(t)} \right)\right]
+ L^2\frac{1}{T}\sum_{t=1}^T \mathbb{E}\left[\left\| \bm{y}^* (\bm{x}^{(t)}) - \bm{y}^{(t)}\right\|^2\right] + 3\beta \eta_x \frac{\sigma^2}{n} \nonumber\\
& \leq  \frac{2}{\eta_x T}\mathbb{E}\left[\Phi(\bm{x}^{(0)})\right]- 2\left(1 - 3\beta \eta_x \right) \Re + 3\beta \eta_x \frac{\sigma^2}{n} \nonumber\\
&  \quad + \left(4 +6\beta\eta_x\right)L^2  \left[10\tau^2 (\eta_x^2 + \eta_y^2) \left( \sigma^2 + \frac{\sigma^2}{n}  \right)+10\tau^2\eta_x^2\zeta_x+ 10\tau^2 \eta_y^2 \zeta_y  \right] \nonumber\\
&  \quad +  \frac{2L^2C\kappa }{T} \mathbb{E}\left[\left\| \bm{y}^{(0)} -  \bm{y}^* (\bm{x}^{(0)}) \right\|^2 \right]+ \left( \frac{2C^2\kappa^2L^2}{C\kappa-1}\right)  \frac{4\eta_y^2\sigma^2}{n}  \nonumber\\
&  \quad +  4C^2 \kappa^4\eta_x^2 L^2\left( \Re  + 6L^2\left[10\tau^2 (\eta_x^2 + \eta_y^2) \left( \sigma^2 + \frac{\sigma^2}{n}  \right)+10\tau^2\eta_x^2\zeta_x+ 10\tau^2 \eta_y^2 \zeta_y  \right] + \frac{3\sigma^2}{n}      \right) \nonumber\\
&  \quad +  \left( \frac{2C^2\kappa^2L^2}{C\kappa-1}\right)  \left(\frac{4\eta_y L^2}{\mu} + 8\eta_y^2L^2\right)\left[10\tau^2 (\eta_x^2 + \eta_y^2) \left( \sigma^2 + \frac{\sigma^2}{n}  \right)+10\tau^2\eta_x^2\zeta_x+ 10\tau^2 \eta_y^2 \zeta_y  \right]. \nonumber\\ 
& \leq  \frac{2}{\eta_x T}\mathbb{E}\left[\Phi(\bm{x}^{(0)})\right] +  \frac{2L^2C\kappa }{T} \mathbb{E}\left[\left\| \bm{y}^{(0)} -  \bm{y}^* (\bm{x}^{(0)}) \right\|^2 \right] - 2\left(1 - 3\beta \eta_x-4C^2 \kappa^4\eta_x^2 L^2 \right) \Re  \nonumber\\
&  \quad + 10\left(4   +6\beta\eta_x   + 24C^2 \kappa^4\eta_x^2 L^2 +\left( \frac{2C^2\kappa^2 }{C\kappa-1}\right)  \left(\frac{4\eta_y L^2}{\mu} + 8\eta_y^2L^2\right)  \right)L^2  \left[\tau^2 (\eta_x^2 + \eta_y^2) \left( \sigma^2 + \frac{\sigma^2}{n}  \right)+\tau^2\eta_x^2\zeta_x+ \tau^2 \eta_y^2 \zeta_y  \right] \nonumber\\ 
&  \quad +    \frac{12C^2 \kappa^4\eta_x^2 L^2\sigma^2}{n}    + 3\beta \eta_x \frac{\sigma^2}{n} + \left( \frac{2C^2\kappa^2L^2}{C\kappa-1}\right)  \frac{4\eta_y^2\sigma^2}{n}. \nonumber 
\end{align}}
By choosing  $\eta_x = \frac{n^{1/3}}{LT^{2/3}},  C = T^{1/2}$ and $T \geq \max\left\{\left(\frac{16 n^{4/3}\kappa^4 + \sqrt{16 n^{4/3}\kappa^8 - 12 \beta n^{1/3}/L}}{2}\right)^3,40^{3/2}, \frac{160^3}{n^2}   \right\}$ in Theorem~\ref{Thm: NCSC} such that
{\begin{equation}
    1 - 3\beta \eta_x-4C^2 \kappa^4\eta_x^2 L^2 \geq 0,\nonumber
\end{equation}}
holds, then we have:
\begin{align}
& \frac{1}{T}\sum_{t=1}^T\mathbb{E}\left[\left\| \nabla \Phi(\bm{x}^{(t)}) \right\|^2\right]   \leq  \frac{2}{\eta_x T}\mathbb{E}\left[\Phi(\bm{x}^{(0)})\right] +  \frac{2L^2C\kappa }{T} \mathbb{E}\left[\left\| \bm{y}^{(0)} -  \bm{y}^* (\bm{x}^{(0)}) \right\|^2 \right]  \nonumber\\
&  + 10\left(4   +6\beta\eta_x   + 24C^2 \kappa^4\eta_x^2 L^2 +\left( \frac{2C^2\kappa^2 }{C\kappa-1}\right)  \left(\frac{4\eta_y L^2}{\mu} + 8\eta_y^2L^2\right)  \right)L^2  \left[\tau^2 (\eta_x^2 + \eta_y^2) \left( \sigma^2 + \frac{\sigma^2}{n}  \right)+\tau^2\eta_x^2\zeta_x+ \tau^2 \eta_y^2 \zeta_y  \right] \nonumber\\ 
&  +    \frac{12C^2 \kappa^4\eta_x^2 L^2\sigma^2}{n}    +  \frac{3\beta \eta_x\sigma^2}{n} + \left( \frac{2C^2\kappa^2L^2}{C\kappa-1}\right)  \frac{4\eta_y^2\sigma^2}{n}. \nonumber 
\end{align}
Plugging in  $\tau = \frac{T^{1/3}}{n^{1/3}}$ and   $\eta_x = \frac{n^{1/3}}{ LT^{2/3}} $ ,$\eta_y = \frac{2}{LT^{\frac{1}{2}}}$, will conclude the proof:
\begin{align}
& \frac{1}{T}\sum_{t=1}^T\mathbb{E}\left[\left\| \nabla \Phi(\bm{x}^{(t)}) \right\|^2\right]   \leq O\left(\frac{L}{(nT)^{1/3}} + \frac{\kappa^4 L^2\sigma^2}{(nT)^{1/3}} + \frac{L^2\zeta_x}{T^{2/3}}+ \frac{L^2\zeta_y}{n^{2/3}T^{1/3}} + \frac{L^2\kappa}{T^{1/2}}\right).\nonumber
\end{align}

\qed

\section{Proof of Local SGDA+ under Nonconvex-PL Setting}\label{sec:ncnc}

\subsection{Overview of proofs}
Now we proceed to the proof of convergence rate in nonconvex-PL setting. In this case we still study the envelope function $\Phi(\cdot)$. The following proposition establishes the smoothness property of these auxiliary functions.
\begin{proposition}[Nouiehed et al~\cite{nouiehed2019solving}]
If a function $F(\bm{x},\cdot)$ satisfies $\mu$-PL condition and $L$ smooth, then $\Phi(\bm{x})$ is $\beta=\kappa L/2 + L$ smooth where $\kappa = L/\mu$.
\end{proposition}

Since $\Phi$ is $\beta$-smooth, then the starting point is similar to what we did in nonconvex-strongly-concave case, to conduct the one iteration analysis scheme for nonconvex smooth function on one iteration as follows:
\begin{align}
\mathbb{E}\left[\Phi (\bm{x}^{(t+1)} )\right] - \mathbb{E}\left[\Phi(\bm{x}^{(t)})\right]  
    & \leq   - \frac{\eta_x}{2}\mathbb{E}\left[\left\| \nabla \Phi(\bm{x}^{(t)})\right\|^2\right] - \left(\frac{\eta_x}{2}-\frac{ \beta\eta_x^2}{2} \right)\mathbb{E}\left[\left\|\frac{1}{n} \sum_{i=1}^n \nabla_x f_i(\bm{x}_i^{(t)}, \bm{y}_i^{(t)} ) \right\|^2\right] \nonumber\\
    & +   \frac{2\eta_x L^2}{\mu}\mathbb{E}\left[ (\Phi(\bm{x}^{(t)} )- F(\bm{x}^{(t)},  \bm{y}^{(t)}))\right]   +   2\eta_x  L^2  \mathbb{E}\left[ \delta_{\bm{x}}^{(t)} + \delta_{\bm{y}}^{(t)}\right]  +   \frac{ \beta \eta_x^2 \sigma^2}{2n}  .\nonumber
\end{align}
We can see the convergence depends on $\delta_{\bm{x}}^{(t)} + \delta_{\bm{y}}^{(t)}$, and  $\mathbb{E}\left[ (\Phi(\bm{x}^{(t)} )- F(\bm{x}^{(t)},  \bm{y}^{(t)}))\right]$. For $\delta_{\bm{x}}^{(t)} + \delta_{\bm{y}}^{(t)}$, we bound it in an analogous way to nonconvex-strongly-concave case.

Another thing is to bound the gap $\mathbb{E}\left[ (\Phi(\bm{x}^{(t)} )- F(\bm{x}^{(t)},  \bm{y}^{(t)}))\right]$. Here we borrow the proof idea from~\cite{reisizadeh2020robust}:

{\begin{align}
&\frac{1}{T}\sum_{t=1}^T \mathbb{E}\left[ \Phi(\bm{x}^{(t)} )- F(\bm{x}^{(t)},  \bm{y}^{(t)})\right] \nonumber\\
&\leq \frac{2\mathbb{E}\left[ \Phi(\bm{x}^{(0)} )- F(\bm{x}^{(0)},  \bm{y}^{(0)})\right]}{\mu \eta_y T}  +  \frac{2}{\mu T}\sum_{t=1}^{T } \left( L^2\eta_x^2 \frac{\sigma^2}{n}  +  2L^2 S\eta_x^2 (G_x^2 + \sigma^2 )+2L^2\mathbb{E}\left[ \delta_{\bm{x}}^{(t)} + \delta_{\bm{y}}^{(t)}\right] \right)\nonumber\\
     & + \left[\frac{2(1-\mu\eta_y )}{\mu \eta_y}\left( \frac{\eta_x^2 L}{2}+\frac{ \beta\eta_x^2}{2}\right)+ L^2\eta_x^2 \right]\frac{1}{T}\left(\mathbb{E}\left[\left\|\frac{1}{n}\sum_{i=1}^n \nabla_x f_i(\bm{x}_i^{(0)}, \bm{y}_i^{(0)})\right\|^2 \right]+\sum_{t=1}^{T }  \mathbb{E}\left[\left\|\frac{1}{n}\sum_{i=1}^n \nabla_x f_i(\bm{x}_i^{(t)}, \bm{y}_i^{(t)})\right\|^2 \right] \right) \nonumber\\
     & + \frac{2(1-\mu\eta_y )}{\mu \eta_y}\frac{1}{T}\left(\sum_{t=1}^{T }\left( \frac{1}{2}\eta_x \mathbb{E}\left[   \left \| \nabla \Phi(\bm{x}^{(t)})\right \|^2 \right] + \frac{\eta_x^2 L\sigma^2}{2n}  \right)+\mathbb{E}\left[   \left \| \nabla \Phi(\bm{x}^{(0)})\right \|^2 \right] \right)\nonumber\\
     &+\frac{2(1-\mu\eta_y )}{\mu \eta_y}\frac{1}{T}\sum_{t=1}^{T }  \left( 2\eta_x  L^2  \mathbb{E}\left[ \delta_{\bm{x}}^{(t)} + \delta_{\bm{y}}^{(t)}\right]  +   \frac{ \beta \eta_x^2 \sigma^2}{2n}  \right)+\frac{\eta_y  L\sigma^2 }{ n}\nonumber.
\end{align}}

Putting these piece together, concludes the proof. 

\subsection{Proof of technical lemmas}
We first introduce some useful lemmas. The following lemma performs one iteration analysis of local SGDA+, on nonconvex-PL objective.
\begin{lemma}\label{Lm: NCPL-1}
For local-SGDA+, under the assumptions in Theorem~\ref{Thm: NCNC}, the following statement holds:
{\begin{align}
\mathbb{E}\left[\Phi (\bm{x}^{(t+1)} )\right] - \mathbb{E}\left[\Phi(\bm{x}^{(t)})\right]  
    & \leq   - \frac{\eta_x}{2}\mathbb{E}\left[\left\| \nabla \Phi(\bm{x}^{(t)})\right\|^2\right] - \left(\frac{\eta_x}{2}-\frac{ \beta\eta_x^2}{2} \right)\mathbb{E}\left[\left\|\frac{1}{n} \sum_{i=1}^n \nabla_x f_i(\bm{x}_i^{(t)}, \bm{y}_i^{(t)} ) \right\|^2\right] \nonumber\\
    & +   \frac{2\eta_x L^2}{\mu}\mathbb{E}\left[ (\Phi(\bm{x}^{(t)} )- F(\bm{x}^{(t)},  \bm{y}^{(t)}))\right]   +   2\eta_x  L^2  \mathbb{E}\left[ \delta_{\bm{x}}^{(t)} + \delta_{\bm{y}}^{(t)}\right]  +   \frac{ \beta \eta_x^2 \sigma^2}{2n}  .\nonumber
\end{align}}
where $\beta=L+ \kappa L/2$.
\begin{proof}
 Since $\Phi(\cdot)$ is $\beta=L+ \kappa L$-smooth, we have:
 {\begin{align}
     \Phi(\bm{x}^{(t+1)}) &\leq  \Phi(\bm{x}^{(t)}) + \left\langle \nabla \Phi(\bm{x}^{(t)}), \bm{x}^{(t+1)}-\bm{x}^{(t)} \right\rangle + \frac{\beta}{2}\left\| \bm{x}^{(t+1)}-\bm{x}^{(t)} \right\|^2\nonumber\\
    & \leq  \Phi(\bm{x}^{(t)}) - \eta_x\left\langle \nabla \Phi(\bm{x}^{(t)}), \frac{1}{n} \sum_{i=1}^n \nabla_x f_i(\bm{x}_i^{(t)}, \bm{y}_i^{(t)};\xi_i^t) \right\rangle + \frac{\beta}{2}\eta^2\left\| \frac{1}{n} \sum_{i=1}^n \nabla_x f_i(\bm{x}_i^{(t)}, \bm{y}_i^{(t)};\xi_i^t) \right\|^2.\nonumber
 \end{align}}
 Taking expectation on both sides yields:
  {\begin{align}
      \mathbb{E}\left[\Phi(\bm{x}^{(t+1)})\right] &\leq  \mathbb{E}\left[\Phi(\bm{x}^{(t)})\right] -  \eta_x\left\langle \nabla \Phi(\bm{x}^{(t)}),\frac{1}{n} \sum_{i=1}^n \nabla_x f_i(\bm{x}_i^{(t)}, \bm{y}_i^{(t)} ) \right\rangle + \frac{\beta}{2}\eta_x^2\mathbb{E}\left[\left\| \frac{1}{n} \sum_{i=1}^n \nabla_x f_i(\bm{x}_i^{(t)}, \bm{y}_i^{(t)};\xi_i^t) \right\|^2\right] \nonumber . 
 \end{align}}
 Using the identity $\langle \bm{a},\bm{b}\rangle = -\frac{1}{2}\|\bm{a} - \bm{b}\|^2 + \frac{1}{2}\|\bm{a}\|^2 + \frac{1}{2}\|\bm{b}\|^2 $, we have:
   {\begin{align}
     & \mathbb{E}\left[\Phi(\bm{x}^{(t+1)})\right]-\mathbb{E}\left[\Phi(\bm{x}^{(t)})\right]  \nonumber\\
     & \leq   - \frac{\eta_x}{2}\mathbb{E}\left[\left\| \nabla \Phi(\bm{x}^{(t)})\right\|^2\right] - \frac{\eta_x}{2}\mathbb{E}\left[\left\|\frac{1}{n} \sum_{i=1}^n \nabla_x f_i(\bm{x}_i^{(t)}, \bm{y}_i^{(t)} ) \right\|^2\right] + \eta_x\mathbb{E}\left[\left\| \nabla \Phi(\bm{x}^{(t)}) - \frac{1}{n} \sum_{i=1}^n \nabla_x f_i(\bm{x}^{(t)}, \bm{y}^{(t)} ) \right\|^2\right] \nonumber\\
    & \quad + \eta_x \mathbb{E}\left[\left\|\frac{1}{n} \sum_{i=1}^n \nabla_x f_i(\bm{x}_i^{(t)}, \bm{y}_i^{(t)} ) - \frac{1}{n} \sum_{i=1}^n \nabla_x f_i(\bm{x}^{(t)}, \bm{y}^{(t)} )\right\|^2\right] \nonumber\\
    & \quad +  \frac{\beta}{2}\eta_x^2\mathbb{E}\left[\left\| \frac{1}{n} \sum_{i=1}^n \nabla_x f_i(\bm{x}_i^{(t)}, \bm{y}_i^{(t)} ) \right\|^2\right] + \frac{\beta\eta_x^2\sigma^2}{2n}\nonumber\\
    & \leq   - \frac{\eta_x}{2}\mathbb{E}\left[\left\| \nabla \Phi(\bm{x}^{(t)})\right\|^2\right] - \left(\frac{\eta_x}{2} - \frac{\beta \eta_x^2}{2}\right)\mathbb{E}\left[\left\|\frac{1}{n} \sum_{i=1}^n \nabla_x f_i(\bm{x}_i^{(t)}, \bm{y}_i^{(t)} ) \right\|^2\right] + \eta_x L^2\mathbb{E}\left[\left\| \phi(\bm{x}^{(t)}) -\bm{y}^{(t)}   \right\|^2\right] \nonumber\\
    &  \quad +  \eta_x L^2  \frac{1}{n} \sum_{i=1}^n  \mathbb{E}\left[2\left\|  \bm{x}_i^{(t)}  -   \bm{x}^{(t)}   \right\|^2 + 2\left\|  \bm{y}_i^{(t)}  -  \bm{y}^{(t)}  \right\|^2\right] + \frac{\beta\eta_x^2\sigma^2}{2n} \nonumber\\ 
     & \leq   - \frac{\eta_x}{2}\mathbb{E}\left[\left\| \nabla \Phi(\bm{x}^{(t)})\right\|^2\right] - \left(\frac{\eta_x}{2}-\frac{ \beta\eta_x^2}{2} \right)\mathbb{E}\left[\left\|\frac{1}{n} \sum_{i=1}^n \nabla_x f_i(\bm{x}_i^{(t)}, \bm{y}_i^{(t)} ) \right\|^2\right] +  \eta_x L^2 \mathbb{E}\left[\left\| \phi(\bm{x}^{(t)}) -\bm{y}^{(t)}   \right\|^2\right] \nonumber\\
    &  \quad  +   2\eta_x  L^2  \mathbb{E}\left[ \delta_{\bm{x}}^{(t)} + \delta_{\bm{y}}^{(t)}\right]  +   \frac{ \beta \eta_x^2 \sigma^2}{2n}  .\nonumber
 \end{align}}
 According to~\cite{karimi2016linear}, PL condition implies quadratic growth, we have:
 \begin{align}
     \left\| \phi(\bm{x}^{(t)}) -\bm{y}^{(t)}   \right\|^2 \leq \frac{2}{\mu} (F(\bm{x}^{(t)}, \phi(\bm{x}^{(t)}))- F(\bm{x}^{(t)},  \bm{y}^{(t)})) = \frac{2}{\mu} (\Phi(\bm{x}^{(t)} )- F(\bm{x}^{(t)},  \bm{y}^{(t)})),
 \end{align}
 which concludes the proof.

\end{proof}
\end{lemma}

The following lemma characterizes the sub-linear convergence of gap $ \mathbb{E}\left[ \Phi(\bm{x}^{(t)} )- F(\bm{x}^{(t)},  \bm{y}^{(t)})\right]$.
\begin{lemma}\label{Lm: NCPL-2}
For local-SGDA+, under the assumptions in Theorem~\ref{Thm: NCNC}, the following statement holds:
{\begin{align}
&\frac{1}{T}\sum_{t=1}^T \mathbb{E}\left[ \Phi(\bm{x}^{(t)} )- F(\bm{x}^{(t)},  \bm{y}^{(t)})\right] \nonumber\\
&\leq \frac{2\mathbb{E}\left[ \Phi(\bm{x}^{(0)} )- F(\bm{x}^{(0)},  \bm{y}^{(0)})\right]}{\mu \eta_y T}  +  \frac{2}{\mu T}\sum_{t=1}^{T } \left( L^2\eta_x^2 \frac{\sigma^2}{n}  +  2L^2 S\eta_x^2 (G_x^2 + \sigma^2 )+2L^2\mathbb{E}\left[ \delta_{\bm{x}}^{(t)} + \delta_{\bm{y}}^{(t)}\right] \right)\nonumber\\
     & + \left[\frac{2(1-\mu\eta_y )}{\mu \eta_y}\left( \frac{\eta_x^2 L}{2}+\frac{ \beta\eta_x^2}{2}\right)+ L^2\eta_x^2 \right]\frac{1}{T}\left(\mathbb{E}\left[\left\|\frac{1}{n}\sum_{i=1}^n \nabla_x f_i(\bm{x}_i^{(0)}, \bm{y}_i^{(0)})\right\|^2 \right]+\sum_{t=1}^{T }  \mathbb{E}\left[\left\|\frac{1}{n}\sum_{i=1}^n \nabla_x f_i(\bm{x}_i^{(t)}, \bm{y}_i^{(t)})\right\|^2 \right] \right) \nonumber\\
     & + \frac{2(1-\mu\eta_y )}{\mu \eta_y}\frac{1}{T}\left(\sum_{t=1}^{T }\left( \frac{1}{2}\eta_x \mathbb{E}\left[   \left \| \nabla \Phi(\bm{x}^{(t)})\right \|^2 \right] + \frac{\eta_x^2 L\sigma^2}{2n}  \right)+\mathbb{E}\left[   \left \| \nabla \Phi(\bm{x}^{(0)})\right \|^2 \right] \right)\nonumber\\
     &+\frac{2(1-\mu\eta_y )}{\mu \eta_y}\frac{1}{T}\sum_{t=1}^{T }  \left( 2\eta_x  L^2  \mathbb{E}\left[ \delta_{\bm{x}}^{(t)} + \delta_{\bm{y}}^{(t)}\right]  +   \frac{ \beta \eta_x^2 \sigma^2}{2n}  \right)+\frac{\eta_y  L\sigma^2 }{ n}\nonumber.
\end{align}}
\begin{proof}
 According to smoothness of $F(\bm{x},\cdot)$, we have
 \begin{align}
    F(\bm{x}^{(t+1)}, \bm{y}^{(t)})  &\leq F(\bm{x}^{(t+1)}, \bm{y}^{(t+1)}) - \left \langle \nabla_y F(\bm{x}^{(t+1)}, \bm{y}^{(t)}), \bm{y}^{(t+1)} - \bm{y}^{(t)}   \right\rangle + \frac{L}{2} \left \| \frac{1}{n} \sum_{i=1}^n\nabla_y f_i(\Tilde{\bm{x}} , \bm{y}_i^{(t)}; \xi_i^t)  \right\|^2\nonumber\\
     &\leq F(\bm{x}^{(t+1)}, \bm{y}^{(t+1)})  - \eta_y\left \langle \nabla_y F(\bm{x}^{(t+1)}, \bm{y}^{(t)}), \frac{1}{n} \nabla_y f_i(\Tilde{\bm{x}} , \bm{y}_i^{(t)}; \xi_i^t)  \right\rangle + \frac{\eta_y^2 L}{2} \left \| \frac{1}{n}\sum_{i=1}^n \nabla_y f_i(\Tilde{\bm{x}} , \bm{y}_i^{(t)}; \xi_i^t)  \right\|^2\nonumber 
 \end{align}
 Taking expectation on both sides yields:
  \begin{align}
     \mathbb{E}[F(\bm{x}^{(t+1)}, \bm{y}^{(t)})]  &\leq \mathbb{E}[F(\bm{x}^{(t+1)}, \bm{y}^{(t+1)})]  - \eta_y\mathbb{E}\left[\left \langle \nabla_y F(\bm{x}^{(t+1)}, \bm{y}^{(t)}), \frac{1}{n}\sum_{i=1}^n \nabla_y f_i(\Tilde{\bm{x}} , \bm{y}_i^{(t)} )  \right\rangle \right]\nonumber\\
     &+ \frac{\eta_y^2 L}{2}  \mathbb{E}\left[\left \| \frac{1}{n}\sum_{i=1}^n \nabla_y f_i(\Tilde{\bm{x}} , \bm{y}_i^{(t)}; \xi_i^t)  \right\|^2\right]\nonumber \\
     &\leq  \mathbb{E}[F(\bm{x}^{(t+1)}, \bm{y}^{(t+1)})] - \frac{\eta_y}{2} \mathbb{E}\left[\left \|  \nabla_y F(\bm{x}^{(t+1)} , \bm{y}_i^{(t)})  \right\|^2\right] +   
     \frac{1}{2}\eta_y\mathbb{E}\left[\left \| \nabla_y F(\bm{x}^{(t+1)}, \bm{y}^{(t)})- \frac{1}{n}\sum_{i=1}^n \nabla_y f_i(\Tilde{\bm{x}} , \bm{y}_i^{(t)} )  \right\|^2 \right]\nonumber\\
     &-\left(\frac{\eta_y}{2}- \frac{\eta_y^2 L}{2} \right) \mathbb{E}\left[\left \| \frac{1}{n}\sum_{i=1}^n \nabla_y f_i(\Tilde{\bm{x}} , \bm{y}_i^{(t)})  \right\|^2\right] +\frac{\eta_y^2 L\sigma^2 }{2n}  \label{eq: NCOC-lm2 1}\\
      &\leq  \mathbb{E}[F(\bm{x}^{(t+1)}, \bm{y}^{(t+1)})] - \frac{\eta_y}{2} \mathbb{E}\left[\left \|  \nabla_y F(\bm{x}^{(t+1)} , \bm{y}_i^{(t)})  \right\|^2\right]-\left(\frac{\eta_y}{2}- \frac{\eta_y^2 L}{2} \right) \mathbb{E}\left[\left \| \frac{1}{n}\sum_{i=1}^n \nabla_y f_i(\Tilde{\bm{x}} , \bm{y}_i^{(t)})  \right\|^2\right] +\frac{\eta_y^2 L\sigma^2 }{2n} \nonumber\\
     &  +   \frac{1}{2}\eta_y\mathbb{E}\left[\left \| \nabla_y F(\bm{x}^{(t+1)}, \bm{y}^{(t)})-\nabla_y F(\bm{x}^{(t)}, \bm{y}^{(t)})+\nabla_y F(\bm{x}^{(t)}, \bm{y}^{(t)})- \frac{1}{n}\sum_{i=1}^n \nabla_y f_i(\Tilde{\bm{x}} , \bm{y}_i^{(t)} )  \right\|^2 \right]\nonumber \\ 
     &\leq  \mathbb{E}[F(\bm{x}^{(t+1)}, \bm{y}^{(t+1)})] - \frac{\eta_y}{2} \mathbb{E}\left[\left \|  \nabla_y F(\bm{x}^{(t+1)} , \bm{y}_i^{(t)})  \right\|^2\right]-\left(\frac{\eta_y}{2}- \frac{\eta_y^2 L}{2} \right) \mathbb{E}\left[\left \| \frac{1}{n}\sum_{i=1}^n \nabla_y f_i(\Tilde{\bm{x}} , \bm{y}_i^{(t)})  \right\|^2\right] +\frac{\eta_y^2 L\sigma^2 }{2n} \nonumber\\
     &  +  \eta_y\underbrace{\mathbb{E}\left[\left \| \nabla_y F(\bm{x}^{(t+1)}, \bm{y}^{(t)})-\nabla_y F(\bm{x}^{(t)}, \bm{y}^{(t)})\right\|^2 \right]}_{T_1}+\eta_y\underbrace{\mathbb{E}\left[\left \| \nabla_y F(\bm{x}^{(t)}, \bm{y}^{(t)})- \frac{1}{n}\sum_{i=1}^n \nabla_y f_i(\Tilde{\bm{x}} , \bm{y}_i^{(t)} )  \right\|^2 \right]}_{T_2}\nonumber,
 \end{align}
 where we use the identity $\langle \bm{a},\bm{b}\rangle = -\frac{1}{2}\|\bm{a} - \bm{b}\|^2 + \frac{1}{2}\|\bm{a}\|^2 + \frac{1}{2}\|\bm{b}\|^2 $.
 
 To bound $T_1$, we notice that:
\begin{align}
    T_1 \leq L^2 \mathbb{E}\left[\left \|  \bm{x}^{(t+1)} - \bm{x}^{(t)} \right\|^2 \right]\leq L^2\eta_x^2 \mathbb{E}\left[\left \|  \frac{1}{n}\sum_{i=1}^n \nabla_x f_i(\bm{x}_i^{(t)},\bm{y}_i^{(t)} ) \right\|^2 \right]+L^2\eta_x^2 \frac{\sigma^2}{n}\nonumber.
\end{align}
For $T_2$, we bound it as follows:
\begin{align}
    T_2 &\leq 2\mathbb{E}\left[\left \| \nabla_y F(\bm{x}^{(t)}, \bm{y}^{(t)})- \nabla_y F(\Tilde{\bm{x}}, \bm{y}^{(t)}) \right\|^2 \right]+2\mathbb{E}\left[\left \| \nabla_y F(\Tilde{\bm{x}}, \bm{y}^{(t)})- \frac{1}{n}\sum_{i=1}^n \nabla_y f_i(\Tilde{\bm{x}} , \bm{y}_i^{(t)} )  \right\|^2 \right]  \nonumber \\
    & \leq 2L^2\mathbb{E}\left[\left \|  \bm{x}^{(t)} -  \Tilde{\bm{x}}  \right\|^2 \right]+2L^2\frac{1}{n}\sum_{i=1}^n \mathbb{E}\left[\left \|  \bm{y}^{(t)} -   \bm{y}_i^{(t)} \right\|^2 \right]  \nonumber \\
     & \leq 2L^2 S\eta_x^2 (G_x^2 + \sigma^2 )+2L^2\frac{1}{n}\sum_{i=1}^n \mathbb{E}\left[\left \|  \bm{y}^{(t)} -   \bm{y}_i^{(t)} \right\|^2 \right]  \nonumber 
\end{align}
Putting these pieces together yields:
  \begin{align}
     \mathbb{E}[F(\bm{x}^{(t+1)}, \bm{y}^{(t)})]   &\leq \mathbb{E}[F(\bm{x}^{(t+1)}, \bm{y}^{(t+1)})]  - \frac{\eta_y}{2} \mathbb{E}\left[\left \|  \nabla_y F(\bm{x}^{(t+1)} , \bm{y}^{(t)})  \right\|^2\right]-\left(\frac{\eta_y}{2}- \frac{\eta_y^2 L}{2} \right) \mathbb{E}\left[\left \| \frac{1}{n}\sum_{i=1}^n \nabla_y f_i(\Tilde{\bm{x}} , \bm{y}_i^{(t)})  \right\|^2\right] +\frac{\eta_y^2 L\sigma^2 }{2n} \nonumber\\
     &  +  \eta_y\left(L^2\eta_x^2 \mathbb{E}\left[\left \|  \frac{1}{n}\sum_{i=1}^n \nabla_x f_i(\bm{x}_i^{(t)},\bm{y}_i^{(t)} ) \right\|^2 \right]+L^2\eta_x^2 \frac{\sigma^2}{n} \right)\nonumber\\
     &+\eta_y  \left(  2L^2 S\eta_x^2 (G_x^2 + \sigma^2 )+2L^2\frac{1}{n}\sum_{i=1}^n \mathbb{E}\left[\left \|  \bm{y}^{(t)} -   \bm{y}_i^{(t)} \right\|^2 \right] \right) \nonumber\\
     &\leq  \mathbb{E}[F(\bm{x}^{(t+1)}, \bm{y}^{(t+1)})] - \frac{\eta_y}{2} \mathbb{E}\left[\left \|  \nabla_y F(\bm{x}^{(t+1)} , \bm{y}^{(t)})  \right\|^2\right]-\left(\frac{\eta_y}{2}- \frac{\eta_y^2 L}{2} \right) \mathbb{E}\left[\left \| \frac{1}{n}\sum_{i=1}^n \nabla_y f_i(\Tilde{\bm{x}} , \bm{y}_i^{(t)})  \right\|^2\right] +\frac{\eta_y^2 L\sigma^2 }{2n} \nonumber\\
     &  +  \eta_y\left(L^2\eta_x^2 \mathbb{E}\left[\left \|  \frac{1}{n}\sum_{i=1}^n \nabla_x f_i(\bm{x}_i^{(t)},\bm{y}_i^{(t)} ) \right\|^2 \right]+L^2\eta_x^2 \frac{\sigma^2}{n} \right)\nonumber\\
     &+\eta_y  \left(  2L^2 S\eta_x^2 (G_x^2 + \sigma^2 )+2L^2\frac{1}{n}\sum_{i=1}^n \mathbb{E}\left[\left \|  \bm{y}^{(t)} -   \bm{y}_i^{(t)} \right\|^2 \right] \right) \nonumber.
 \end{align}
 Now, applying the PL condition to substitute $\left \|  \nabla_y F(\bm{x}^{(t+1)} , \bm{y}^{(t)})  \right\|^2$:
 \begin{align}
     \left \|  \nabla_y F(\bm{x}^{(t+1)} , \bm{y}^{(t)})  \right\|^2 \geq 2 \mu  (\Phi(\bm{x}^{(t+1)}) - F(\bm{x}^{(t+1)} , \bm{y}^{(t)})).
 \end{align}
 Thus we have:
   \begin{align}
    \eta_y\mu \mathbb{E}\left[  (\Phi(\bm{x}^{(t+1)}) - F(\bm{x}^{(t+1)} , \bm{y}^{(t)}))\right]   
     &\leq  \mathbb{E}[F(\bm{x}^{(t+1)}, \bm{y}^{(t+1)})]-\mathbb{E}[F(\bm{x}^{(t+1)}, \bm{y}^{(t)})]    +\frac{\eta_y^2 L\sigma^2 }{2n} \nonumber\\
     & \quad +  \eta_y\left(L^2\eta_x^2 \mathbb{E}\left[\left \|  \frac{1}{n}\sum_{i=1}^n \nabla_x f_i(\bm{x}_i^{(t)},\bm{y}_i^{(t)} ) \right\|^2 \right]+L^2\eta_x^2 \frac{\sigma^2}{n} \right)\nonumber\\
     &\quad+\eta_y  \left(  2L^2 S\eta_x^2 (G_x^2 + \sigma^2 )+2L^2\frac{1}{n}\sum_{i=1}^n \mathbb{E}\left[\left \|  \bm{y}^{(t)} -   \bm{y}_i^{(t)} \right\|^2 \right] \right) \nonumber.
 \end{align}
 Re-arranging the terms yields:
    \begin{align}
    \mathbb{E}\left[  (\Phi(\bm{x}^{(t+1)}) - F(\bm{x}^{(t+1)} , \bm{y}^{(t+1)}))\right]   
     &\leq   (1-\mu\eta_y)\mathbb{E}\left[  (\Phi(\bm{x}^{(t+1)}) - F(\bm{x}^{(t+1)} , \bm{y}^{(t)}))\right]     +\frac{\eta_y^2 L\sigma^2 }{2n} \nonumber\\
     & \quad +  \eta_y\left(L^2\eta_x^2 \mathbb{E}\left[\left \|  \frac{1}{n}\sum_{i=1}^n \nabla_x f_i(\bm{x}_i^{(t)},\bm{y}_i^{(t)} ) \right\|^2 \right]+L^2\eta_x^2 \frac{\sigma^2}{n} \right)\nonumber\\
     &\quad+\eta_y  \left(  2L^2 S\eta_x^2 (G_x^2 + \sigma^2 )+2L^2\frac{1}{n}\sum_{i=1}^n \mathbb{E}\left[\left \|  \bm{y}^{(t)} -   \bm{y}_i^{(t)} \right\|^2 \right] \right) \nonumber.
 \end{align}
 
 Notice that in RHS:
 \begin{align}
\mathbb{E}[ \Phi(\bm{x}^{(t+1)}) - F(\bm{x}^{(t+1)} , \bm{y}^{(t)})] =\mathbb{E}[\Phi(\bm{x}^{(t)}) - F(\bm{x}^{(t)} , \bm{y}^{(t)})] +\underbrace{\mathbb{E}[ \Phi(\bm{x}^{(t+1)}) - \Phi(\bm{x}^{(t)})]}_{T_3} + \underbrace{\mathbb{E}[F(\bm{x}^{(t)} , \bm{y}^{(t)}) -F(\bm{x}^{(t+1)} , \bm{y}^{(t)})]}_{T_4} 
 \end{align}
 According to Lemma~\ref{Lm: NCPL-1} we can bound $T_3$ as:
  \begin{align}
  T_3 &   \leq   - \frac{\eta_x}{2}\mathbb{E}\left[\left\| \nabla \Phi(\bm{x}^{(t)})\right\|^2\right] - \left(\frac{\eta_x}{2}-\frac{ \beta\eta_x^2}{2} \right)\mathbb{E}\left[\left\|\frac{1}{n} \sum_{i=1}^n \nabla_x f_i(\bm{x}_i^{(t)}, \bm{y}_i^{(t)} ) \right\|^2\right] \nonumber\\
    & +   \frac{2\eta_x L^2}{\mu}\mathbb{E}\left[ (\Phi(\bm{x}^{(t)} )- F(\bm{x}^{(t)},  \bm{y}^{(t)}))\right]   +   2\eta_x  L^2  \mathbb{E}\left[ \delta_{\bm{x}}^{(t)} + \delta_{\bm{y}}^{(t)}\right]  +   \frac{ \beta \eta_x^2 \sigma^2}{2n}  .\nonumber
 \end{align}
 For $T_4$, applying smoothness of $F(\cdot,\bm{y}^{(t)})$ gives:
   \begin{align}
  T_4 &= \mathbb{E}[ F(\bm{x}^{(t)} , \bm{y}^{(t)}) -F(\bm{x}^{(t+1)} , \bm{y}^{(t)}) ] \leq  \mathbb{E}[  - \left \langle \nabla_x F(\bm{x}^{(t)},\bm{y}^{(t)}), \bm{x}^{(t+1)}-\bm{x}^{(t)} \right \rangle ]+ \frac{L}{2}\mathbb{E}\left[\left\| \bm{x}^{(t+1)}-\bm{x}^{(t)} \right\|^2 \right]\nonumber\\
  &= \eta_x \mathbb{E}\left[   \left \langle \nabla_x F(\bm{x}^{(t)},\bm{y}^{(t)}), \frac{1}{n}\sum_{i=1}^n \nabla_x f_i(\bm{x}_i^{(t)}, \bm{y}_i^{(t)}) \right \rangle \right]+ \frac{\eta_x^2 L}{2}\mathbb{E}\left[\left\|\frac{1}{n}\sum_{i=1}^n \nabla_x f_i(\bm{x}_i^{(t)}, \bm{y}_i^{(t)})\right\|^2 \right] + \frac{\eta_x^2 L\sigma^2}{2n}\nonumber\\
  & \leq  \frac{1}{2}\eta_x \mathbb{E}\left[   \left \| \nabla_x F(\bm{x}^{(t)},\bm{y}^{(t)})\right \|^2 \right]+\frac{1}{2}\eta_x \mathbb{E}\left[   \left \| \frac{1}{n}\sum_{i=1}^n \nabla_x f_i(\bm{x}_i^{(t)}, \bm{y}_i^{(t)}) \right \|^2 \right]+ \frac{\eta_x^2 L}{2}\mathbb{E}\left[\left\|\frac{1}{n}\sum_{i=1}^n \nabla_x f_i(\bm{x}_i^{(t)}, \bm{y}_i^{(t)})\right\|^2 \right] + \frac{\eta_x^2 L\sigma^2}{2n}\nonumber\\
    & \leq \eta_x \mathbb{E}\left[   \left \| \nabla \Phi(\bm{x}^{(t)})\right \|^2 \right]+ \eta_x \mathbb{E}\left[   \left \| \nabla_x F(\bm{x}^{(t)},\bm{y}^{(t)})-\nabla \Phi(\bm{x}^{(t)})\right \|^2 \right] + \left(\frac{1}{2}\eta_x+\frac{\eta_x^2 L}{2}\right)\mathbb{E}\left[\left\|\frac{1}{n}\sum_{i=1}^n \nabla_x f_i(\bm{x}_i^{(t)}, \bm{y}_i^{(t)})\right\|^2 \right] + \frac{\eta_x^2 L\sigma^2}{2n}\nonumber. 
  \end{align}
  For $\mathbb{E}\left[   \left \| \nabla_x F(\bm{x}^{(t)},\bm{y}^{(t)})-\nabla \Phi(\bm{x}^{(t)})\right \|^2 \right]$, we apply the smoothness of $F$ and quadratic growth of $F(\bm{x},\cdot)$ to get:
  
  \begin{align}
      \mathbb{E}\left[   \left \| \nabla_x F(\bm{x}^{(t)},\bm{y}^{(t)})-\nabla \Phi(\bm{x}^{(t)})\right \|^2 \right] \leq L^2 \mathbb{E}\left[   \left \|  \bm{y}^{(t)} - \bm{y}^*(\bm{x}^{(t)})\right \|^2 \right] \leq   \frac{2L^2}{\mu} \mathbb{E}\left[F(\bm{x}^{(t)}, \bm{y}^*(\bm{x}^{(t)}))- F(\bm{x}^{(t)},  \bm{y}^{(t)}) \right]\nonumber.
  \end{align}
 Using above bound to replace $\mathbb{E}\left[   \left \| \nabla_x F(\bm{x}^{(t)},\bm{y}^{(t)})-\nabla \Phi(\bm{x}^{(t)})\right \|^2 \right]$ we can finally bound $T_4$ as:
    \begin{align}
  T_4   & \leq \eta_x \mathbb{E}\left[   \left \| \nabla \Phi(\bm{x}^{(t)})\right \|^2 \right]+ \eta_x \frac{2L^2}{\mu} \mathbb{E}\left[\Phi(\bm{x}^{(t)} )- F(\bm{x}^{(t)},  \bm{y}^{(t)}) \right]\nonumber\\
  &+ \left(\frac{1}{2}\eta_x+\frac{\eta_x^2 L}{2}\right)\mathbb{E}\left[\left\|\frac{1}{n}\sum_{i=1}^n \nabla_x f_i(\bm{x}_i^{(t)}, \bm{y}_i^{(t)})\right\|^2 \right] + \frac{\eta_x^2 L\sigma^2}{2n}\nonumber. 
  \end{align}
Plugging $T_3$ and $T_4$ back yields:
\begin{align}
   & \mathbb{E}\left[  (\Phi(\bm{x}^{(t+1)}) - F(\bm{x}^{(t+1)} , \bm{y}^{(t+1)}))\right] \nonumber\\  
     &\leq   (1-\mu\eta_y)\left(1+\eta_x \frac{4L^2}{\mu}\right)\mathbb{E}\left[  (\Phi(\bm{x}^{(t)}) - F(\bm{x}^{(t)} , \bm{y}^{(t)}))\right]     +\frac{\eta_y^2 L\sigma^2 }{2n} \nonumber\\
     & \quad +  (1-\mu\eta_y)\left( \eta_x \mathbb{E}\left[   \left \| \nabla \Phi(\bm{x}^{(t)})\right \|^2 \right]+ \left(\frac{1}{2}\eta_x+\frac{\eta_x^2 L}{2}\right)\mathbb{E}\left[\left\|\frac{1}{n}\sum_{i=1}^n \nabla_x f_i(\bm{x}_i^{(t)}, \bm{y}_i^{(t)})\right\|^2 \right] + \frac{\eta_x^2 L\sigma^2}{2n}  \right)\nonumber\\
     & \quad + (1-\mu\eta_y) \left( - \frac{\eta_x}{2}\mathbb{E}\left[\left\| \nabla \Phi(\bm{x}^{(t)})\right\|^2\right] - \left(\frac{\eta_x}{2}-\frac{ \beta}{2} \eta_x^2\right)\mathbb{E}\left[\left\|\frac{1}{n} \sum_{i=1}^n \nabla_x f_i(\bm{x}_i^{(t)}, \bm{y}_i^{(t)} ) \right\|^2\right]+   2\eta_x  L^2  \mathbb{E}\left[ \delta_{\bm{x}}^{(t)} + \delta_{\bm{y}}^{(t)}\right]  +   \frac{ \beta \eta_x^2 \sigma^2}{2n}  \right)  \nonumber\\ 
     &  \quad +  \eta_y\left(L^2\eta_x^2 \mathbb{E}\left[\left \|  \frac{1}{n}\sum_{i=1}^n \nabla_x f_i(\bm{x}_i^{(t)},\bm{y}_i^{(t)} ) \right\|^2 \right]+L^2\eta_x^2 \frac{\sigma^2}{n} \right)\nonumber\\
     & \quad +\eta_y  \left(  2L^2 S\eta_x^2 (G_x^2 + \sigma^2 )+2L^2\frac{1}{n}\sum_{i=1}^n \mathbb{E}\left[\left \|  \bm{y}^{(t)} -   \bm{y}_i^{(t)} \right\|^2 \right] \right) \nonumber\\
      &\leq   \left(1-\frac{\mu\eta_y}{2}\right) \mathbb{E}\left[  (\Phi(\bm{x}^{(t)}) - F(\bm{x}^{(t)} , \bm{y}^{(t)}))\right]     +\frac{\eta_y^2 L\sigma^2 }{2n} \nonumber\\
     & \quad +  (1-\mu\eta_y)\left( \frac{1}{2}\eta_x \mathbb{E}\left[   \left \| \nabla \Phi(\bm{x}^{(t)})\right \|^2 \right] + \frac{\eta_x^2 L\sigma^2}{2n}  \right)+\left[(1-\mu\eta_y )\left( \frac{\eta_x^2 L}{2}+\frac{ \beta\eta_x^2}{2}\right)+\eta_yL^2\eta_x^2 \right]\mathbb{E}\left[\left\|\frac{1}{n}\sum_{i=1}^n \nabla_x f_i(\bm{x}_i^{(t)}, \bm{y}_i^{(t)})\right\|^2 \right] \nonumber\\
     &  \quad + (1-\mu\eta_y) \left(  2\eta_x  L^2  \mathbb{E}\left[ \delta_{\bm{x}}^{(t)} + \delta_{\bm{y}}^{(t)}\right]  +   \frac{ \beta \eta_x^2 \sigma^2}{2n}  \right)  \nonumber\\ 
     &  \quad +  \eta_y\left( L^2\eta_x^2 \frac{\sigma^2}{n}  +  2L^2 S\eta_x^2 (G_x^2 + \sigma^2 )+2L^2\frac{1}{n}\sum_{i=1}^n \mathbb{E}\left[\left \|  \bm{y}^{(t)} -   \bm{y}_i^{(t)} \right\|^2 \right] \right) \nonumber,
 \end{align}
 where we use the fact $(1-\mu\eta_y)(1+\frac{4L^2 \eta_x}{\mu}) \leq(1-\frac{\mu\eta_y}{2})$ due to $\eta_x \leq \frac{\mu \eta_y}{2(4L^2/\mu - 4L^2\eta_y)}$.  Denote $A_t = \mathbb{E}\left[  (\Phi(\bm{x}^{(t)}) - F(\bm{x}^{(t)} , \bm{y}^{(t)}))\right]$. It is obvious that $A_t \geq 0$ for all $t$. Then, based on the above inequality and do the summation:
 \begin{align}
     \frac{1}{T}\sum_{t=1}^T A_t &\leq \frac{1}{T}\sum_{t=0}^{T-1}\left(1-\frac{\mu\eta_y}{2}\right) A_t + \frac{1}{T}\sum_{t=0}^{T-1}  \left[(1-\mu\eta_y )\left( \frac{\eta_x^2 L}{2}+\frac{ \beta\eta_x^2}{2}\right)+\eta_yL^2\eta_x^2 \right]\mathbb{E}\left[\left\|\frac{1}{n}\sum_{i=1}^n \nabla_x f_i(\bm{x}_i^{(t)}, \bm{y}_i^{(t)})\right\|^2 \right] \nonumber\\
     & + \frac{1}{T}\sum_{t=0}^{T-1}(1-\mu\eta_y)\left( \frac{1}{2}\eta_x \mathbb{E}\left[   \left \| \nabla \Phi(\bm{x}^{(t)})\right \|^2 \right] + \frac{\eta_x^2 L\sigma^2}{2n}  \right) +\frac{\eta_y^2 L\sigma^2 }{2n} \nonumber\\
     &+\frac{1}{T}\sum_{t=0}^{T-1} (1-\mu\eta_y) \left( 2\eta_x  L^2  \mathbb{E}\left[ \delta_{\bm{x}}^{(t)} + \delta_{\bm{y}}^{(t)}\right]  +   \frac{ \beta \eta_x^2 \sigma^2}{2n}  \right)\nonumber\\
     & + \frac{1}{T}\sum_{t=0}^{T-1} \eta_y\left( L^2\eta_x^2 \frac{\sigma^2}{n}  +  2L^2 S\eta_x^2 (G_x^2 + \sigma^2 )+2L^2\frac{1}{n}\sum_{i=1}^n \mathbb{E}\left[\left \|  \bm{y}^{(t)} -   \bm{y}_i^{(t)} \right\|^2 \right] \right) \nonumber\\
     &\leq (1-\frac{\mu\eta_y}{2})\frac{1}{T}\left(A_0 + \sum_{t=1}^{T } A_t\right)+\frac{\eta_y^2 L\sigma^2 }{2n}\nonumber\\
     & + \left[(1-\mu\eta_y )\left( \frac{\eta_x^2 L}{2}+\frac{ \beta\eta_x^2}{2}\right)+\eta_yL^2\eta_x^2 \right]\frac{1}{T}\left(\mathbb{E}\left[\left\|\frac{1}{n}\sum_{i=1}^n \nabla_x f_i(\bm{x}_i^{(0)}, \bm{y}_i^{(0)})\right\|^2 \right]+\sum_{t=1}^{T }  \mathbb{E}\left[\left\|\frac{1}{n}\sum_{i=1}^n \nabla_x f_i(\bm{x}_i^{(t)}, \bm{y}_i^{(t)})\right\|^2 \right] \right) \nonumber\\
     & + (1-\mu\eta_y)\frac{1}{T}\left(\sum_{t=1}^{T }\left( \frac{1}{2}\eta_x \mathbb{E}\left[   \left \| \nabla \Phi(\bm{x}^{(t)})\right \|^2 \right] + \frac{\eta_x^2 L\sigma^2}{2n}  \right)+\mathbb{E}\left[   \left \| \nabla \Phi(\bm{x}^{(0)})\right \|^2 \right] \right)\nonumber\\
     &+(1-\mu\eta_y)\frac{1}{T}\sum_{t=1}^{T }  \left(  2\eta_x  L^2  \mathbb{E}\left[ \delta_{\bm{x}}^{(t)} + \delta_{\bm{y}}^{(t)}\right]  +   \frac{ \beta \eta_x^2 \sigma^2}{2n}  \right)\nonumber\\
     & + \eta_y\frac{1}{T}\sum_{t=1}^{T } \left( L^2\eta_x^2 \frac{\sigma^2}{n}  +  2L^2 S\eta_x^2 (G_x^2 + \sigma^2 )+2L^2\mathbb{E}\left[ \delta_{\bm{x}}^{(t)} + \delta_{\bm{y}}^{(t)}\right] \right) \nonumber
 \end{align}
 Re-arranging the terms will conclude the proof:
  \begin{align}
     \frac{1}{T}\sum_{t=1}^T A_t 
     &\leq \frac{2A_0}{\mu \eta_y T} +\frac{\eta_y  L\sigma^2 }{ n}\nonumber\\
     & + \left[\frac{2(1-\mu\eta_y )}{\mu \eta_y}\left( \frac{\eta_x^2 L}{2}+\frac{ \beta\eta_x^2}{2}\right)+ L^2\eta_x^2 \right]\frac{1}{T}\left(\mathbb{E}\left[\left\|\frac{1}{n}\sum_{i=1}^n \nabla_x f_i(\bm{x}_i^{(0)}, \bm{y}_i^{(0)})\right\|^2 \right]+\sum_{t=1}^{T }  \mathbb{E}\left[\left\|\frac{1}{n}\sum_{i=1}^n \nabla_x f_i(\bm{x}_i^{(t)}, \bm{y}_i^{(t)})\right\|^2 \right] \right) \nonumber\\
     & + \frac{2(1-\mu\eta_y )}{\mu \eta_y}\frac{1}{T}\left(\sum_{t=1}^{T }\left( \frac{1}{2}\eta_x \mathbb{E}\left[   \left \| \nabla \Phi(\bm{x}^{(t)})\right \|^2 \right] + \frac{\eta_x^2 L\sigma^2}{2n}  \right)+\mathbb{E}\left[   \left \| \nabla \Phi(\bm{x}^{(0)})\right \|^2 \right] \right)\nonumber\\
     &+\frac{2(1-\mu\eta_y )}{\mu \eta_y}\frac{1}{T}\sum_{t=1}^{T }  \left( 2\eta_x  L^2  \mathbb{E}\left[ \delta_{\bm{x}}^{(t)} + \delta_{\bm{y}}^{(t)}\right]  +   \frac{ \beta \eta_x^2 \sigma^2}{2n}  \right)\nonumber\\
     & +  \frac{2}{\mu T}\sum_{t=1}^{T } \left( L^2\eta_x^2 \frac{\sigma^2}{n}  +  2L^2 S\eta_x^2 (G_x^2 + \sigma^2 )+2L^2\mathbb{E}\left[ \delta_{\bm{x}}^{(t)} + \delta_{\bm{y}}^{(t)}\right] \right) \nonumber.
 \end{align}
\end{proof}
\end{lemma}

The next lemma bounds the local model deviations on nonconvex-PL objective.

\begin{lemma}\label{lm: nonconvex lm3}

For local-SGDA+, under assumptions of Theorem~\ref{Thm: NCNC}, the following statement holds true:
\begin{align}
 \frac{1}{T}\sum_{t=1}^{T}\frac{1}{n}\sum_{i=1}^n\mathbb{E}\left[\left\|\bm{x}^{(t)}-\bm{x}^{(t)}_i \right\|^2 \right]+\mathbb{E}\left[\left\|\bm{y}^{(t)}-\bm{y}^{(t)}_i \right\|^2 \right]   \leq  10\tau^2 (\eta_x^2 + \eta_y^2) \left( \sigma^2 + \frac{\sigma^2}{n}  \right)+10\tau^2\eta_x^2\zeta_x+ 10\tau^2 \eta_y^2 \zeta_y  .\nonumber
\end{align}
\begin{proof}
 Similarly, for the second statement, we define $\gamma^t = \frac{1}{n}\sum_{i=1}^n\mathbb{E}\left[\left\|\bm{x}^{(t)}-\bm{x}^{(t)}_i \right\|^2 \right]+\mathbb{E}\left[\left\|\bm{y}^{(t)}-\bm{y}^{(t)}_i \right\|^2 \right]   $, then we have:
 
 \begin{align}
 \gamma^t &\leq \frac{1}{n}\sum_{i=1}^n\frac{1}{n}\sum_{k=1}^n\mathbb{E}\left[\left\| \bm{x}^{r\tau} - \sum_{j=r\tau}^{(r+1)\tau} \eta_x  \nabla_x f_k(\bm{x}_k^{(j)},\bm{y}_k^{(j)};\xi_k^j)  -\left(\bm{x}^{r\tau}  -\sum_{j=r\tau}^{(r+1)\tau}\eta_x \nabla_x f_i(\bm{x}_i^{(j)},\bm{y}_i^{(j)};\xi_i^j) \right) \right\|^2\right]\nonumber\\
 & \quad +\frac{1}{n}\sum_{i=1}^n\frac{1}{n}\sum_{k=1}^n\mathbb{E}\left[\left\| \bm{y}^{r\tau} - \sum_{j=r\tau}^{(r+1)\tau} \eta_y  \nabla_y f_k(\Tilde{\bm{x}} ,\bm{y}_k^{(j)};\xi_k^j)  -\left(\bm{y}^{r\tau}  -\sum_{j=r\tau}^{(r+1)\tau}\eta_y \nabla_y f_i(\Tilde{\bm{x}},\bm{y}_i^{(j)};\xi_i^j) \right) \right\|^2\right]\nonumber\\
 & \leq\tau \sum_{j=r\tau}^{(r+1)\tau}\frac{\eta_x^2}{n}\sum_{i=1}^n \frac{1}{n}\sum_{k=1}^n \mathbb{E}\left[\left\|  \nabla_x f_k(\bm{x}_k^{(j)},\bm{y}_k^{(j)};\xi_k^j)  -  \nabla_x f_i(\bm{x}_i^{(j)},\bm{y}_i^{(j)};\xi_i^j)  \right\|^2\right]\nonumber\\
 & \quad + \tau\sum_{j=r\tau}^{(r+1)\tau}\frac{\eta_y}{n}\sum_{i=1}^n  \frac{1}{n}\sum_{k=1}^n \mathbb{E}\left[\left\|  \nabla_y f_k(\Tilde{\bm{x}},\bm{y}_k^{(j)};\xi_k^j)  -  \nabla_y f_i(\Tilde{\bm{x}},\bm{y}_i^{(j)};\xi_i^j)  \right\|^2\right]\nonumber\\
  &  \leq \tau \sum_{j=r\tau}^{(r+1)\tau}\frac{\eta_x^2}{n} \sum_{i=1}^n \frac{1}{n}\sum_{k=1}^n\mathbb{E}\left[\left\|   \nabla_x f_k(\bm{x}_k^{(j)},\bm{y}_k^{(j)};\xi_k^j)  - \nabla_x f_k(\bm{x}_k^{(j)},\bm{y}_k^{(j)} ) +\nabla_x f_k(\bm{x}_k^{(j)},\bm{y}_k^{(j)} ) -\nabla_x f_k(\bm{x}^{(j)},\bm{y}^{(j)} )  \right.\right. \nonumber\\
  &\quad \left.\left. + \nabla_x f_k(\bm{x}^{(j)},\bm{y}^{(j)} ) -\nabla_x f_i(\bm{x}^{(j)},\bm{y}^{(j)} )+\nabla_x f_i(\bm{x}^{(j)},\bm{y}^{(j)} ) -\nabla_x f_i(\bm{x}_i^{(j)},\bm{y}_i^{(j)} ) +\nabla_x f_i(\bm{x}_i^{(j)},\bm{y}_i^{(j)} ) - \nabla_x f_i(\bm{x}_i^{(j)},\bm{y}_i^{(j)};\xi_i^t )  \right\|^2\right] \nonumber\\
  & \quad +\tau\sum_{j=r\tau}^{(r+1)\tau}\frac{\eta_y^2}{n}  \sum_{i=1}^n   \frac{1}{n}\sum_{k=1}^n \mathbb{E}\left[\left\| \nabla_y f_k(\Tilde{\bm{x}},\bm{y}_k^{(j)};\xi_k^j)  - \nabla_y f_k(\Tilde{\bm{x}},\bm{y}_k^{(j)} ) +\nabla_y f_k(\Tilde{\bm{x}},\bm{y}_k^{(j)} ) -\nabla_y f_k(\Tilde{\bm{x}},\bm{y}^{(j)} )  \right.\right. \nonumber\\
  & \quad \left.\left. + \nabla_y f_k(\Tilde{\bm{x}},\bm{y}^{(j)} )-\nabla_y f_i(\Tilde{\bm{x}},\bm{y}^{(j)} )+\nabla_y f_i(\Tilde{\bm{x}},\bm{y}^{(j)} ) -\nabla_y f_i(\Tilde{\bm{x}},\bm{y}_i^{(j)} ) +\nabla_y f_i(\Tilde{\bm{x}},\bm{y}_i^{(j)} ) - \nabla_y f_i(\Tilde{\bm{x}},\bm{y}_i^{(j)};\xi_i^j )  \right\|^2\right] \nonumber\\
  &  \leq \sum_{j=r\tau}^{(r+1)\tau}5\eta_x^2 \left( \sigma^2 + \frac{\sigma^2}{n} +2L^2\gamma^j + \zeta_x \right) + 5\eta_y^2 \left( \sigma^2 + \frac{\sigma^2}{n} +2L^2\gamma^j + \zeta_y \right). \nonumber  
 \end{align}
 
 Summing over $t$ from $r\tau$ to $(r+1)\tau$ yields:
 \begin{align}
  \sum_{t=r\tau}^{(r+1)\tau} \gamma^t &\leq   \sum_{t=r\tau}^{(r+1)\tau}\sum_{j=r\tau}^{(r+1)\tau}5\tau\eta_x^2 \left( \sigma^2 + \frac{\sigma^2}{n} +2L^2\gamma^j + \zeta_x \right) + 5\tau\eta_y^2 \left( \sigma^2 + \frac{\sigma^2}{n} +2L^2\gamma^j + \zeta_y \right)\nonumber\\
 & \leq  10L^2   \tau^2    (\eta_x^2 + \eta_y^2)\sum_{j=r\tau}^{(r+1)\tau} \gamma^j + 5\tau^2 (\eta_x^2 + \eta_y^2) \left( \sigma^2 + \frac{\sigma^2}{n}  \right)+5 \tau^2\eta_x^2\zeta_x+ 5 \tau^2 \eta_y^2 \zeta_y.  
 \end{align}
 Since $10L^2   \tau^2    (\eta_x^2 + \eta_y^2)\leq \frac{1}{2}$, by re-arranging the terms we have:
  \begin{align}
  \sum_{t=r\tau+1}^{(r+1)\tau} \gamma^t   \leq   10\tau^3 (\eta_x^2 + \eta_y^2) \left( \sigma^2 + \frac{\sigma^2}{n}  \right)+10\tau^3\eta_x^2\zeta_x+ 10\tau^3 \eta_y^2 \zeta_y .\nonumber
 \end{align}
 Summing over $r$ from $0$ to $T/\tau - 1$, and dividing both sides by $T$ can conclude the proof of the first statement:
  \begin{align}
 & \frac{1}{T}\sum_{t=1}^{T} \gamma^t   \leq   10\tau^2 (\eta_x^2 + \eta_y^2) \left( \sigma^2 + \frac{\sigma^2}{n}  \right)+10\tau^2\eta_x^2\zeta_x+ 10\tau^2 \eta_y^2 \zeta_y  .\nonumber
 \end{align} 
\end{proof}
\end{lemma}

\subsection{Proof of Theorem~\ref{Thm: NCNC}}\label{sec:proof_thm_ncnc}
According to Lemma~\ref{Lm: NCPL-1}, we sum over $t=1$ to $T$, and divide both sides with $T$:
{\begin{align}
\frac{1}{T}\left(\mathbb{E}\left[\Phi (\bm{x}^{(T+1)} )\right] - \mathbb{E}\left[\Phi(\bm{x}^{(1)})\right]  \right)
    & \leq   - \frac{\eta_x}{2}\frac{1}{T}\sum_{t=1}^T\mathbb{E}\left[\left\| \nabla \Phi(\bm{x}^{(t)})\right\|^2\right] - \left(\frac{\eta_x}{2}-\frac{ \beta\eta_x^2}{2} \right)\frac{1}{T}\sum_{t=1}^T\mathbb{E}\left[\left\|\frac{1}{n} \sum_{i=1}^n \nabla_x f_i(\bm{x}_i^{(t)}, \bm{y}_i^{(t)} ) \right\|^2\right] \nonumber\\
    & +   \frac{2\eta_x L^2}{\mu}\frac{1}{T}\sum_{t=1}^T\mathbb{E}\left[ (\Phi(\bm{x}^{(t)} )- F(\bm{x}^{(t)},  \bm{y}^{(t)}))\right]   + \frac{1}{T}\sum_{t=1}^T  2\eta_x  L^2  \mathbb{E}\left[ \delta_{\bm{x}}^{(t)} + \delta_{\bm{y}}^{(t)}\right]  +   \frac{ \beta \eta_x^2 \sigma^2}{2n}  .\nonumber
\end{align}}
Plugging in Lemma~\ref{Lm: NCPL-2} yields:

{\begin{align}
&\frac{1}{T}\left(\mathbb{E}\left[\Phi (\bm{x}^{(T+1)} )\right] - \mathbb{E}\left[\Phi(\bm{x}^{(1)})\right]  \right)\nonumber\\
    & \leq   - \underbrace{\left(\frac{\eta_x}{2}- \frac{4(1-\mu\eta_y)L^2}{\mu^2\eta_y}\eta_x^2\right)}_{\spadesuit}\frac{1}{T}\sum_{t=1}^T\mathbb{E}\left[\left\| \nabla \Phi(\bm{x}^{(t)})\right\|^2\right] + \frac{8\eta_x^3 L^4}{\mu^2}S(G_x^2+\sigma^2) + \frac{2\eta_x L^2}{\mu}\frac{\eta_y  L\sigma^2 }{ n}\nonumber\\
    &- \underbrace{\left(\frac{\eta_x}{2}-\frac{ \beta\eta_x^2}{2}-\frac{2\eta_x L^2}{\mu} \left[\frac{2(1-\mu\eta_y )}{\mu \eta_y}\left( \frac{\eta_x^2 L}{2}+\frac{ \beta\eta_x^2}{2}\right)+ L^2\eta_x^2 \right] \right)}_{\clubsuit}\frac{1}{T}\sum_{t=1}^T\mathbb{E}\left[\left\|\frac{1}{n} \sum_{i=1}^n \nabla_x f_i(\bm{x}_i^{(t)}, \bm{y}_i^{(t)} ) \right\|^2\right] \nonumber\\
    & +  \left( 2\eta_x L^2 + \frac{8\eta_x L^4}{\mu^2}+\frac{8(1-\mu\eta_y)L^4}{\mu^2 \eta_y} \eta_x^2\right) \frac{1}{T}\sum_{t=1}^T  \mathbb{E}\left[ \delta_{\bm{x}}^{(t)} + \delta_{\bm{y}}^{(t)}\right]  +  \left( \frac{8(1-\mu\eta_y)\eta_x L^2(L+\beta)}{\mu^2 \eta_y}  +  \beta+ \frac{8\eta_x L^4}{\mu^2}  \right) \frac{  \eta_x^2 \sigma^2}{2n}   \nonumber\\
    & + \frac{2\eta_x L^2}{\mu}\left(\frac{\mathbb{E}\left[\Phi (\bm{x}^{(0)} )\right] - \mathbb{E}\left[F(\bm{x}^{(0)},\bm{y}^{(0)})\right] }{\mu \eta_y T} + \left[\frac{2(1-\mu\eta_y )}{\mu \eta_y}\left( \frac{\eta_x^2 L}{2}+\frac{ \beta\eta_x^2}{2}\right)+ L^2\eta_x^2 \right]\frac{\mathbb{E}\left[\left\|\frac{1}{n} \sum_{i=1}^n \nabla_x f_i(\bm{x}^{(0)}, \bm{y}^{(0)} ) \right\|^2\right] }{T}\right) \nonumber\\
    & + \frac{2\eta_x L^2}{\mu}\frac{2(1-\mu\eta_y) }{\mu \eta_y}\left( \frac{\mathbb{E}\left[\left\| \nabla_x \Phi(\bm{x}^{(0)}) \right\|^2\right] }{T}\right)\nonumber.
\end{align}}
Recall that we choose: $\eta_x = \frac{n^{1/3}}{LT^{2/3}}$, $\eta_y = \frac{n^{1/3}}{LT^{1/2}}$, $\tau = \frac{T^{1/3}}{n^{2/3}}$, $S = \frac{T^{1/3}}{n^{2/3}}$, and
\begin{align}
    T \geq \max \left\{\left( \frac{\beta n^{1/3}}{2L} + \sqrt{\frac{\beta^2 n^{2/3}}{4L^2}+ \frac{8L(L+\beta)n^{1/3}}{\mu^2}} + \frac{4L^2 n^{2/3}}{\mu} \right)^{3/2}, (8\kappa^2)^6\right\} \nonumber,
\end{align}
so we know that $\spadesuit \geq \frac{\eta_x}{4}$ and $\clubsuit \geq 0$. Plugging in $\eta_x, \eta_y, \tau, S$, and plugging in Lemma~\ref{lm: nonconvex lm3} will conclude the proof for Theorem~\ref{Thm: NCNC}:
\begin{align}
    \frac{1}{T}\sum_{t=1}^T\mathbb{E}\left[\left\| \nabla \Phi(\bm{x}^{(t)})\right\|^2\right] \leq O\left(\frac{\beta \sigma^2}{(nT)^{1/3}}+\frac{\kappa^2 L^2 \zeta_y}{n^{2/3}T^{1/3}} + \frac{\kappa^2 L^2\zeta_x}{ n^{2/3}T} + \frac{\kappa^2 L^2 G^2_x}{ T}+\frac{\kappa^2 }{ n^{1/3}T^{1/2}}\right).
\end{align}
\qed

\section{Proof of Local SGDA+ under Nonconvex-One-Point-Concave Setting}\label{sec:ncoc}
\subsection{Overview of the proof techniques}
In this section we are going to present the proof of convergence of local SGDA+, under the setting that $F$ is nonconvex in $\bm{x}$ but one point concave in $\bm{y}$. In this setting, $\Phi(\bm{x})$ is no longer smooth any more, and $\bm{y}^{*}(\bm{x})$ is not Lipschitz. As we mentioned in the main paper, we study the Moreau evenlope function: $\Phi_{1/2L}(\bm{x})$. The proof mainly contains two parts: \textbf{one iteration analysis of Moreau envelope} and \textbf{Convergence of SGA under one point concave condition}.

\paragraph{Step I: One iteration analysis of Moreau envelope.}By examining one iteration of local SGDA+,
 we have the following relation:
\begin{align}
     \mathbb{E}[\Phi_{1/2L} (\bm{x}^{(t)})] &
\leq \mathbb{E}\left[\Phi_{1/2L} (\bm{x}^{(t-1)})\right]   +L \eta_x^2(G_x^2+\sigma_x^2) +   2\eta_x L^2 \frac{1}{n}\sum_{i=1}^n\mathbb{E}\left[ \left\|  \bm{x}^{(t-1)}_i  -  \bm{x}^{(t-1)}  \right\|^2 + \left\|  \bm{y}^{(t-1)}_i  -  \bm{y}^{(t-1)}  \right\|^2 \right]  \nonumber\\
    &+ 2L\eta_x\left( \mathbb{E}\left[\Phi( \bm{x}^{(t-1)} )\right] - \mathbb{E}\left[F( \bm{x}^{(t-1)},\bm{y}^{(t-1)})\right]  \right) - \frac{\eta_x}{8} \mathbb{E}\left[\left\|\nabla \Phi_{1/2L} (\bm{x}^{(t-1)}) \right\|^2\right].\nonumber
\end{align}
It turns out our next job is to bound local model deviation $\mathbb{E}\left[ \left\|  \bm{x}^{(t-1)}_i  -  \bm{x}^{(t-1)}  \right\| + \left\|  \bm{y}^{(t-1)}_i  -  \bm{y}^{(t-1)}  \right\| \right]$ and the gap $\mathbb{E}[\Phi(\bm{x}^{(t-1)})]- \mathbb{E}[F(\bm{x}^{(t-1)},\bm{y}^{(t-1)})]$. The the analysis of deviation term is similar to what we did in nonconvex-strongly-concave setting. The remaining tricky part is how to bound $\mathbb{E}[\Phi(\bm{x}^{(t-1)})]- \mathbb{E}[F(\bm{x}^{(t-1)},\bm{y}^{(t-1)})]$.

\paragraph{Step II: Convergence of SGA under one point concave condition.}To deal with $\mathbb{E}[\Phi(\bm{x}^{(t)})]- \mathbb{E}[F(\bm{x}^{(t)},\bm{y}^{(t)}]$, we first notice that:
\begin{align}
     \mathbb{E}[\Phi(\bm{x}^{(t)})]- \mathbb{E}[F(\bm{x}^{(t)},\bm{y}^{(t)})]
     &=\mathbb{E}[F(\bm{x}^{(t)},\bm{y}^*(\bm{x}^t))]- \mathbb{E}[F(\Tilde{\bm{x}} ,\bm{y}^*(\Tilde{\bm{x}}))]+\mathbb{E}[F(\Tilde{\bm{x}},\bm{y}^*(\Tilde{\bm{x}}))]- \mathbb{E}[F(\bm{x}^{(t)},\bm{y}^{(t)})] \nonumber \\
     &\leq \mathbb{E}[F(\bm{x}^{(t)},\bm{y}^*(\bm{x}^t))]- \mathbb{E}[F(\Tilde{\bm{x}} ,\bm{y}^*(\bm{x}^t))]+\mathbb{E}[F(\Tilde{\bm{x}},\bm{y}^*(\Tilde{\bm{x}}))]- \mathbb{E}[F(\bm{x}^{(t)},\bm{y}^{(t)})] \nonumber\\
      &\leq \underbrace{\mathbb{E}[F(\bm{x}^{(t)},\bm{y}^*(\bm{x}^t))]- \mathbb{E}[F(\Tilde{\bm{x}} ,\bm{y}^*(\bm{x}^t))]}_{T_1}+\underbrace{\mathbb{E}[F(\Tilde{\bm{x}},\bm{y}^*(\Tilde{\bm{x}}))]- \mathbb{E}[F(\Tilde{\bm{x}},\bm{y}^{(t)})]}_{T_2} \nonumber \\
      & +\underbrace{\mathbb{E}[F(\Tilde{\bm{x}},\bm{y}^{(t)})]- \mathbb{E}[F(\bm{x}^{(t)},\bm{y}^{(t)})]}_{T_3}\nonumber.
 \end{align}
According to the Lipschitz continuity of $F$, and the fact that $\Tilde{\bm{x}}$ will be updated every $S$ iterations, we can bound $T_1$ and $T_3$ by $\eta_x S G_x \sqrt{G_x^2 + \sigma^2}$. 

The tricky part is to handle $T_2$. Basically fixing $\Tilde{\bm{x}}$, we wish to know how fast $\mathbb{E}[F(\Tilde{\bm{x}},\bm{y}^{(t)})]$ converges to $\mathbb{E}[F(\Tilde{\bm{x}},\bm{y}^*(\Tilde{\bm{x}}))]$. Thanks to one point concave property and the updating rule of local SGDA+ where we fixed $\Tilde{\bm{x}}$ while updating $\bm{y}$, we can show that:
\begin{align}
     \sum_{t = kS+1}^{(k+1)S} \mathbb{E}\left[F(\Tilde{\bm{x}}, \bm{y}^*(\Tilde{\bm{x}}))-F(\Tilde{\bm{x}}, \bm{y}^{(t)}) \right]
    &\leq \frac{D}{\eta_y}   + L\sum_{t = kS+1}^{(k+1)S} \frac{1}{n} \mathbb{E}\left[\left\| \bm{y}^{(t)}_i-\bm{y}^{(t)}\right\|^2\right]  + 2\eta_y L^2 \sum_{t=kS+1}^{(k+1)S}\frac{1}{n} \mathbb{E}\left[\left\|\  \bm{y}^{(t)}_i-\bm{y}^{(t)}\right\|^2\right] + \frac{\eta_y S\sigma^2}{n} \nonumber.
\end{align}
 Putting these pieces together will conclude the proof.

\subsection{Proof of technical lemmas}

\begin{lemma}[One iteration analysis]\label{lm: nonconvex-oc lm1}
For local SGDA+, under Theorem~\ref{Thm: NCOC}'s assumption, the following statement holds:
\begin{align}
     \mathbb{E}[\Phi_{1/2L} (\bm{x}^{(t)})] &
\leq \mathbb{E}\left[\Phi_{1/2L} (\bm{x}^{(t-1)})\right]   +L \eta_x^2(G_x^2+\sigma_x^2) +   2\eta_x L^2 \frac{1}{n}\sum_{i=1}^n\mathbb{E}\left[ \left\|  \bm{x}^{(t-1)}_i  -  \bm{x}^{(t-1)}  \right\|^2 + \left\|  \bm{y}^{(t-1)}_i  -  \bm{y}^{(t-1)}  \right\|^2 \right]  \nonumber\\
    &+ 2L\eta_x\left( \mathbb{E}\left[\Phi( \bm{x}^{(t-1)} )\right] - \mathbb{E}\left[F( \bm{x}^{(t-1)},\bm{y}^{(t-1)})\right]  \right) - \frac{\eta_x}{8} \mathbb{E}\left[\left\|\nabla \Phi_{1/2L} (\bm{x}^{(t-1)}) \right\|^2\right].\nonumber
\end{align}
\begin{proof}
Define $\hat{\bm{x}}^{(t)} = \arg \min_{\bm{x}\in\mathcal{X}} \Phi(\bm{x}) + L\|\bm{x}-\bm{x}^{(t)}\|^2$, the by the definition of $\Phi_{1/2L}$ we have:
\begin{align}
    \Phi_{1/2L} (\bm{x}^{(t)}) \leq \Phi (\hat{\bm{x}}^{(t-1)}) + L\|\hat{\bm{x}}^{(t-1)}-\bm{x}^{(t)}\|^2.\label{eq: nonconvex-oc lm1 1}
\end{align}
Meanwhile according to updating rule we have:
\begin{align}
    \mathbb{E}\left[\left\|\hat{\bm{x}}^{(t-1)}-\bm{x}^{(t)}\right\|^2\right] &=    \mathbb{E}\left[\left\|\bm{x}^{(t-1)} - \eta_x\frac{1}{n}\sum_{i =1}^n \nabla_x f_i(\bm{x}^{(t-1)}_i,\bm{y}^{(t-1)}_i;\xi_i^{(t)})\right\|^2\right]\nonumber\\
    &\leq \mathbb{E}\left[\left\|\hat{\bm{x}}^{(t-1)}-\bm{x}^{(t-1)} \right\|^2\right] + \eta_x^2\mathbb{E}\left[\left\|\frac{1}{n}\sum_{i =1}^n \nabla_x f_i(\bm{x}^{(t-1)}_i,\bm{y}^{(t-1)}_i;\xi_i^{(t)}) \right\|^2\right]  \nonumber\\
    & \quad+ 2\eta_x\mathbb{E}\left[\left\langle  \hat{\bm{x}}^{(t-1)}-\bm{x}^{(t-1)}, \frac{1}{n}\sum_{i =1}^n \nabla_x f_i(\bm{x}^{(t-1)}_i,\bm{y}^{(t-1)}_i )\right\rangle\right]\nonumber\\
    & \leq \mathbb{E}\left[\left\|\hat{\bm{x}}^{(t-1)}-\bm{x}^{(t-1)} \right\|^2\right] + \eta_x^2(G_w^2+\sigma_w^2) + 2\eta_x\left\langle  \hat{\bm{x}}^{(t-1)}-\bm{x}^{(t-1)}, \frac{1}{n}\sum_{i =1}^n \nabla_x f_i(\bm{x}^{(t-1)},\bm{y}^{(t-1)})\right\rangle \nonumber\\
    &\quad+ \eta_x \left( \frac{L}{2} \mathbb{E}\left[\left\|\hat{\bm{x}}^{(t-1)}-\bm{x}^{(t-1)} \right\|^2 \right] + \frac{2}{L}\mathbb{E}\left[ \frac{1}{n}\sum_{i =1}^n \left\|\nabla_x f_i(\bm{x}^{(t-1)}_i,\bm{y}^{(t-1)}_i) - \nabla_x f_i(\bm{x}^{(t-1)},\bm{y}^{(t-1)} )\right\|^2 \right]\right) \nonumber\\
     & \leq \mathbb{E}\left[\left\|\hat{\bm{x}}^{(t-1)}-\bm{x}^{(t-1)} \right\|^2\right] + \eta_x^2(G_w^2+\sigma_w^2) +  \eta_x  2L \frac{1}{n}\sum_{i =1}^n\mathbb{E}\left[ \left\|  \bm{x}^{(t-1)}_i  -  \bm{x}^{(t-1)}  \right\|^2 + \left\|  \bm{y}^{(t-1)}_i  -  \bm{y}^{(t-1)}  \right\|^2 \right] \nonumber\\
    &\quad + 2\eta_x\mathbb{E}\left[\left\langle  \hat{\bm{x}}^{(t-1)}-\bm{x}^{(t-1)},  \nabla_x F(\bm{x}^{(t-1)},\bm{y}^{(t-1)}  )\right\rangle\right]+\frac{\eta_x L}{2} \mathbb{E}\left[\left\|\hat{\bm{x}}^{(t-1)}-\bm{x}^{(t-1)} \right\|^2 \right] .\label{eq: nonconvex-oc lm1 2}
\end{align}

According to smoothness of $F$ we obtain:
 \begin{align}
     &\mathbb{E}\left[\left\langle  \hat{\bm{x}}^{(t-1)}-\bm{x}^{(t-1)},  \nabla_x F(\bm{x}^{(t-1)},\bm{y}^{(t-1)}  )\right\rangle\right] \nonumber\\
     &\leq \mathbb{E}\left[F(\hat{\bm{x}}^{(t-1)},\bm{y}^{(t-1)})\right] - \mathbb{E}\left[F( \bm{x}^{(t-1)},\bm{y}^{(t-1)})\right] + \frac{L}{2}\mathbb{E}\left[\left\|\hat{\bm{x}}^{(t-1)}-\bm{x}^{(t-1)} \right\|^2\right] \nonumber\\
     & \leq \mathbb{E}\left[\Phi(\hat{\bm{x}}^{(t-1)} )\right]- \mathbb{E}\left[F( \bm{x}^{(t-1)},\bm{y}^{(t-1)})\right]+ \frac{L}{2}\mathbb{E}\left[\left\|\hat{\bm{x}}^{(t-1)}-\bm{x}^{(t-1)} \right\|^2\right]  \nonumber\\
       & \leq \underbrace{ \mathbb{E}\left[\Phi(\hat{\bm{x}}^{(t-1)} )\right]+ L\mathbb{E}\left[\left\|\hat{\bm{x}}^{(t-1)}-\bm{x}^{(t-1)} \right\|^2\right]}_{\leq \mathbb{E}\left[\Phi( \bm{x}^{(t-1)} )\right]+ L\mathbb{E}\left[\left\|\bm{x}^{(t-1)}-\bm{x}^{(t-1)} \right\|^2\right]} - \mathbb{E}\left[F( \bm{x}^{(t-1)},\bm{y}^{(t-1)})\right] - \frac{L}{2}\mathbb{E}\left[\left\|\hat{\bm{x}}^{(t-1)}-\bm{x}^{(t-1)} \right\|^2\right] \nonumber\\
     & \leq \mathbb{E}\left[\Phi( \bm{x}^{(t-1)} )\right] - \mathbb{E}\left[F( \bm{x}^{(t-1)},\bm{y}^{(t-1)})\right] - \frac{L}{2}\mathbb{E}\left[\left\|\hat{\bm{x}}^{(t-1)}-\bm{x}^{(t-1)} \right\|^2\right].\label{eq: nonconvex-oc lm1 3}
\end{align} 
Plugging (\ref{eq: nonconvex-oc lm1 2}) and (\ref{eq: nonconvex-oc lm1 3}) into (\ref{eq: nonconvex-oc lm1 1}) yields:
 \begin{align}
   \mathbb{E}\left[ \Phi_{1/2L} (\bm{x}^{(t)})\right] &\leq \mathbb{E}\left[\Phi (\hat{\bm{x}}^{(t-1)}) \right]+ L  \mathbb{E}\left[\left\|\hat{\bm{x}}^{(t-1)}-\bm{x}^{(t-1)} \right\|^2\right] +  2\eta_x L^2 \frac{1}{n}\sum_{i=1}^n\mathbb{E}\left[ \left\|  \bm{x}^{(t-1)}_i  -  \bm{x}^{(t-1)}  \right\|^2 + \left\|  \bm{y}^{(t-1)}_i  -  \bm{y}^{(t-1)}  \right\|^2 \right]  \nonumber\\
    &+ 2\eta_x L \left( \mathbb{E}\left[\Phi( \bm{x}^{(t-1)} )\right] - \mathbb{E}\left[F( \bm{x}^{(t-1)},\bm{y}^{(t-1)})\right] - \frac{L}{2}\mathbb{E}\left[\left\|\hat{\bm{x}}^{(t-1)}-\bm{x}^{(t-1)} \right\|^2\right]\right) \nonumber\\
    &+L \eta_x^2(G_w^2+\sigma_w^2)+\frac{\eta_x L^2}{2}\mathbb{E}\left[\left\|\hat{\bm{x}}^{(t-1)}-\bm{x}^{(t-1)} \right\|^2\right]\nonumber\\
    & \leq \mathbb{E}\left[\Phi_{1/2L} (\bm{x}^{(t-1)})\right]   +L \eta_x^2(G_w^2+\sigma_w^2) +   2\eta_x L^2 \frac{1}{n}\sum_{i=1}^n\mathbb{E}\left[ \left\|  \bm{x}^{(t-1)}_i  -  \bm{x}^{(t-1)}  \right\|^2 + \left\|  \bm{y}^{(t-1)}_i  -  \bm{y}^{(t-1)}  \right\|^2 \right]  \nonumber\\
    &+ 2L\eta_x\left( \mathbb{E}\left[\Phi( \bm{x}^{(t-1)} )\right] - \mathbb{E}\left[F( \bm{x}^{(t-1)},\bm{y}^{(t-1)})\right]  \right) - \frac{\eta_x}{8} \mathbb{E}\left[\left\|\nabla \Phi_{1/2L} (\bm{x}^{(t-1)}) \right\|^2\right],\nonumber
\end{align} 
where we use the result from Lemma 2.8 in~\cite{lin2019gradient}: $\nabla \Phi_{1/2L}(\bm{x}) = 2L (\bm{x}-\hat{\bm{x}})$.
\end{proof}
\end{lemma}

The following lemma derives the convergence rate of the gap $\mathbb{E}[\Phi(\bm{x}^{(t)})]- \mathbb{E}[F(\bm{x}^{(t)},\bm{y}^{(t)})] $.
\begin{lemma}\label{lm: nonconvex-oc lm2}
For local $SGDA+$, under Theorem~\ref{Thm: NCOC}'s assumption,  the following statement holds:
\begin{align}
\frac{1}{T}\sum_{t= 1}^{T} \mathbb{E}[\Phi(\bm{x}^{(t)})]- \mathbb{E}[F(\bm{x}^{(t)},\bm{y}^{(t)})]  
      &\leq 2\eta_x SG_x\sqrt{G_x^2 + \sigma^2} +\frac{D}{S\eta_y}   + (L+4\eta_y L^2) \frac{1}{T}\sum_{t=1}^T\frac{1}{n}\sum_{i=1}^n \mathbb{E}\left[\left\| \bm{y}^{(t)}_i-\bm{y}^{(t)}\right\|^2\right]  + \frac{\eta_y \sigma^2}{n}\nonumber.
\end{align}
\begin{proof}
 Consider $t = kS+1$ to $(k+1)S$. Let $\Tilde{\bm{x}}$ denote the latest snapshot iterate. Observe that:
 \begin{align}
     \mathbb{E}[\Phi(\bm{x}^{(t)})]- \mathbb{E}[F(\bm{x}^{(t)},\bm{y}^{(t)})] &\leq \mathbb{E}[F(\bm{x}^{(t)},\bm{y}^*(\bm{x}^t))]- \mathbb{E}[F(\Tilde{\bm{x}} ,\bm{y}^*(\bm{x}^t))]+\mathbb{E}[F(\Tilde{\bm{x}},\bm{y}^*(\Tilde{\bm{x}}))]- \mathbb{E}[F(\bm{x}^{(t)},\bm{y}^{(t)})] \nonumber\\
     &\leq G_x \mathbb{E}\|\bm{x}^{(t)} - \Tilde{\bm{x}}\| +\mathbb{E}[F(\Tilde{\bm{x}},\bm{y}^*(\Tilde{\bm{x}})]- \mathbb{E}[F(\Tilde{\bm{x}},\bm{y}^{(t)} )] +\mathbb{E}[F(\Tilde{\bm{x}},\bm{y}^{(t)})]- \mathbb{E}[F(\bm{x}^{(t)},\bm{y}^{(t)}]\nonumber\\
      &\leq 2\eta_x SG_x \sqrt{G_x^2 + \sigma^2}  + \mathbb{E}[F(\Tilde{\bm{x}},\bm{y}^*(\Tilde{\bm{x}})]- \mathbb{E}[F(\Tilde{\bm{x}},\bm{y}^{(t)} )]\label{eq: ncnc lm2 eq0}.  
 \end{align}
 where we use the fact $f(\cdot,\bm{y})$ is $G_x$-Lipschitz, so that:
 \begin{align*}
     &\mathbb{E}[F(\bm{x}^{(t)},\bm{y}^*(\bm{x}^t))]- \mathbb{E}[F(\Tilde{\bm{x}} ,\bm{y}^*(\bm{x}^t))] \leq G_x \mathbb{E} \|\bm{x}^{(t)} - \Tilde{\bm{x}}\| \leq \eta_xS G_x \sqrt{G_x^2 + \sigma^2},\\
     &\mathbb{E}[F( \Tilde{\bm{x}},\bm{y}^{(t)})]- \mathbb{E}[F(\bm{x}^{(t)} ,\bm{y}^{(t)})] \leq G_x \mathbb{E} \|\bm{x}^{(t)} - \Tilde{\bm{x}}\| \leq \eta_x SG_x \sqrt{G_x^2 + \sigma^2}.
 \end{align*}
Summing over $t = kS+1$ to $(k+1)S$ in (\ref{eq: ncnc lm2 eq0}), and dividing both sides with $T$ yields:
  \begin{align}
      \sum_{t= kS}^{(k+1)S-1} \mathbb{E}[\Phi(\bm{x}^{(t)})]- \mathbb{E}[F(\bm{x}^{(t)},\bm{y}^{(t)})]  
      &\leq 2\eta_x S^2G_x \sqrt{G_x^2 + \sigma^2}  +  \sum_{t= kS}^{(k+1)S-1} \mathbb{E}[F(\Tilde{\bm{x}},\bm{y}^*(\Tilde{\bm{x}})]- \mathbb{E}[F(\Tilde{\bm{x}},\bm{y}^{(t)} )].\label{eq: ncnc-oc lm2 eq2}  
 \end{align}
 
 Now let us study the convergence of $\mathbb{E}[F(\Tilde{\bm{x}},\bm{y}^*(\Tilde{\bm{x}})]- \mathbb{E}[F(\Tilde{\bm{x}},\bm{y}^{(t)} )]$.

 By the updating rule of $\bm{y}$ we have:
 
 \begin{align}
  & \mathbb{E}\left[ \| \bm{y}^{(t+1)} - \bm{y}^*(\Tilde{\bm{x}}) \|^2 \right]\nonumber\\
  &= \mathbb{E}\left[\left \| \bm{y}^{(t)} + \eta_y \frac{1}{n}\sum_{i=1}^n\nabla_y f_i(\Tilde{\bm{x}}, \bm{y}^{(t)}_i;\xi_i^t)- \bm{y}^*(\Tilde{\bm{x}})\right\|^2 \right]\nonumber\\
   &= \mathbb{E}\left[\left \| \bm{y}^{(t)}  - \bm{y}^*(\Tilde{\bm{x}})\right\|^2 \right] + 2\eta_y\mathbb{E}\left[ \left\langle\frac{1}{n}\sum_{i=1}^n\nabla_y f_i(\Tilde{\bm{x}}, \bm{y}^{(t)}_i;\xi_i^t),  \bm{y}^{(t)} - \bm{y}^*(\Tilde{\bm{x}})\right\rangle\right]  + \eta_y^2\mathbb{E}\left[ \left\|\frac{1}{n}\sum_{i=1}^n\nabla_y f_i(\Tilde{\bm{x}}, \bm{y}^{(t)}_i;\xi_i^t)\right\|^2\right] \nonumber\\ 
   &\leq \mathbb{E}\left[\left \| \bm{y}^{(t)}  - \bm{y}^*(\Tilde{\bm{x}})\right\|^2 \right] + 2\eta_y\mathbb{E}\left[ \left\langle\frac{1}{n}\sum_{i=1}^n\nabla_y f_i(\Tilde{\bm{x}}, \bm{y}^{(t)}_i),  \bm{y}^{(t)} -\bm{y}^{(t)}_i  \right\rangle\right] \nonumber\\
   &  + 2\eta_y\mathbb{E}\left[ \left\langle\frac{1}{n}\sum_{i=1}^n\nabla_y f_i(\Tilde{\bm{x}}, \bm{y}^{(t)}_i),  \bm{y}^{(t)}_i- \bm{y}^*(\Tilde{\bm{x}})\right\rangle \right]+ \eta_y^2 \mathbb{E}\left[\left\|\frac{1}{n}\sum_{i=1}^n\nabla_y f_i(\Tilde{\bm{x}}, \bm{y}^{(t)}_i)\right\|^2\right] + \frac{\eta_y^2 \sigma^2}{n}\nonumber.
 \end{align}
 Applying one point concavity and $L$-smoothness of $f_i(\Tilde{\bm{x}}, \cdot)$ we have:
  \begin{align}
   \mathbb{E}\left[ \| \bm{y}^{(t+1)} - \bm{y}^*(\Tilde{\bm{x}}) \|^2 \right] &\leq \mathbb{E}\left[\left \| \bm{y}^{(t)}  - \bm{y}^*(\Tilde{\bm{x}})\right\|^2 \right] + 2\eta_y  \frac{1}{n}\sum_{i=1}^n \mathbb{E}\left[f_i(\Tilde{\bm{x}}, \bm{y}^{(t)}) -f_i(\Tilde{\bm{x}}, \bm{y}^*(\Tilde{\bm{x}}))\right]  + \eta_y L \frac{1}{n} \sum_{i=1}^n\mathbb{E}\left[\left\| \bm{y}^{(t)}_i-\bm{y}^{(t)}\right\|^2\right]  \nonumber\\
   &  + 4\eta_y^2 L  \mathbb{E}\left[F(\Tilde{\bm{x}}, \bm{y}^*(\Tilde{\bm{x}}))- F(\Tilde{\bm{x}}, \bm{y}^{(t)})\right] + 2\eta_y^2 L^2 \frac{1}{n} \sum_{i=1}^n\mathbb{E}\left[\left\|\  \bm{y}^{(t)}_i-\bm{y}^{(t)}\right\|^2\right]+ \frac{\eta_y^2 \sigma^2}{n}\nonumber\\
   &\leq \mathbb{E}\left[\left \| \bm{y}^{(t)}  - \bm{y}^*(\Tilde{\bm{x}})\right\|^2 \right] + \underbrace{(2\eta_y- 4\eta_y^2 L)}_{\geq \eta_y} \mathbb{E}\left[F(\Tilde{\bm{x}}, \bm{y}^{(t)}) -F(\Tilde{\bm{x}}, \bm{y}^*(\Tilde{\bm{x}}))\right]+ \frac{\eta_y^2 \sigma^2}{n} \nonumber\\
   &+ \eta_y L \frac{1}{n}\sum_{i=1}^n \mathbb{E}\left[\left\| \bm{y}^{(t)}_i-\bm{y}^{(t)}\right\|^2\right]  + 2\eta_y^2 L^2 \frac{1}{n} \mathbb{E}\left[\left\|\  \bm{y}^{(t)}_i-\bm{y}^{(t)}\right\|^2\right]\nonumber\\
   &\leq \mathbb{E}\left[\left \| \bm{y}^{(t)}  - \bm{y}^*(\Tilde{\bm{x}})\right\|^2 \right] - \eta_y    \mathbb{E}\left[F(\Tilde{\bm{x}}, \bm{y}^*(\Tilde{\bm{x}}))-F(\Tilde{\bm{x}}, \bm{y}^{(t)}) \right]+ \frac{\eta_y^2 \sigma^2}{n} \nonumber\\
   &+ \eta_y L \frac{1}{n} \sum_{i=1}^n \mathbb{E}\left[\left\| \bm{y}^{(t)}_i-\bm{y}^{(t)}\right\|^2\right]  + 2\eta_y^2 L^2 \frac{1}{n} \sum_{i=1}^n\mathbb{E}\left[\left\|\  \bm{y}^{(t)}_i-\bm{y}^{(t)}\right\|^2\right]\nonumber. 
 \end{align}
 Re-arranging the terms, and summing $t = kS+1$ to $(k+1)S$ yields:
   \begin{align}
 \sum_{t = kS+1}^{(k+1)S} \mathbb{E}\left[F(\Tilde{\bm{x}}, \bm{y}^*(\Tilde{\bm{x}}))-F(\Tilde{\bm{x}}, \bm{y}^{(t)}) \right]
   &\leq \frac{1}{\eta_y} \left(\mathbb{E}\left[\left \| \bm{y}^{(kS+1)}  - \bm{y}^*(\Tilde{\bm{x}})\right\|^2 \right] -  \mathbb{E}\left[ \| \bm{y}^{((k+1)S)} - \bm{y}^*(\Tilde{\bm{x}}) \|^2 \right]\right) + \frac{\eta_y S\sigma^2}{n}  \nonumber\\
   &+ L \sum_{t = kS+1}^{(k+1)S}\frac{1}{n} \mathbb{E}\left[\left\| \bm{y}^{(t)}_i-\bm{y}^{(t)}\right\|^2\right]  + 2\eta_y L^2 \sum_{t = kS+1}^{(k+1)S}\frac{1}{n} \mathbb{E}\left[\left\|\  \bm{y}^{(t)}_i-\bm{y}^{(t)}\right\|^2\right]\nonumber \\
    &\leq \frac{D}{\eta_y}   + L\sum_{t = kS+1}^{(k+1)S} \frac{1}{n} \mathbb{E}\left[\left\| \bm{y}^{(t)}_i-\bm{y}^{(t)}\right\|^2\right]  + 2\eta_y L^2 \sum_{t=kS+1}^{(k+1)S}\frac{1}{n} \mathbb{E}\left[\left\|\  \bm{y}^{(t)}_i-\bm{y}^{(t)}\right\|^2\right] + \frac{\eta_y S\sigma^2}{n} \nonumber.
 \end{align}
 Plugging above bound into (\ref{eq: ncnc-oc lm2 eq2}) yields:
   \begin{align}
      \sum_{t= kS}^{(k+1)S-1} \mathbb{E}[\Phi(\bm{x}^{(t)})]- \mathbb{E}[F(\bm{x}^{(t)},\bm{y}^{(t)}]  
      &\leq 2\eta_x S^2G_x\sqrt{G_x^2 + \sigma^2} +\frac{D}{\eta_y}   + (L+4\eta_y L^2) \sum_{t = kS+1}^{(k+1)S}\frac{1}{n}\sum_{i=1}^n \mathbb{E}\left[\left\| \bm{y}^{(t)}_i-\bm{y}^{(t)}\right\|^2\right]  + \frac{S\eta_y \sigma^2}{n}\nonumber. 
 \end{align}
 Finally, summing $k = 0$ to $T/S-1$, and dividing both sides by $T$ will conclude the proof:
 \begin{align}
      \frac{1}{T}\sum_{t= 1}^{T} \mathbb{E}[\Phi(\bm{x}^{(t)})]- \mathbb{E}[F(\bm{x}^{(t)},\bm{y}^{(t)})]  
      &\leq 2\eta_x SG_x\sqrt{G_x^2 + \sigma^2} +\frac{D}{S\eta_y}   + (L+4\eta_y L^2) \frac{1}{T}\sum_{t=1}^T\frac{1}{n}\sum_{i=1}^n \mathbb{E}\left[\left\| \bm{y}^{(t)}_i-\bm{y}^{(t)}\right\|^2\right]  + \frac{\eta_y \sigma^2}{n}\nonumber. 
 \end{align}

\end{proof}
\end{lemma}

\subsection{Proof of Theorem~\ref{Thm: NCOC}}\label{sec:proof_thm_ncoc}
In this section we provide the full proof of Theorem~\ref{Thm: NCOC}. We first sum over $t = 1$ to $T$ in Lemma~\ref{lm: nonconvex-oc lm1}, and divide both sides with $T$:
 \begin{align}
     \frac{1}{T} \sum_{t=1}^T\mathbb{E}\left[\left\|\nabla \Phi_{1/2L} (\bm{x}^{(t)})\right\|^2\right]  &
\leq \frac{8\mathbb{E}[\Phi_{1/2L} (\bm{x}^{(0)})] - 8\mathbb{E}[\Phi_{1/2L} (\bm{x}^{(T)})]}{\eta_x T} + 16\frac{1}{T}\sum_{t=1}^T L^2 \mathbb{E}\left[ \frac{1}{n}\sum_{i=1}^n \left\|   \bm{x}^{(t)}_i  -  \bm{x}^{(t)}  \right\| + \left\|   \bm{y}^{(t)}_i  -  \bm{y}^{(t-1)}  \right\|\right] \nonumber \\
    &\quad + 16 L\frac{1}{T}\sum_{t=1}^T \left( \mathbb{E}[\Phi(\bm{x}^{(t)})]- \mathbb{E}[F(\bm{x}^{(t)},\bm{y}^{(t)}]  \right)+8L \eta_x^2(G^2_x+\sigma^2).\nonumber
\end{align} 
Plugging in Lemma~\ref{lm: nonconvex-oc lm2} and \ref{lm: nonconvex lm3} yields:
 \begin{align}
     & \frac{1}{T} \sum_{t=1}^T\mathbb{E}\left[\left\|\nabla \Phi_{1/2L} (\bm{x}^{(t)})\right\|^2\right]\nonumber\\
     & \leq \frac{8\mathbb{E}[\Phi_{1/2L} (\bm{x}^{(0)})] }{\eta_x T} +      16 L^2 \left(  10\tau^2 (\eta_x^2 + \eta_y^2) \left( \sigma^2 + \frac{\sigma^2}{n}  \right)+10\tau^2\eta_x^2\zeta_x+ 10\tau^2 \eta_y^2 \zeta_y   \right)+8L \eta_x(G^2_x+\sigma^2) \nonumber \\
    &+ 8 L  \left( 2\eta_x SG_x\sqrt{G_x^2 + \sigma^2} +\frac{D}{S\eta_y}   + (L+4\eta_y L^2) \frac{1}{T}\sum_{t=1}^T\frac{1}{n}\sum_{i=1}^n \mathbb{E}\left[\left\| \bm{y}^{(t)}_i-\bm{y}^{(t)}\right\|^2\right]  + \frac{\eta_y \sigma^2}{n} \right).\nonumber\\
    &  \leq \frac{8\mathbb{E}[\Phi_{1/2L} (\bm{x}^{(0)})] }{\eta_x T} +      16 L^2 \left(  10\tau^2 (\eta_x^2 + \eta_y^2) \left( \sigma^2 + \frac{\sigma^2}{n}  \right)+10\tau^2\eta_x^2\zeta_x+ 10\tau^2 \eta_y^2 \zeta_y   \right)+8L \eta_x(G^2_x+\sigma^2) \nonumber \\
    &+ 8 L  \left( 2\eta_x SG_x\sqrt{G_x^2 + \sigma^2} +\frac{D}{S\eta_y}   + (L+4\eta_y L^2)  \left(10\tau^2 (\eta_x^2 + \eta_y^2) \left( \sigma^2 + \frac{\sigma^2}{n}  \right)+10\tau^2\eta_x^2\zeta_x+ 10\tau^2 \eta_y^2 \zeta_y    \right)  + \frac{\eta_y \sigma^2}{n} \right)\nonumber\\ 
    &  \leq \frac{8\mathbb{E}[\Phi_{1/2L} (\bm{x}^{(0)})] }{\eta_x T} +      (16 L^2 +8 L(L+4\eta_y L^2) )  \left(  10\tau^2 (\eta_x^2 + \eta_y^2) \left( \sigma^2 + \frac{\sigma^2}{n}  \right)+10\tau^2\eta_x^2\zeta_x+ 10\tau^2 \eta_y^2 \zeta_y   \right)+8L \eta_x(G^2_x+\sigma^2) \nonumber \\
    &+ 8 L  \left( 2\eta_x SG_x\sqrt{G_x^2 + \sigma^2} +\frac{D}{S\eta_y}    + \frac{\eta_y \sigma^2}{n} \right)\nonumber  
\end{align} 
If we choose $\eta_x = \frac{1}{LT^{\frac{5}{6}}}$, $\eta_y = \frac{1}{4LT^{\frac{1}{2}}}$, $\tau = T^{\frac{1}{3}}/n^{\frac{1}{6}}$, $S = T^{\frac{2}{3}}$ we recover the rate:
 \begin{align}
    \frac{1}{T}\sum_{t=1}^T \mathbb{E}\left[\left\|\nabla \Phi_{1/2L} (\bm{x}^{(t)})\right\|^2\right] 
    & \leq  O\left(\frac{L\sigma^2 }{T^{\frac{1}{6}}}\right) +  O\left(\frac{D }{T^{\frac{1}{6}}}\right)+    O\left(  \frac{L^2 \sigma^2}{(nT)^{\frac{1}{3}}} + \frac{L^2\zeta_x}{n^{\frac{1}{3}}T}   + \frac{L^2\zeta_y}{(nT)^{\frac{1}{3}}}\right)  +O\left(\frac{LG_x^2}{T^{\frac{1}{6}}}\right)  +O\left(\frac{\sigma^2}{nT^{\frac{1}{6}}}\right) \nonumber, 
\end{align} 
as stated by the theorem.
\qed

\end{document}